\documentclass[10pt,twocolumn,letterpaper]{article}

\usepackage{cvpr}              % To produce the CAMERA-READY version
\usepackage[T1]{fontenc}

\usepackage{multirow}
\usepackage{graphicx}
\usepackage{amsmath}
\usepackage{svg}
\usepackage{ragged2e}
\usepackage{twemojis}
\usepackage{cuted}
\usepackage{wrapfig}
\usepackage{marvosym}
\usepackage[export]{adjustbox}
\usepackage{rotating}
\usepackage{soul}

\usepackage{dblfloatfix}

\usepackage[nolist,nohyperlinks]{acronym}

\begin{acronym}
\acro{llm}[LLM]{Large Language Model}
\acro{mllm}[MLLM]{Multimodal Large Language Model}
\acro{vlm}[VLM]{Vision-Language Model}
\acro{mcq}[MCQ]{Multiple Choice Questions}
\end{acronym}

\newcommand{\cgpt}[1]{GPT-4o}
\newcommand{\datasub}[1]{label categories}
\newcommand{\qwen}[1]{Qwen3-VL}
\newcommand{\gt}[1]{ImGT}
\newcommand{\regt}[1]{ReGT}

\definecolor{cvprblue}{rgb}{0.21,0.49,0.74}
\usepackage[pagebackref,breaklinks,colorlinks,allcolors=cvprblue]{hyperref}
\usepackage{amsmath}

\def\paperID{} % *** Enter the Paper ID here
\def\confName{CVPR}
\def\confYear{2026}

\title{Multimodal Large Language Models as Image Classifiers}

\author{Nikita Kisel$^*$~~ Illia Volkov$^*$~~ Klara Janouskova\textsuperscript{\Letter}~~ Jiri Matas\\\\
Visual Recognition Group, Czech Technical University in Prague \\
{\tt\small kiselnik@fel.cvut.cz, volkoill@cvut.cz, janoukl1@fel.cvut.cz, matas@fel.cvut.cz} \\
{\tt\small $^*$Equal contribution~~ \textsuperscript{\Letter}Corresponding author}
}

\usepackage{svg}
\setsvg{inkscapelatex=true}

\begin{document}
\maketitle
\begin{abstract}
\ac{mllm} classification performance depends critically on evaluation protocol and ground truth quality.  Studies comparing \acp{mllm} with supervised and \acp{vlm} report conflicting conclusions, and we show these conflicts stem from protocols that either inflate or underestimate performance. Across the most common evaluation protocols, we identify and fix key issues: model outputs that fall outside the provided class list and are discarded, inflated results from weak multiple-choice distractors, and open-world setting that underperforms only due to poor output mapping. We additionally quantify the impact of commonly overlooked design choices —  batch size, image ordering, and text encoder selection — showing they substantially affect accuracy.

Evaluating on ReGT, our multilabel reannotation of 625 ImageNet-1k classes, reveals that \acp{mllm} benefit most from corrected labels (up to +10.8\%), substantially narrowing the perceived gap with supervised models. Much of the reported \ac{mllm} underperformance on classification is thus an artifact of noisy ground truth and flawed evaluation protocol rather than genuine model deficiency. Models less reliant on supervised training signals prove most sensitive to annotation quality. Finally, we show that \acp{mllm} can assist human annotators: in a controlled case study, annotators confirmed or integrated \ac{mllm} predictions in approximately 50\% of difficult cases, demonstrating their potential for large-scale dataset curation. This work is part of the Aiming for Perfect ImageNet-1k project, see \href{https://klarajanouskova.github.io/ImageNet/}{\nolinkurl{klarajanouskova.github.io/ImageNet}}.

\end{abstract}    
\section{Introduction}
\label{sec:intro}

\begin{figure}[!htb]
    \centering
    \includegraphics[width=1.0\columnwidth]{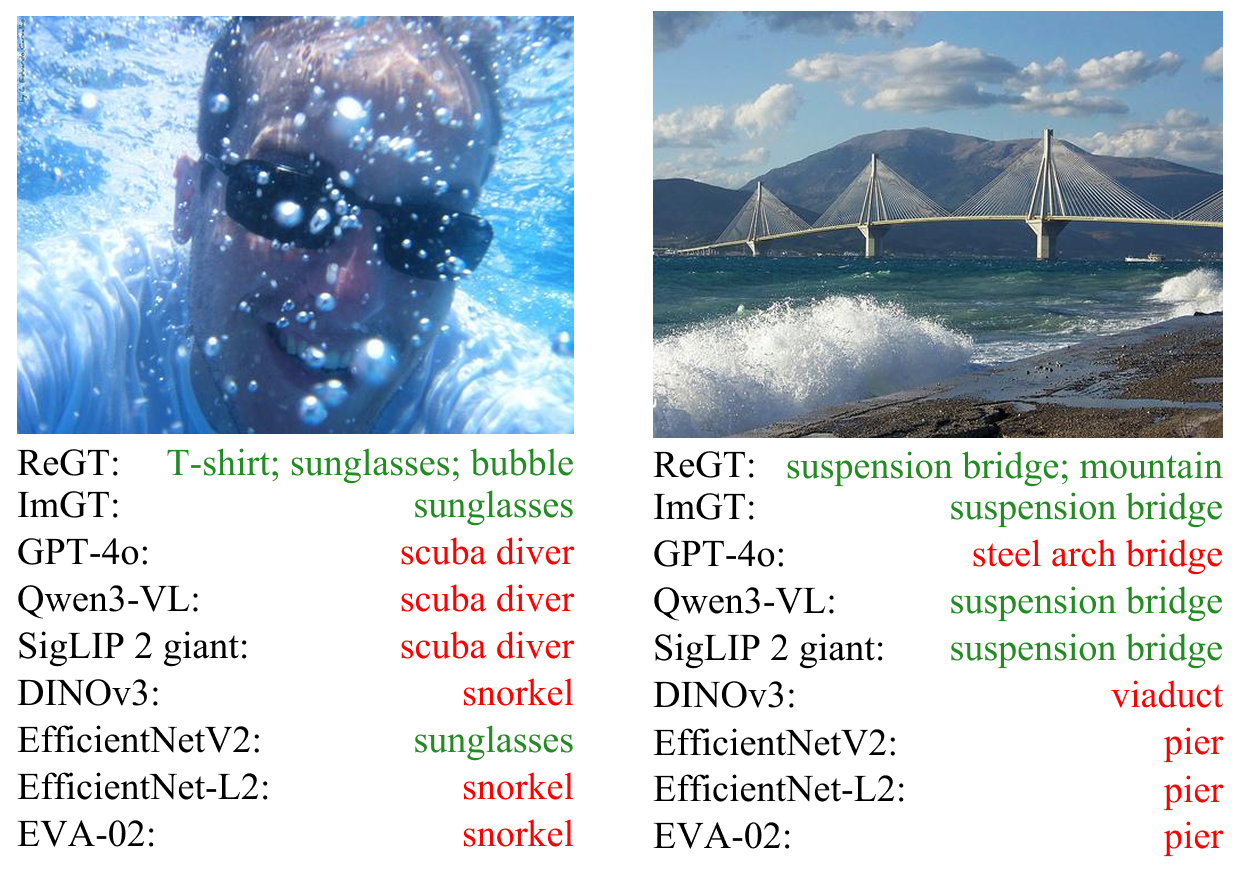}
    \caption{\hspace{-0.5ex}Challenging visual recognition cases from ImageNet-1k.\\
    ImGT: original single ImageNet label, ReGT: our reannotations. 
    Predictions of representative models, including DINOv3, EfficientNetV2, EfficientNet-L2, EVA-02 (self-)supervised on ImageNet, are often \textcolor{red}{wrong} on such data.
    % A trained human annotator, however, recognizes all classes. 
    Correct predictions in \textcolor{ForestGreen}{green}.
    }
    \label{fig:cover}
\end{figure}

%  LLMs: Language-only to multimodal
%% \acfp{llm} have transformed the way we work.
% %and 
% They have increasingly demonstrated competence in knowledge-intensive domains that traditionally required human expertise, including summarization, translation, legal text analysis, medical reasoning, and programming.

% 

\begin{figure*}[t!]
    \centering
    \resizebox{1.0\linewidth}{!}{%
        \includegraphics{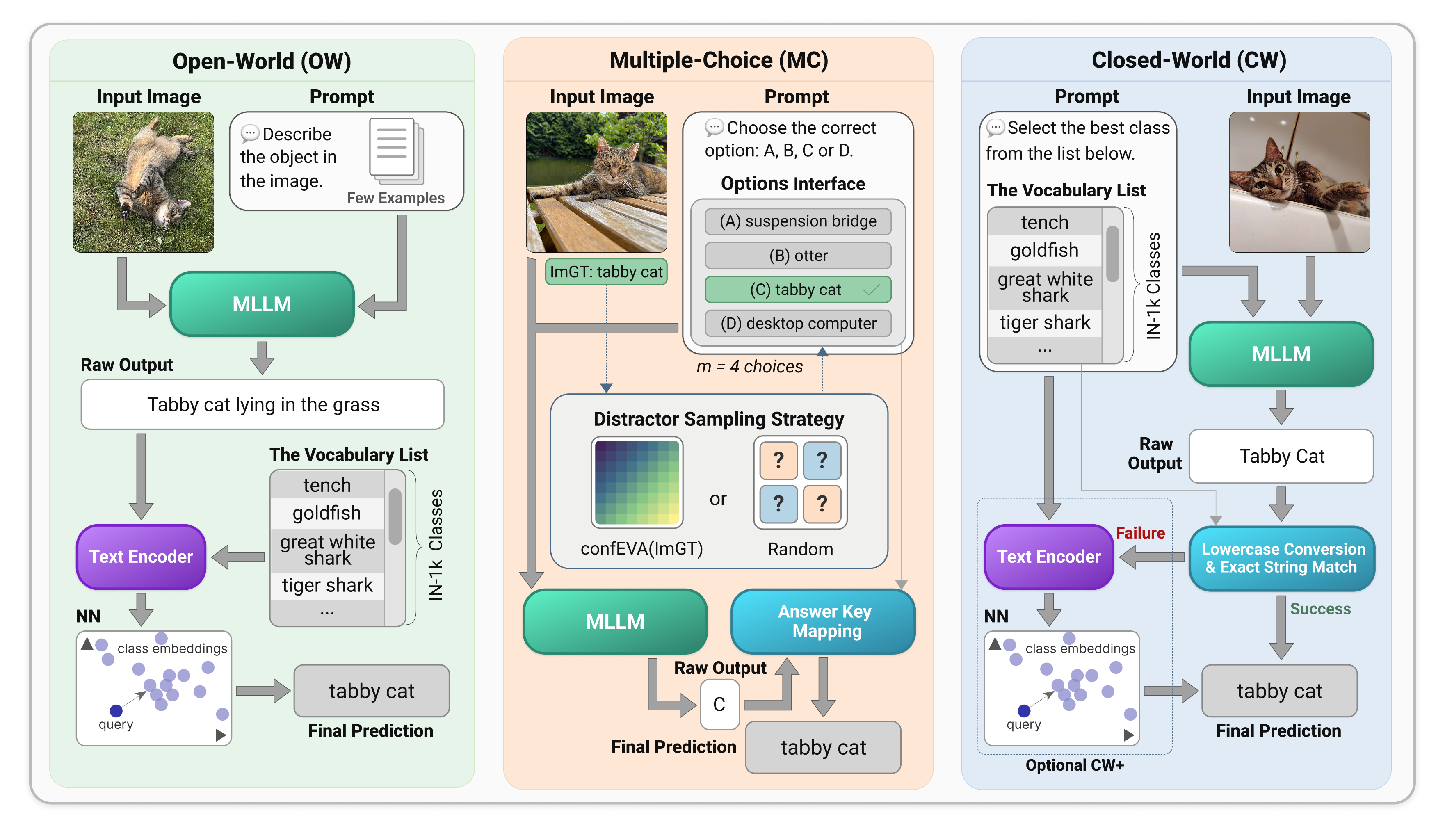}%
    }
 \caption{The three evaluated classification tasks — OW, MC, and CW(+) — described in \cref{subsec:tasks}
 }

\label{fig:tasks}
\end{figure*}

\acfp{mllm} \cite{openai_chatgpt_2025,openai2024gpt4ocard,google_gemini_2025,anthropic_claude3_2025, liu2023visualinstructiontuning, bai2023qwenvlversatilevisionlanguagemodel, chen2024internvlscalingvisionfoundation, deitke2024molmopixmoopenweights}, a recent extension of \acfp{llm} \cite{brown2020languagemodelsfewshotlearners, devlin2019bertpretrainingdeepbidirectional, 
raffel2023exploringlimitstransferlearning, Radford2019LanguageMA} into the visual domain, are capable of performing complex vision-language reasoning that integrates linguistic understanding with visual perception. \acp{mllm} performance has been extensively benchmarked \cite{fu2025mmecomprehensiveevaluationbenchmark, liu2024mmbenchmultimodalmodelallaround,li2023seedbenchbenchmarkingmultimodalllms, liu2024mibenchevaluatingmultimodallarge}, primarily assessing high-level multimodal reasoning and instruction following through multiple-choice question answering. 
%
% adoption
\acp{mllm} exhibit remarkable generalization and are increasingly used by both researchers
and the general public.  Use-cases range from safety-critical ones such as medical diagnostics \cite{caffagni2024revolutionmultimodallargelanguage, tu2023generalistbiomedicalai, li2023llavamedtraininglargelanguageandvision} and dangerous species classification \cite{du_wang_zhao_2024}, to helping with mundane office work \cite{Yin_2024, he2024webvoyagerbuildingendtoendweb}.

% MLLMs 
% 

% closed x open x self-sueprvised vlms x supervised

{\it Classification performance} provides a natural basis for comparing \acp{mllm} with established computer vision methods, both supervised and \acp{vlm} (CLIP \cite{radford2021learningtransferablevisualmodels}, SigLIP \cite{zhai2023sigmoidlosslanguageimage, tschannen2025siglip2multilingualvisionlanguage}). Recent works have begun to explore this comparison \cite{wu2024gpt4visgpt4zeroshotvisual, zhang2024visuallygroundedlanguagemodelsbad, liu2024revisitingmllmsindepthanalysis}. However, depending on the evaluation protocol and task formulation, studies report conflicting conclusions: while some find that \acp{mllm} perform substantially worse than classical vision models \cite{zhang2024visuallygroundedlanguagemodelsbad}, others suggest performance comparable to \acp{vlm} \cite{conti2025largemultimodalmodelsopenworld, liu2024revisitingmllmsindepthanalysis}.

%Moreover, none of these protocols places \acp{mllm} on equal footing with supervised 
%classifiers and \acp{vlm}, which predict directly over the full set of classes, making cross-paradigm comparisons fundamentally unfair. \todo{Think where to place this sentence}
%\todo{say that different protocols have problems and that we solve them, but we start with the protocols overview} 

\acp{mllm} produce free-form outputs, whereas image classification requires selecting a label from a predefined class set, so model outputs must be mapped to these classes.
In the literature, this mapping has been achieved by formulating \ac{mllm} classification as one of the following tasks -- Open-World (OW) \cite{conti2025largemultimodalmodelsopenworld, zhang2024visuallygroundedlanguagemodelsbad}, Multiple-Choice (MC) question answering \cite{liu2024revisitingmllmsindepthanalysis} and Closed-World (CW) \cite{wu2024gpt4visgpt4zeroshotvisual, zhang2024visuallygroundedlanguagemodelsbad}.  

Open-World (OW) is closest to everyday \ac{mllm} use: the model generates a free-form image description, which is then mapped to dataset classes, either through simple heuristics (\ie, text inclusion) \cite{zhang2024visuallygroundedlanguagemodelsbad, conti2025largemultimodalmodelsopenworld} or nearest-neighbor search in a text-embedding space \cite{conti2025largemultimodalmodelsopenworld}. Following \citeauthor{conti2025largemultimodalmodelsopenworld}, who found embedding-based mapping to be the most effective, we adopt this strategy and further show that OW outperforms CW for half of the evaluated models. This contradicts prior work \cite{zhang2024visuallygroundedlanguagemodelsbad}, which relied on string matching, indicating that previous OW limitations were due to the mapping method rather than the task itself.

%We adopt the embedding-based approach and, unlike prior work relying on string matching \cite{zhang2024visuallygroundedlanguagemodelsbad}, find that OW outperforms CW for half the evaluated models. We also observe that Sentence-BERT \cite{reimers2019sentencebertsentenceembeddingsusing} embeddings \cite{conti2025largemultimodalmodelsopenworld} perform worse than modern text encoders, suggesting that previously reported OW limitations were largely due to the mapping strategy rather than the task itself.

Multiple-choice (MC) question answering is commonly used to benchmark \acp{mllm} and is therefore often adopted for classification evaluation. The model selects from a set of candidate class names (up to 26 in prior work), including exactly one ground-truth label and several distractors. Distractors are sampled per-image, ranging from random to more challenging alternatives semantically close to the ground-truth label.
Prior work reported that harder distractors cause only a “slight decrease” in performance \cite{liu2024revisitingmllmsindepthanalysis}. We find this does not hold for newer models, where our confidence-interval-backed experiments show a larger drop under their sampling strategy. We also propose a distractor sampling method that further increases task difficulty. Yet, MC evaluation still inflates performance compared to other tasks, yielding an overly optimistic view of model ability.

The Closed-World (CW) task is designed to mimic classification with supervised models and \acp{vlm}. The model is prompted with candidate classes, ideally covering the full set of dataset classes. However, such a full-class formulation has been avoided due to input token limitations of older models \cite{zhang2024visuallygroundedlanguagemodelsbad, liu2024revisitingmllmsindepthanalysis}; in our work, we are the first to use the full set of classes. As the number of candidate classes increases, out-of-prompt (OOP) predictions become more frequent, where the model generates a label that is not contained in the provided candidate list despite explicit instructions to select from it. While OOP predictions are theoretically possible in MC, they are not observed in practice due to the small number of candidates. In the literature, CW is typically avoided in favor of MC \cite{liu2024revisitingmllmsindepthanalysis}; when it is used, OOP predictions are counted as incorrect \cite{zhang2024visuallygroundedlanguagemodelsbad}. We introduce CW+, which resolves OOP predictions by adopting the embedding-space nearest neighbour mapping from OW. With this approach, the main reason for avoiding CW is removed, making MC unnecessary unless the model is constrained by input token length.

Even the best evaluation protocol is only as good as its ground truth. %Regardless of the task used, the evaluation cannot be reliable if the underlying data are noisy or ambiguous, making the conclusions about the \acp{mllm} classification ability tightly coupled to the quality of the ground truth. 
ImageNet-1k has long
% long been considered the gold standard benchmark in computer vision,
served as a foundation for pretraining and evaluating model performance. 
Thanks to its
% long-term and
widespread adoption, its limitations are well-documented \cite{kisel2025flaws, beyer2020imagenet, tsipras2020imagenetimageclassificationcontextualizing, pmlr-v119-shankar20c, vasudevan2022doesdoughbagelanalyzing, northcutt2021pervasivelabelerrorstest, yun2021relabelingimagenetsinglemultilabels}. They go beyond simple image label errors ($\sim$20\% error rate in validation set \cite{northcutt2021pervasivelabelerrorstest}), including 15–21\% of validation images containing multiple objects from different classes \cite{beyer2020imagenet}, making single-label evaluation unreliable, overlapping class definitions \cite{vasudevan2022doesdoughbagelanalyzing}, distribution shifts between training and validation sets \cite{kisel2025flaws}, and duplicate images \cite{vasudevan2022doesdoughbagelanalyzing}.
% and motivating the “ReaL accuracy” metric.
%Others have focused on label noise, with Northcutt et al. \cite{northcutt2021pervasivelabelerrorstest} estimating a 20\% error rate in validation labels. 
Despite these limitations, ImageNet remains the de facto standard for evaluating visual recognition systems, including \acp{mllm} \cite{zhang2024visuallygroundedlanguagemodelsbad, liu2024revisitingmllmsindepthanalysis, wu2024gpt4visgpt4zeroshotvisual}. 

We reannotate a substantial portion of the ImageNet-1k validation set (625 classes),
mainly excluding challenging fine-grained animal categories that require extensive domain knowledge,
greatly reducing annotation noise, 
ambiguities and other imperfections mentioned. The new labels, \regt{}, are not yet public and thus unseen by any current model which enables controlled analysis of how supervised, self-supervised vision--language, and instruction-tuned \acp{mllm} respond to improved ground truth. 

The gap between top-performing \acp{mllm} and supervised models is almost halved on the new labels. \regt{} also partitions images into label categories based on agreement and multiplicity with \gt{}—the original ImageNet-1k ground truth. Although supervised models remain strongest on average, self-supervised \acp{vlm} and \acp{mllm} outperform them on images where \regt{} disagrees with \gt{}.

%We are the first to introduce a benchmark that not just unifies the most widely used evaluation protocols (CW, OW, and MC), but modifies them to address their flaws, rather than solely reevaluating models under the existing task formulations. In a CW setup, we use the full list of 1,000 classes, in contrast to the previously used maximum of 100 \cite{zhang2024visuallygroundedlanguagemodelsbad}, making the direct comparison to the supervised models and \acp{vlm} fair. We propose a new CW+ post-processing mapping of nearest neighbors in the text-embedding space to handle OOP predictions. We further strengthen the OW mapping procedure by utilizing text-embedding similarity akin to CW+. As a result, we show that, contrary to previous reports \cite{ zhang2024visuallygroundedlanguagemodelsbad},
%OW outperforms CW for multiple models. We also examine the impact of distractor selection strategies in MC and show that this task substantially inflates model accuracy, to a greater extent than previously reported \cite{liu2024revisitingmllmsindepthanalysis}. Finally, we find that model rankings are inconsistent across evaluation protocols, highlighting a broader measurement problem in multimodal evaluation.

In our work we benchmark five \acp{mllm} --- the closed-source \cgpt{} \cite{openai2024gpt4ocard} and the open-source \qwen{} \cite{yang2025qwen3technicalreport}, LLaVA-OneVision \cite{li2024llavaonevision}, InternVL3.5 \cite{chen2024internvlscalingvisionfoundation}, and PaliGemma~2 \cite{steiner2024paligemma2} on the reannotated data. We also conduct preliminary experiments to study prompt design, reproducibility, encoder selection for CW+ and OW, and the influence of batch size and image order.

%We find that OOPs are not random: they are far more common on multilabel images, challenging cases, or images that contain no ImageNet-1k object at all. 

%Beyond assessing accuracy, .  

In a focused case study on the subset where ChatGPT disagreed with our \regt{}, we conducted a second annotation pass in which annotators reviewed each image alongside anonymized predictions from \cgpt{}, SigLIP~2, \gt{}, and \regt{}. For roughly half of these disputed cases with single-label, annotators either fully replaced or augmented the labels with \cgpt{}’s predictions, revealing substantial residual label noise even after an initial careful reannotation. This demonstrates that \acp{mllm} can serve as powerful annotation assistants.

To assess performance on fine-grained biological categories excluded from reannotation, we additionally evaluate on the expert-curated weasel-family subset from \cite{kisel2025flaws}.

The main contributions are:
\begin{enumerate}
\item \textbf{Improved \ac{mllm} Image Classification Benchmark.} We systematically evaluate five \acp{mllm} on \mbox{ImageNet-1k} across Closed-World, Open-World, and Multiple-Choice setups within a single framework, enabling direct comparison with vision-language and supervised baselines. We introduce CW+, a lightweight embedding-based post-processing that resolves out-of-prompt predictions without costly constrained decoding, enabling full 1,000-class Closed-World evaluation. To support fair assessment, we provide a new multilabel reannotation (ReGT) of 625 classes that holistically addresses known ImageNet labeling issues.

\item \textbf{Label Noise Sensitivity Across Learning Paradigms.} Leveraging ReGT, we show that \acp{mllm} benefit most from corrected labels (up to +10.8\%), substantially narrowing the perceived performance gap with supervised models. This reveals that much of the reported \ac{mllm} weakness on classification is an artifact of noisy ground truth rather than genuine model deficiency, and that models less reliant on supervised training signals are more sensitive to annotation quality.

\item \textbf{Evaluation Protocol Sensitivity.} We quantify the impact of commonly overlooked design choices—distractor selection, batch size, image ordering, output format, and text encoder—showing that they substantially affect reported accuracy (\eg confusion-matrix distractors cause a 10--15\% drop over random ones), calling into question prior results obtained under inflated conditions.

\item \textbf{\acp{mllm} as Annotation Assistants.} In a controlled case study, human annotators confirmed or integrated the \ac{mllm} prediction in approximately 50\% of difficult cases, demonstrating that \acp{mllm} can effectively flag 
% residual 
annotation errors and assist large-scale dataset curation.

\end{enumerate}

\section{Evaluation setup}
\label{sec:setup}

% We first describe our new, detailed reannotation for 625 classes of the ImageNet-1k validation set, which was created to address numerous well-documented label issues in 
% \cref{subsec:dataset}. We then, in \cref{subsec:eval_metric}, propose an adapted accuracy metric designed to properly evaluate predictions against multi-label ground truth, accounting for images with no valid labels and for semantically equivalent classes. In \cref{subsec:tasks}, we detail the three distinct evaluation tasks: Closed World (CW), Multiple-Choice (MC), and Open World (OW). Finally, we specify the models, text encoders, and class name conventions used in our experiments in \cref{subsec:models_and_class_names}.

\subsection{Dataset and labels}
\label{subsec:dataset}
% This might nto be a section on its own, a short version could go to setup where we explain notation etc., and long detailed reannotation stuff to supplementary.

All experiments are conducted on the ImageNet-1k dataset.
% Its limitations are well documented \cite{kisel2025flaws, beyer2020imagenet, tsipras2020imagenetimageclassificationcontextualizing, pmlr-v119-shankar20c, vasudevan2022doesdoughbagelanalyzing, northcutt2021pervasivelabelerrorstest, yun2021relabelingimagenetsinglemultilabels}, and 
Several prior works reannotated portions or even the entire validation set \cite{beyer2020imagenet, tsipras2020imagenetimageclassificationcontextualizing, yun2021relabelingimagenetsinglemultilabels, northcutt2021pervasivelabelerrorstest}. However, each of these efforts addressed only a subset of the problems
% , without taking into account the full scope of issues present in the dataset
\cite{kisel2025flaws}.

We introduce reannotated labels (\regt{}) for 625 classes, holistically addressing known issues. We excluded fine-grained wildlife categories (except for dog breeds and species easily distinguishable without expert domain knowledge), as prior work \cite{kisel2025flaws, bugs_in_data} showed them to be extremely challenging to annotate. To mitigate selection bias, we additionally evaluate on expert-provided \textit{Mustelidae} reannotations \cite{kisel2025flaws} in the Supplementary.

%We introduce reannotated labels for 625 classes of ImageNet-1k, addressing all known issues identified to date. Previous studies \cite{kisel2025flaws, bugs_in_data} have shown that wildlife-related categories are particularly challenging for annotators without biological expertise. The selected subset excludes fine-grained wildlife categories — except for a few that can be easily distinguished without any prior domain knowledge simply by studying the Web, as well as dog breeds. Although this selection introduces bias, we mitigate it by also conducting experiments presented in the Supplementary using biologist-provided reannotations of classes from the \textit{Mustelidae} family presented in Kisel et al \cite{kisel2025flaws}.

Annotators, trained to recognize and avoid common issues, were asked to label all objects from ImageNet-1k classes (or indicate that none is present). 
% They underwent periodic performance evaluations, and received continuous feedback. 
To make it easier to remember all classes, the annotation tool displayed top-20 model predictions for each image. While potentially introducing bias, this prevented many
% the significant
label omissions found in ImGT. 
% Under our active supervision,
The annotators collaborated via group chat to refine class definitions and challenging cases.

\noindent\textbf{Label categories.} The categorization of images based on the number of labels in \regt{} and agreement with \gt{} is provided in \cref{tab:reannot_defs}.
% For each image $i$, let $gt_i$ be the ground-truth label, $L_i$ the set of reannotated labels, and $p_i$ the predicted label.  
% We define $A = N \cup S \cup M$, where:
% \[
% \begin{gathered}
% N = \{i: |L_i| = 0\},\\[2pt]
% S = \{i: |L_i| = 1\}, \quad
% M = \{i: |L_i| > 1\}.
% \end{gathered}
% \]

% Further, $S = S^+ \cup S^-$ with $S^+ = \{i: gt_i \in L_i\}$ and $S^- = \{i: gt_i \notin L_i\}$;  
% similarly, $M = M^+ \cup M^-$ with $M^+ = \{i: gt_i \in L_i\}$ and $M^- = \{i: gt_i \notin L_i\}$.

\begin{table}[t!]
\centering
\small
\setlength{\tabcolsep}{4pt}
\begin{tabular}{ll|ll}
\toprule
Cat. & Definition & Cat. & Definition \\
\midrule
$A$ & $N \cup S \cup M$ & $S^+$ & $S \cap \{ i : gt_i \in L_i \}$ \\
$N$ & $\{ i : |L_i| = 0 \}$ & $S^-$ & $S \cap \{ i : gt_i \notin L_i \}$ \\
$S$ & $\{ i : |L_i| = 1 \}$ & $M^+$ & $M \cap \{ i : gt_i \in L_i \}$ \\
$M$ & $\{ i : |L_i| > 1 \}$ & $M^-$ & $M \cap \{ i : gt_i \notin L_i \}$ \\
\bottomrule
\end{tabular}
\caption{
Definitions of ImageNet-1k label categories based on \regt{} and \gt{} differences.
For image $i$, $gt_i$ is \gt{} label, and $L_i$ the set of \regt{} labels. $A, S, M$ correspond to all, single-label, and multi-label respectively, +/- denotes agreement/disagreement of \regt{} and \gt{}.
}
\label{tab:reannot_defs}
\end{table}

\begin{table}[t!]
\centering
\setlength{\tabcolsep}{2pt}
\small

\begin{tabular}{lrrrrrrrr}
\toprule
 & \multicolumn{1}{c}{$\mathrm{A}$} & \multicolumn{1}{c}{$\mathrm{S}$} & \multicolumn{1}{c}{$\mathrm{{S+}}$} & \multicolumn{1}{c}{$\mathrm{{S-}}$} & \multicolumn{1}{c}{$\mathrm{M}$} & \multicolumn{1}{c}{$\mathrm{{M+}}$} & \multicolumn{1}{c}{$\mathrm{{M-}}$} & \multicolumn{1}{c}{$\mathrm{{N}}$} \\
 \midrule
\# & 31250 & 18071 & 16177 & 1894 & 11834 & 10756 & 1078 & 1345 \\
\% & 100 & 57.8 & 51.8 & 6.1 & 37.9 & 34.4 & 3.5 & 4.3 \\
\bottomrule
\end{tabular}

\caption{The number of images in each label category (\#) and their fraction among reannotated images (\%). Images with multiple labels ($\mathrm{M}$) account for roughly 40\,\% of image in the selected 625 classes.
Approximately 5\,\% of the images have no correct corresponding ImageNet-1k label ($\mathrm{N}$). }
\label{tab:set_size}
\end{table}

\begin{table*}[ht]
\footnotesize
\centering
\setlength{\tabcolsep}{3.5pt}
\begin{tabular}{lllclllllllc}
% \begin{tabular}{lrrrrrrrr}
\toprule

% \multicolumn{1}{r}{\# images} & $31250$ & $31250$ & $17532$ & $15862$ & $1670$ & $12373$ & $11071$ & $1302$ \\

& & \multicolumn{1}{c}{\gt{}} & \multicolumn{8}{c}{ReGT} & \multirow{2}{*}{Im $\cap$ Re}\\ 

\cmidrule{3-3} \cmidrule{5-11} 

Model & Task & \multicolumn{1}{c}{$\mathrm{A_{31250}}$} & & \multicolumn{1}{c}{$\mathrm{A_{31250}}$} & \multicolumn{1}{c}{$\mathrm{S_{18071}}$} & \multicolumn{1}{c}{$\mathrm{{S+}_{16177}}$} & \multicolumn{1}{c}{$\mathrm{{S-}_{1894}}$} & \multicolumn{1}{c}{$\mathrm{M_{11834}}$} & \multicolumn{1}{c}{$\mathrm{{M+}_{10756}}$} & \multicolumn{1}{c}{$\mathrm{{M-}_{1078}}$}  \\
%&Model & $\mathrm{A^{\gt{}}_{31250}}$ & &$\mathrm{A_{31250}}$ & $\mathrm{S_{17532}}$ & $\mathrm{{S+}_{15862}}$ & $\mathrm{{S-}_{1670}}$ & $\mathrm{M_{12373}}$ & $\mathrm{{M+}_{11071}}$ & $\mathrm{{M-}_{1302}}$ \\
\cmidrule{1-11}

PaliGemma2-mix-28B/448 & OW &
37.11 {\scriptsize(13)} & & 
47.94 {\scriptsize\textcolor{ForestGreen}{+10.8}, (13)} & 
39.74 {\scriptsize(13)} & 
41.49 {\scriptsize(13)} & 
24.76 {\scriptsize(13)} & 
54.55 {\scriptsize(13)} & 
56.24 {\scriptsize(13)} & 
37.66 {\scriptsize(5)} & 35.78\\

LLaVA-OneVision-72B-Chat & OW &
62.00 {\scriptsize(12)} & & 
70.58 {\scriptsize\textcolor{ForestGreen}{+8.6}, (12)} & 
67.11 {\scriptsize(12)} & 
70.67 {\scriptsize(12)} & 
36.69 {\scriptsize(4)} & 
72.52 {\scriptsize(12)} & 
75.87 {\scriptsize(12)} & 
39.05 {\scriptsize\textcolor{Brown}{(3)}} & 59.39 \\

InternVL3.5-38B & CW+ &
64.31 {\scriptsize(11)} & & 
72.57 {\scriptsize\textcolor{ForestGreen}{+8.3}, (11)} & 
67.69 {\scriptsize(11)} & 
71.05 {\scriptsize(11)} & 
39.02 {\scriptsize\textcolor{Dandelion}{(1)}} & 
76.91 {\scriptsize(11)} & 
80.43 {\scriptsize(11)} & 
41.74 {\scriptsize\textcolor{Dandelion}{(1)}} & 61.53 \\

Qwen3-VL-235B-A22B-Inst & OW &
68.74 {\scriptsize(10)} & & 
76.68 {\scriptsize\textcolor{ForestGreen}{+7.9}, (10)} & 
73.61 {\scriptsize(10)} & 
77.79 {\scriptsize(10)} & 
37.91 {\scriptsize\textcolor{Brown}{(3)}} & 
78.71 {\scriptsize(10)} & 
82.46 {\scriptsize(10)} & 
41.28 {\scriptsize\textcolor{gray}{(2)}} & 65.87 \\

GPT-4o-2024-08-06 & CW+ &
76.40 {\scriptsize(9)} & & 
82.36 {\scriptsize\textcolor{ForestGreen}{+6.0}, (9)} & 
81.11 {\scriptsize(9)} & 
86.12 {\scriptsize(9)} & 
38.28 {\scriptsize\textcolor{gray}{(2)}} & 
82.27 {\scriptsize(9)} & 
86.74 {\scriptsize(9)} & 
37.66 {\scriptsize(5)} & 72.79 \\

\midrule

DINOv3 ViT-L/16 (dino.txt) & CW &
80.96 {\scriptsize(8)} & & 
85.02 {\scriptsize\textcolor{ForestGreen}{+4.1}, (8)} & 
84.57 {\scriptsize(8)} & 
90.26 {\scriptsize(8)} & 
36.01 {\scriptsize(6)} & 
84.00 {\scriptsize(7)} & 
88.69 {\scriptsize(8)} & 
37.29 {\scriptsize(6)} & 76.79 \\   

SigLIP so400M/14-384 & CW &
82.40 {\scriptsize(7)} & & 
86.45 {\scriptsize\textcolor{ForestGreen}{+4.0}, (6)} & 
85.59 {\scriptsize(7)} & 
91.35 {\scriptsize(7)} & 
36.38 {\scriptsize(5)} & 
86.23 {\scriptsize(5)} & 
90.96 {\scriptsize(6)} & 
38.96 {\scriptsize(4)} & 78.27 \\

SigLIP~2 so400M/16-384 & CW &
83.30 {\scriptsize(6)} & & 
86.83 {\scriptsize\textcolor{ForestGreen}{+3.5}, (4)} & 86.06 {\scriptsize(4)} & 91.99 {\scriptsize(6)} & 35.32 {\scriptsize(7)} & 86.52 {\scriptsize(4)} & 91.47 {\scriptsize(5)} & 37.20 {\scriptsize(7)} & 78.93 \\

SigLIP~2 ViT/g-opt-384 & CW &
84.14 {\scriptsize(4)} & & 
87.12 {\scriptsize\textcolor{ForestGreen}{+3.0}, \textcolor{Brown}{(3)}} & 
86.40 {\scriptsize\textcolor{Brown}{(3)}} & 
92.55 {\scriptsize(4)} & 
33.90 {\scriptsize(8)} & 
86.75 {\scriptsize\textcolor{Brown}{(3)}} & 
91.88 {\scriptsize\textcolor{Brown}{(3)}} & 
35.53 {\scriptsize(8)} & 79.61 \\

\midrule

DINOv3 ViT-7B $k$=11 & CW &
84.18 {\scriptsize(5)} & & 
85.61 {\scriptsize\textcolor{ForestGreen}{+1.4}, (7)} & 
85.63 {\scriptsize(6)} & 
92.48 {\scriptsize(5)} & 
27.14 {\scriptsize(11)} & 
83.95 {\scriptsize(8)} & 
89.29 {\scriptsize(7)} & 
30.71 {\scriptsize(11)} & 79.07 \\

EfficientNetV2-XL & CW &
85.28 {\scriptsize\textcolor{Brown}{(3)}} & & 
86.53 {\scriptsize\textcolor{ForestGreen}{+1.3}, (5)} & 
85.96 {\scriptsize(5)} & 
92.67 {\scriptsize\textcolor{Brown}{(3)}} & 
28.56 {\scriptsize(9)} & 
85.88 {\scriptsize(6)} & 
91.74 {\scriptsize(4)} & 
27.37 {\scriptsize(12)} & 80.07 \\

EfficientNet-L2 & CW &
88.17 {\scriptsize\textcolor{gray}{(2)}} & & 
88.91 {\scriptsize\textcolor{ForestGreen}{+0.7}, \textcolor{gray}{(2)}} & 
87.94 {\scriptsize\textcolor{gray}{(2)}} & 
95.05 {\scriptsize\textcolor{gray}{(2)}} & 
27.24 {\scriptsize(10)} & 
89.12 {\scriptsize\textcolor{gray}{(2)}} & 
94.77 {\scriptsize\textcolor{gray}{(2)}} & 
32.75 {\scriptsize(9)} & 82.69 \\

EVA-02 ViT-L-14-448 & CW &
90.13 {\scriptsize\textcolor{Dandelion}{(1)}} & & 
90.04 {\scriptsize\textcolor{red}{-0.1}, \textcolor{Dandelion}{(1)}} & 
89.08 {\scriptsize\textcolor{Dandelion}{(1)}} & 
96.37 {\scriptsize\textcolor{Dandelion}{(1)}} & 
26.82 {\scriptsize(12)} & 
90.38 {\scriptsize\textcolor{Dandelion}{(1)}} & 
96.35 {\scriptsize\textcolor{Dandelion}{(1)}} & 
30.80 {\scriptsize(10)} & 84.37 \\
\bottomrule
\end{tabular}
\caption{Recognition accuracy on \datasub{} of ImageNet-1k defined in \cref{tab:reannot_defs}. The highest-performing task (OW/CW+) is selected for the \acp{mllm}. \textcolor{ForestGreen}{Increase} or \textcolor{red}{decrease} in classification accuracy on \regt{} w.r.t. ImGT. Model (rank)s are also shown. All models achieve higher accuracy on single-label images. The more dependence there is on the training data, the less the model benefits from the reannotation. The Im\ $\cap$ Re column reports the intersection of the two correctness sets: images on which the model is correct under \gt{} and images on which it is correct under \regt{}. This value indicates the overlap between these two correctness subsets, which are largely different.}
\label{tab:all_model_comp}
\end{table*}

% for \acp{mllm} (top), \acp{vlm} (middle) and supervised (bottom) models

\subsection{Evaluation metric}
\label{subsec:eval_metric}
For evaluation on \regt{}, we adapt the top-1 ReaL accuracy \cite{beyer2020imagenet}. In ReaL,
a prediction is counted as correct if it matches any of an image ground-truth labels, 
making it suitable for multilabel images.
We treat predictions on images in $N$ (those without any valid ground-truth label) as correct, 
since no meaningful label supervision exists for them.
We also account for semantically equivalent classes\footnote{Full list is in the Supplementary.} (\eg “notebook computer” vs.\ “laptop”), a known issue in ImageNet \cite{kisel2025flaws, vasudevan2022doesdoughbagelanalyzing}. We treat each pair in a predefined equivalence set $E$ as interchangeable, marking a prediction correct if it matches any label in the corresponding equivalence group. This applies to both \gt{} and \regt{}.
Formally, for the image $i$ we define a set of admissible labels:
\begin{equation}
L_i^* = L_i \;\cup\; \{ b \mid \exists a \in L_i,~\{a,b\} \in E \}.
\end{equation}
Then, the accuracy over a set of images $I$ is
\begin{equation}
\mathrm{Acc} = \frac{1}{|I|}\sum_{i \in I} \alpha_i, 
\quad
\alpha_i =
\begin{cases}
1, & p_i \in L_i^* \text{ or } |L_i| = 0,\\
0, & \text{otherwise,}
\end{cases}
\end{equation}
where $p_i$ is the prediction for image $i$.

\subsection{Tasks} 
\label{subsec:tasks}

% The \acp{mllm} are evaluated in three different setups.
% An overview, including prompts, is in Fig TODO.

\noindent\textbf{Task 1: Open-World (OW).}
The model is asked to classify the image without any predefined set of categories. A few prediction examples are given in the prompt to define the desired level of granularity.
The predicted output is then embedded via a text encoder and matched with the nearest embedded class name to give the final prediction.\footnote{Prompts for all Tasks are available in the Supplementary.}

\noindent\textbf{Task 2: Multiple-Choice (MC).}
The prompt specifies $n$ choices for each image and the model is asked to select one as the output. The choices always contain 1 correct answer and $n-1$ distractor answers. This task can be considered a special case of Task 1 with only a subset of the classes used, but the prompt is also different. We adopt the standard \cite{fu2025mmecomprehensiveevaluationbenchmark, liu2024mmbenchmultimodalmodelallaround,li2023seedbenchbenchmarkingmultimodalllms, liu2024mibenchevaluatingmultimodallarge} $n=4$ in all experiments.
We explore multiple variants, differing in the distractor sampling strategy. The 4 options always include either \gt{}, \regt{}, or both. The distractors can be selected at random, or based on the confusion matrix of the supervised EVA-02 model, denoted as \texttt{confEVA(c)}, to favour classes that are likely to be confused with the class \texttt{c}. It is ensured that the ImGT label distractors do not include any ReGT image labels. The evaluation is performed multiple times, each time varying the choices order and the random distractors.

\noindent\textbf{Task 3: Closed-World (CW).}
The prompt provides a list of all 1000 dataset classes and the model is asked to output a single one from the list. Since many objects may be present in the image, the model is asked to select only one class name that best describes the main object. Both the prediction and the class names are converted to lowercase; if they do not match, the prediction is considered incorrect.

%In the latter case, the confused classes for the class $c$ are identified based on their weighted frequency across EVA-02’s top-5 predictions for the class $c$ images. 

\begin{figure}[t!]
    \centering
    \resizebox{0.85\columnwidth}{!}{%
        \includegraphics{fig/model_deltas_plot.png}%
    }
 \caption{Classification accuracy on ImageNet-1k, change from
 the original (\gt{}) to reannotated labels (ReGT).
 }
 %   \caption{Classification accuracy change on ImageNet-1k reannotated labels (ReGT) compared to the original labels (\gt{}). }
    \label{fig:accuracy_change}
\end{figure}

%The more dependence there is on the training data, the less the model benefits from the reannotation.

\subsection{Model and class names overview}
\label{subsec:models_and_class_names}

\begin{figure*}[t]
    \centering
    \includegraphics[width=1.0\textwidth]{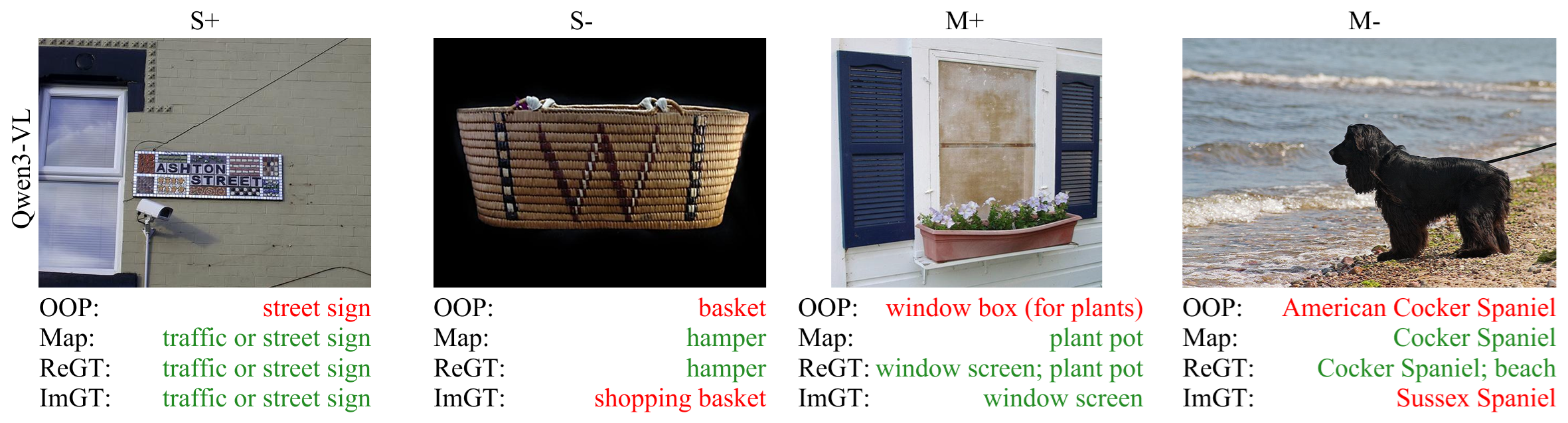}
    \caption{To address \acp{mllm} out-of-prompt ``OOP'' predictions, often referred to as \emph{hallucinations} \cite{liu2024revisitingmllmsindepthanalysis, zhang2024visuallygroundedlanguagemodelsbad}, we map model outputs that fall outside the provided class list to the nearest in-prompt class using the best model-specific encoder. Examples where the mapping is correct are shown. Columns correspond to the label categories from which the images were sampled. The row shows \qwen{} predictions, for which 38.75\% OOP predictions are correctly mapped. \textcolor{ForestGreen}{Correct} or \textcolor{red}{incorrect} image label.}
    \label{fig:chatgpt_mapping}
\end{figure*}

\textbf{Evaluated models.} We evaluate five \acp{mllm}: closed-source \cgpt{} \cite{openai2024gpt4ocard} and open-source \qwen{} \cite{yang2025qwen3technicalreport}, LLaVA-OneVision \cite{li2024llavaonevision}, InternVL3.5 \cite{chen2024internvlscalingvisionfoundation}, and PaliGemma~2 \cite{steiner2024paligemma2}. \cgpt{} was selected instead of the recent GPT-5 model due to ongoing API issues that make reliable evaluation difficult. \qwen{} was chosen as a representative leading open-source \ac{mllm}, while LLaVA-OneVision, InternVL3.5, and PaliGemma~2 cover a broader range of open-source architectures\footnote{See \cref{tab:model_architectures} in the Supplementary for details.}. 
For the CW experiment, we additionally evaluate two families of \acp{vlm}: SigLIP so400M/14-384 \cite{zhai2023sigmoidlosslanguageimage} and SigLIP~2 in two sizes (so400M/16-384 and ViT/g-opt-384) \cite{tschannen2025siglip2multilingualvisionlanguage}; DINOv3 as a zero-shot text-guided classifier `dino.txt` (ViT-L/16) and as a $k$-NN classifier using ImageNet-1k training set (ViT-7B, $k{=}11$); and the supervised classifiers EfficientNet-L2 (trained to be robust to label noise), EfficientNetV2-XL \cite{tan2021efficientnetv2smallermodelsfaster} (trained on ImageNet21k), and transformer-based EVA-02 (best-performing model on ImageNet-1k 
% according to the Timm leaderboard
\cite{rw2019timm}). 

\noindent\textbf{Text encoder.}
Text-embeddings (required for OW and CW+ are computed using the best model-specific text encoder in the main text. The standard mean over multiple prompt templates is used following \cite{radford2021learningtransferablevisualmodels}. Ablations for alternative encoders are provided in the Suppl., \cref{tab:lang_knn_extended}.

\begin{table*}[ht]
\small
\centering
\setlength{\tabcolsep}{4.5pt}
\begin{tabular}{lllllllllll}
\toprule
& & \multicolumn{1}{c}{ImGT} & & \multicolumn{7}{c}{ReGT} \\ \cmidrule{3-3} \cmidrule{5-11} 
& Task & \multicolumn{1}{c}{$\mathrm{A_{31250}}$} & & \multicolumn{1}{c}{$\mathrm{A_{31250}}$} & \multicolumn{1}{c}{$\mathrm{S_{18071}}$} & \multicolumn{1}{c}{$\mathrm{{S+}_{16177}}$} & \multicolumn{1}{c}{$\mathrm{{S-}_{1894}}$} & \multicolumn{1}{c}{$\mathrm{M_{11834}}$} & \multicolumn{1}{c}{$\mathrm{{M+}_{10756}}$} & \multicolumn{1}{c}{$\mathrm{{M-}_{1078}}$} \\
\cmidrule{1-11}
\multirow{1}{*}{PaliGemma~2} 
& OW & 
37.11 & &
47.94 & 39.74 & 41.49 & 24.76 & 54.55 & 56.24 & 37.66 \\ \midrule

\multirow{3}{*}{LLaVA-OV} 
& OW & 
62.00 {\scriptsize\textcolor{ForestGreen}{+18.6}} & &
70.58 {\scriptsize\textcolor{ForestGreen}{+17.8}} &
67.11 {\scriptsize\textcolor{ForestGreen}{+17.5}} &
70.67 {\scriptsize\textcolor{ForestGreen}{+18.7}} &
36.69 {\scriptsize\textcolor{ForestGreen}{+7.1}} &
72.52 {\scriptsize\textcolor{ForestGreen}{+20.3}} &
75.87 {\scriptsize\textcolor{ForestGreen}{+22.0}} &
39.05 {\scriptsize\textcolor{ForestGreen}{+4.1}} \\

& CW+ & 
52.66 {\scriptsize\textcolor{ForestGreen}{+9.3}} & &
62.82 {\scriptsize\textcolor{ForestGreen}{+10.1}} &
59.54 {\scriptsize\textcolor{ForestGreen}{+10.0}} &
62.32 {\scriptsize\textcolor{ForestGreen}{+10.4}} &
35.80 {\scriptsize\textcolor{ForestGreen}{+6.2}} &
63.61 {\scriptsize\textcolor{ForestGreen}{+11.4}} &
65.84 {\scriptsize\textcolor{ForestGreen}{+11.9}} &
41.37 {\scriptsize\textcolor{ForestGreen}{+6.4}}
 \\

& CW & 
43.40 & & 52.75 & 49.59 & 51.94 & 29.57 & 52.20 & 53.92 & 34.97 \\ \midrule

\multirow{3}{*}{InternVL3.5} 
& OW & 
59.23 {\scriptsize\textcolor{red}{-4.9}} & &
68.18 {\scriptsize\textcolor{red}{-4.2}} &
62.44 {\scriptsize\textcolor{red}{-5.1}} &
65.78 {\scriptsize\textcolor{red}{-5.1}} &
33.90 {\scriptsize\textcolor{red}{-5.1}} &
73.33 {\scriptsize\textcolor{red}{-3.5}} &
76.66 {\scriptsize\textcolor{red}{-3.7}} &
40.07 {\scriptsize\textcolor{red}{-1.4}} \\

& CW+ & 
64.31 {\scriptsize\textcolor{ForestGreen}{+0.2}} & &
72.57 {\scriptsize\textcolor{ForestGreen}{+0.2}} &
67.69 {\scriptsize\textcolor{ForestGreen}{+0.2}} &
71.05 {\scriptsize\textcolor{ForestGreen}{+0.2}} &
39.02 {\scriptsize\textcolor{ForestGreen}{+0.0}} &
76.91 {\scriptsize\textcolor{ForestGreen}{+0.1}} &
80.43 {\scriptsize\textcolor{ForestGreen}{+0.1}} &
41.74 {\scriptsize\textcolor{ForestGreen}{+0.3}} \\

& CW & 
64.14 & & 72.42 & 67.51 & 70.84 & 39.02 & 76.78 & 80.32 & 41.47 \\ \midrule

\multirow{3}{*}{\qwen{}} 
& OW & 
68.74 {\scriptsize\textcolor{ForestGreen}{+4.9}} & &
76.68 {\scriptsize\textcolor{ForestGreen}{+5.3}} &
73.61 {\scriptsize\textcolor{ForestGreen}{+3.5}} &
77.79 {\scriptsize\textcolor{ForestGreen}{+3.7}} &
37.91 {\scriptsize\textcolor{ForestGreen}{+1.9}} &
78.71 {\scriptsize\textcolor{ForestGreen}{+8.7}} &
82.46 {\scriptsize\textcolor{ForestGreen}{+8.9}} &
41.28 {\scriptsize\textcolor{ForestGreen}{+7.0}} \\

& CW+ & 
66.74 {\scriptsize\textcolor{ForestGreen}{+2.9}} & &
74.43 {\scriptsize\textcolor{ForestGreen}{+3.0}} &
72.90 {\scriptsize\textcolor{ForestGreen}{+2.8}} &
76.95 {\scriptsize\textcolor{ForestGreen}{+2.8}} &
38.23 {\scriptsize\textcolor{ForestGreen}{+2.2}} &
73.86 {\scriptsize\textcolor{ForestGreen}{+3.8}} &
77.53 {\scriptsize\textcolor{ForestGreen}{+3.9}} &
37.29 {\scriptsize\textcolor{ForestGreen}{+3.0}} \\

& CW & 
63.86 & & 71.39 & 70.15 & 74.14 & 36.06 & 70.04 & 73.61 & 34.32 \\ \midrule

\multirow{3}{*}{\cgpt{}}
& OW & 
70.92 {\scriptsize\textcolor{red}{-4.5}} & &
77.58 {\scriptsize\textcolor{red}{-3.7}} &
76.59 {\scriptsize\textcolor{red}{-3.7}} &
81.08 {\scriptsize\textcolor{red}{-4.3}} &
38.23 {\scriptsize\textcolor{ForestGreen}{+1.2}} &
76.55 {\scriptsize\textcolor{red}{-4.1}} &
80.75 {\scriptsize\textcolor{red}{-4.4}} &
34.69 {\scriptsize\textcolor{red}{-1.6}} \\

& CW+ &
76.40 {\scriptsize\textcolor{ForestGreen}{+1.0}} & &
82.36 {\scriptsize\textcolor{ForestGreen}{+1.1}} &
81.11 {\scriptsize\textcolor{ForestGreen}{+0.8}} &
86.12 {\scriptsize\textcolor{ForestGreen}{+0.7}} &
38.28 {\scriptsize\textcolor{ForestGreen}{+1.2}} &
82.27 {\scriptsize\textcolor{ForestGreen}{+1.6}} &
86.74 {\scriptsize\textcolor{ForestGreen}{+1.6}} &
37.66 {\scriptsize\textcolor{ForestGreen}{+1.4}} \\

& CW & 
75.40 & & 81.29 & 80.32 & 85.39 & 37.06 & 80.65 & 85.10 & 36.27 \\

\bottomrule
\end{tabular}
\caption{Recognition accuracy for CW(+) and OW tasks. PaliGemma~2 results are reported only for the OW setup, since it is not feasible to perform CW with all class names included in a single prompt. Changes relative to the Closed World (CW) baseline are indicated as \textcolor{ForestGreen}{increase} or \textcolor{red}{decrease}. The CW+ setup extends the CW task by applying the same text-embedding-space output mapping used in the Open World (OW) task, addressing OOP predictions in CW that would otherwise be considered incorrect. Results are obtained using the best-performing model-specific encoder; for a comparison of all encoders, see \cref{tab:lang_knn_extended}. As expected, accuracy in CW+ is higher than in standard CW. Notably, LLaVA-OV and \qwen{} outperform their CW results in the OW setup, an unexpected trend that contrasts prior findings on \acp{mllm}~\cite{liu2024revisitingmllmsindepthanalysis, zhang2024visuallygroundedlanguagemodelsbad}. In contrast, InternVL3.5 and \cgpt{} achieve the highest performance in CW+.}
\label{tab:ov_res}
\end{table*}

\begin{table}[t!]
\centering
\setlength{\tabcolsep}{3pt}
\small

\begin{tabular}{lrrrrrrrr}
\toprule
 Model & \multicolumn{1}{c}{$\mathrm{A}$} & \multicolumn{1}{c}{$\mathrm{S}$} & \multicolumn{1}{c}{$\mathrm{{S+}}$} & \multicolumn{1}{c}{$\mathrm{{S-}}$} & \multicolumn{1}{c}{$\mathrm{M}$} & \multicolumn{1}{c}{$\mathrm{{M+}}$} & \multicolumn{1}{c}{$\mathrm{{M-}}$} & \multicolumn{1}{c}{$\mathrm{{N}}$} \\
 \midrule
 InternVL3.5 & 0.9 & 0.8 & 0.8 & 0.8 & 0.8 & 0.7 & 1.2 & 2.5 \\
\cgpt{} & 5.3 & 4.0 & 3.4 & 8.7 & 6.0 & 5.6 & 9.9 & 16.4 \\
\qwen{} & 10.5 & 9.3 & 8.7 & 13.8 & 10.9 & 10.6 & 14.8 & 24.2 \\
LLaVA-OV & 26.8 & 24.1 & 23.6 & 28.1 & 29.0 & 28.8 & 31.2 & 42.5 \\
\bottomrule
\end{tabular}

\caption{\acp{mllm} out-of-prompt (OOP) rate, the (\%) of cases when the output is none of the classes specified, per \datasub{}. 
Models hallucinate more on 1. multilabel images than single label ones, 2. images that do not contain their ImGT in \regt{} (S- and M-), 3. images from N category (contain no ImageNet labels in \regt{}). On the last, the hallucination rate is highest for all models.
%For both models, hallucination rate is higher on multi-label images than on a single-label ones, and is highest on the N category (images that contain no ImageNet labels in ReGT). \qwen{} hallucinates roughly twice as often as \cgpt{}. 
}
\label{tab:ood}
\end{table}

\noindent\textbf{Class names.}
For CW and MC, we adopt the set of hand-crafted ImageNet-1k class names introduced in Kisel et al. \cite{kisel2025flaws}. Although we find that they contain imperfections, these are the most up-to-date version available for the dataset. It is possible to learn “optimal” class representations \cite{Zhou_2022, zhou2022conditionalpromptlearningvisionlanguage} and thereby increase accuracy, but this approach would no longer be zero-shot, which contradicts the goal of our \acp{mllm} classification evaluation.

\section{Results}
\label{sec:experiments}
% All experimentsd on the ImageNet-1k \cite{deng2009imagenet} Classification in three different setups.

\begin{table*}[ht]
\footnotesize
\setlength{\tabcolsep}{2pt}
\centering
\begin{tabular}{lllllllllll}
\toprule
% Model & Distractors & $\mathrm{Acc}_{gt}^{625}$ & $\mathrm{Acc}_A^{625}$ & $\mathrm{Acc}_S^{340}$ & $\mathrm{Acc}_{S+}^{309}$ & $\mathrm{Acc}_{S-}^{31}$ & $\mathrm{Acc}_M^{252}$ & $\mathrm{Acc}_{M+}^{228}$ & $\mathrm{Acc}_{M-}^{24}$ \\
& & \multicolumn{1}{c}{ImGT} & & \multicolumn{7}{c}{ReGT} \\ \cmidrule{3-3} \cmidrule{5-11} 
 & Distractors & $\mathrm{A_{625}}$ & & $\mathrm{A_{625}}$ & $\mathrm{S_{352}}$ & $\mathrm{{S+}_{316}}$ & $\mathrm{{S-}_{36}}$ & $\mathrm{M_{240}}$ & $\mathrm{{M+}_{221}}$ & $\mathrm{{M-}_{19}}$ \\
\midrule

% --- GPT-4o results first (multirow) ---
\multirow{6}{*}{\rotatebox[origin=c]{90}{\cgpt{}}} 
& \gt{} + random & 99.62 \textsuperscript{\scriptsize \textcolor{gray}{±0.07}} & & 90.95 \textsuperscript{\scriptsize \textcolor{gray}{±0.05}} & 89.49 \textsuperscript{\scriptsize \textcolor{gray}{±0.07}} & 99.67 \textsuperscript{\scriptsize \textcolor{gray}{±0.07}} & 0.09 \textsuperscript{\scriptsize \textcolor{gray}{±0.18}} & 91.85 \textsuperscript{\scriptsize \textcolor{gray}{±0.10}} & 99.75 \textsuperscript{\scriptsize \textcolor{gray}{±0.11}} & 0.00 \textsuperscript{\scriptsize \textcolor{gray}{±0.00}} \\
& \gt{} + \texttt{confEVA(\gt{})} & 90.66 \textsuperscript{\scriptsize \textcolor{gray}{±0.22}} & & 85.12 \textsuperscript{\scriptsize \textcolor{gray}{±0.21}} & 84.27 \textsuperscript{\scriptsize \textcolor{gray}{±0.26}} & 93.87 \textsuperscript{\scriptsize \textcolor{gray}{±0.29}} & 0.00 \textsuperscript{\scriptsize \textcolor{gray}{±0.00}} & 84.31 \textsuperscript{\scriptsize \textcolor{gray}{±0.39}} & 91.56 \textsuperscript{\scriptsize \textcolor{gray}{±0.42}} & 0.00 \textsuperscript{\scriptsize \textcolor{gray}{±0.00}} \\

& ReGT + \texttt{confEVA(ReGT)} & 51.35 \textsuperscript{\scriptsize \textcolor{gray}{±0.21}} & & 84.26 \textsuperscript{\scriptsize \textcolor{gray}{±0.37}} & 92.45 \textsuperscript{\scriptsize \textcolor{gray}{±0.36}} & 93.90 \textsuperscript{\scriptsize \textcolor{gray}{±0.31}} & 79.66 \textsuperscript{\scriptsize \textcolor{gray}{±1.50}} & 70.09 \textsuperscript{\scriptsize \textcolor{gray}{±0.89}} & 70.78 \textsuperscript{\scriptsize \textcolor{gray}{±0.83}} & 62.14 \textsuperscript{\scriptsize \textcolor{gray}{±3.36}} \\

& \gt{} + ReGT + random & 91.77 \textsuperscript{\scriptsize \textcolor{gray}{±0.26}} & & 94.89 \textsuperscript{\scriptsize \textcolor{gray}{±0.13}} & 94.79 \textsuperscript{\scriptsize \textcolor{gray}{±0.16}} & 99.67 \textsuperscript{\scriptsize \textcolor{gray}{±0.07}} & 51.97 \textsuperscript{\scriptsize \textcolor{gray}{±1.35}} & 94.31 \textsuperscript{\scriptsize \textcolor{gray}{±0.27}} & 99.80 \textsuperscript{\scriptsize \textcolor{gray}{±0.09}} & 30.56 \textsuperscript{\scriptsize \textcolor{gray}{±2.97}} \\

& \gt{} + ReGT + \texttt{confEVA(\gt{},ReGT)} & 88.31 \textsuperscript{\scriptsize \textcolor{gray}{±0.29}} & & 92.52 \textsuperscript{\scriptsize \textcolor{gray}{±0.15}} & 91.59 \textsuperscript{\scriptsize \textcolor{gray}{±0.20}} & 96.36 \textsuperscript{\scriptsize \textcolor{gray}{±0.15}} & 49.73 \textsuperscript{\scriptsize \textcolor{gray}{±1.27}} & 92.86 \textsuperscript{\scriptsize \textcolor{gray}{±0.31}} & 98.16 \textsuperscript{\scriptsize \textcolor{gray}{±0.20}} & 31.24 \textsuperscript{\scriptsize \textcolor{gray}{±3.30}}  \\

& \gt{} + 999 (Closed World (CW)) & 74.69 \textsuperscript{\scriptsize \textcolor{gray}{±0.19}} & & 81.32 \textsuperscript{\scriptsize \textcolor{gray}{±0.18}} & 79.87 \textsuperscript{\scriptsize \textcolor{gray}{±0.19}} & 84.49 \textsuperscript{\scriptsize \textcolor{gray}{±0.21}} & 39.25 \textsuperscript{\scriptsize \textcolor{gray}{±0.82}} & 80.89 \textsuperscript{\scriptsize \textcolor{gray}{±0.35}} & 84.72 \textsuperscript{\scriptsize \textcolor{gray}{±0.36}} & 36.33 \textsuperscript{\scriptsize \textcolor{gray}{±0.58}} \\

\bottomrule
\end{tabular}
\caption{4-way multiple-choice (MC) task results on a randomly sampled subset of 625 images (one per class) with 95\,\% confidence intervals (CI). The function \texttt{confEVA()} generates challenging distractors based on the confusion matrix of EVA-02. The Closed World result corresponds to 999 distractors. Only \cgpt{} results are shown here, as the trend of gradually declining accuracy with more challenging distractors is similar across models. Full multiple-choice task results for all evaluated \acp{mllm} are provided in \cref{tab:mc_ext_res}.}
\label{tab:mc_res}
\end{table*}

\begin{figure*}[ht]
    \centering
    \resizebox{1.0\textwidth}{!}{%
        \includegraphics{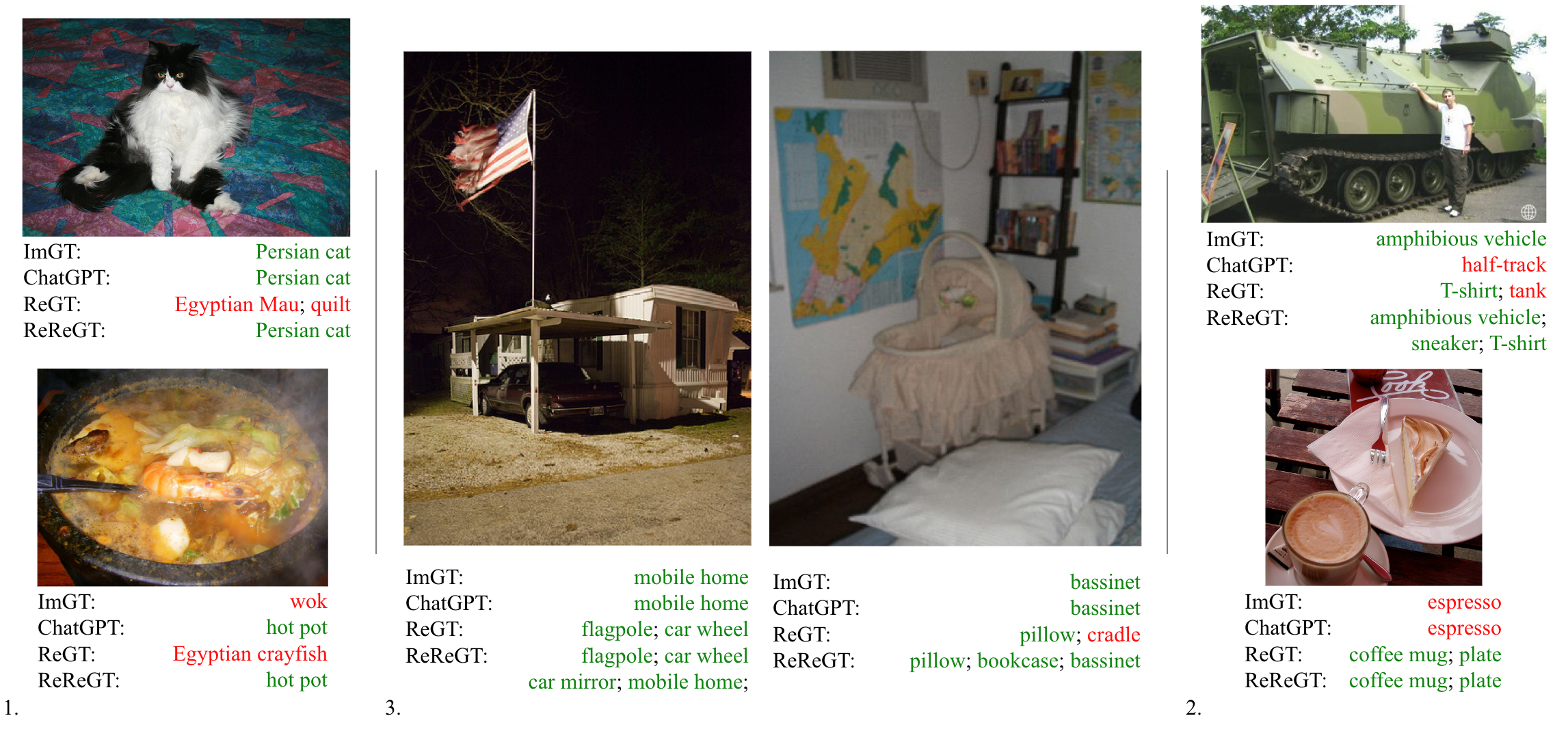}%
    }
    \caption{Images from the second annotation pass, where annotators 1. preferred \cgpt{} over \regt{} (left), 2. kept part of \regt{} without adding \cgpt{} (right), and 3. combined \regt{} with \cgpt{} predictions (middle). ``ReReGT'' indicates second pass reannotations. Annotators in the second pass preserved \textcolor{ForestGreen}{correct} \regt{} or added \cgpt{} predictions while also correcting \textcolor{red}{erroneous} first-pass labels.}
    \label{fig:case_study}
\end{figure*}

We compare \acp{mllm}, \acp{vlm}, and vision-only architectures on ImageNet-1k with ImGT and ReGT. The results reveal differences in robustness and label dependence, with \regt{} reducing the performance gap and an embedding-based mapping further mitigating hallucination errors.

\noindent\textbf{Preliminary experiments}
assessing the impact of batching (size, ordering and composition) and prompting 
% in the CW setup 
with either class names or class IDs 
% (see \cref{subsec:case_study}). 
% of these experiments 
are provided in the Supplementary.
% The best-performing parameters are selected and used throughout all the experiments.

\subsection{Closed-World}

\noindent\textbf{Results on ImGT} 
% of all models 
are reported in
\cref{tab:all_model_comp}, first column.  Among \acp{mllm}, PaliGemma~2 scores 37\,\%, while LLaVA-OV (62\,\%), InternVL3.5 (64\,\%) and \qwen{} (69\,\%) score closely. \cgpt{} dominates at 76\,\%. Despite this impressive result, there is still a gap between \cgpt{} and vision-specific self-supervised models, with the leading
% VLM
SigLIP~2 
% ViT/g-opt-384
reaching 84\,\%.
Supervised models still outperform all others, with EfficientNetV2-XL and EfficientNet-L2 scoring 85\,\% and 88\,\%, respectively, while EVA-02 achieves the highest accuracy of 90\,\%.

\noindent\textbf{Results on \regt{}} yield the largest gains for \acp{mllm}: PaliGemma~2 improves by 10.8\,\%, LLaVA-OV by 8.6\,\%, InternVL3.5 by 8.3\,\%, \qwen{} by 7.9\,\%, and \cgpt{} by 6.0\,\%. The performance of \acp{vlm} still shows a solid improvement of 3--4.1\,\%. Supervised models improve negligibly: EfficientNetV2-XL gains 1.3\,\%, EfficientNet-L2 improves by 0.7\,\% and \mbox{EVA-02} decreases by 0.1\,\%.
This confirms the intuition that the more a model relies on the training data, the smaller the reannotation benefit. The results also show \acp{vlm} and \acp{mllm} do not overfit the official ImageNet-1k validation labels, as reflected on the S- and M- subsets, where the top ranks go to self-supervised models rather than supervised ones. Deltas between ImGT and ReGT are presented in \cref{fig:accuracy_change}.

Notably, the gap between \acp{mllm} and supervised models got reduced significantly, for the better performing \mbox{\cgpt{}} by roughly 6\,\%. As shown in \cref{tab:set_size}, our non-OOD reannotated labels do not contain the ground-truth label in 9.6\,\% of images ($\mathrm{{S-}}$, $\mathrm{{M-}}$), which is close to the error rate of the best-performing supervised model EVA-02. 

%\noindent\textbf{Results on single-label and multi-label images.} All the models perform better on single-label images. The results on single-label and multilabel images also reveal that supervised models performance drops mainly on multilabel images, where its performance deteriorates compared to SigLIP~2 and \acp{mllm}. Similar trend can be observed for DINOv3.

\begin{figure*}[!htb]
    \centering
    \includegraphics[width=1\textwidth]{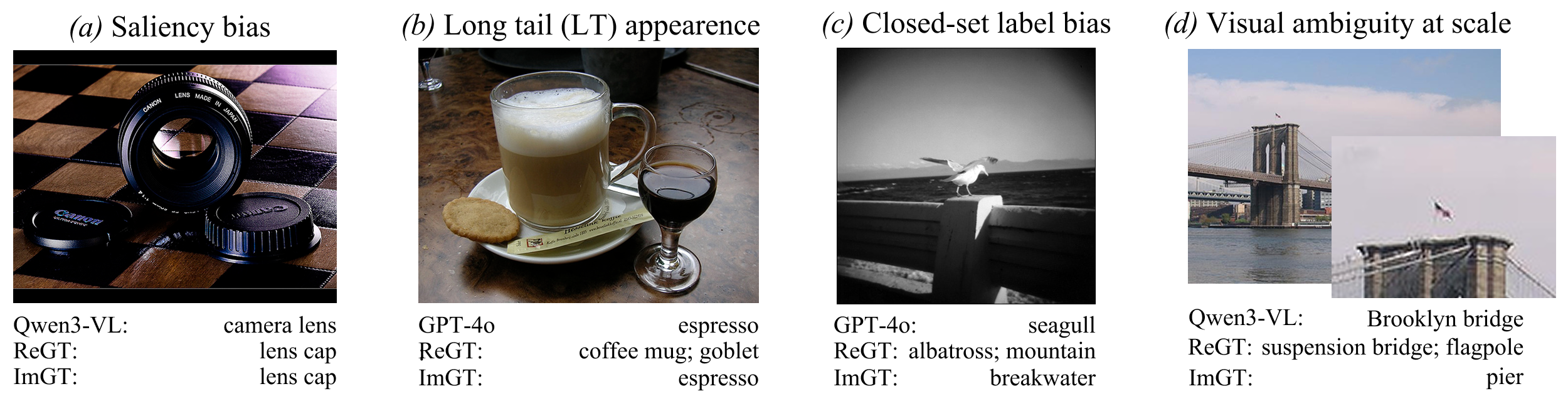}
%     \caption{Challenging classification cases.
% \textit{(a)}: The dominant object,
% which \acp{mllm} tend to predict,  does not always correspond to the intended class. Here, the dominant object, the lens, $\notin$ ImageNet-1k class list.
% \textit{(b)}: Appearance variations make it difficult to recognize long-tail content without additional context. An espresso served in a goblet for latte preparation differs from the typical presentation in a cup, confusing models and humans alike.
% \textit{(c)}: Fixed class lists push annotators/models to select a label, even when the image content falls outside the defined categories (``seagull'' $\notin$ ImageNet1-k). \textit{(d)}: Recognizability depends on scale and perceptual context.
% For smaller or visually ambiguous objects, determining what is seen versus inferred by contextual associations becomes increasingly challenging.}
\caption{Challenging classification cases.
\textit{(a)} Dominant objects do not always match the target class; e.g., a ``lens'' $\notin$ ImageNet-1k labels.
\textit{(b)} Appearance variations hinder recognition of long-tail content, e.g., espresso served in a goblet vs. a cup.
\textit{(c)} Fixed label sets force assignment even for out-of-distribution objects (e.g., ``seagull'' $\notin$ ImageNet-1k).
\textit{(d)} Recognizability depends on scale and context; small or ambiguous objects are harder to identify.}
\label{fig:discussion}
\end{figure*}

\begin{figure}[ht]
    \centering
    \resizebox{\columnwidth}{!}{
        \includegraphics{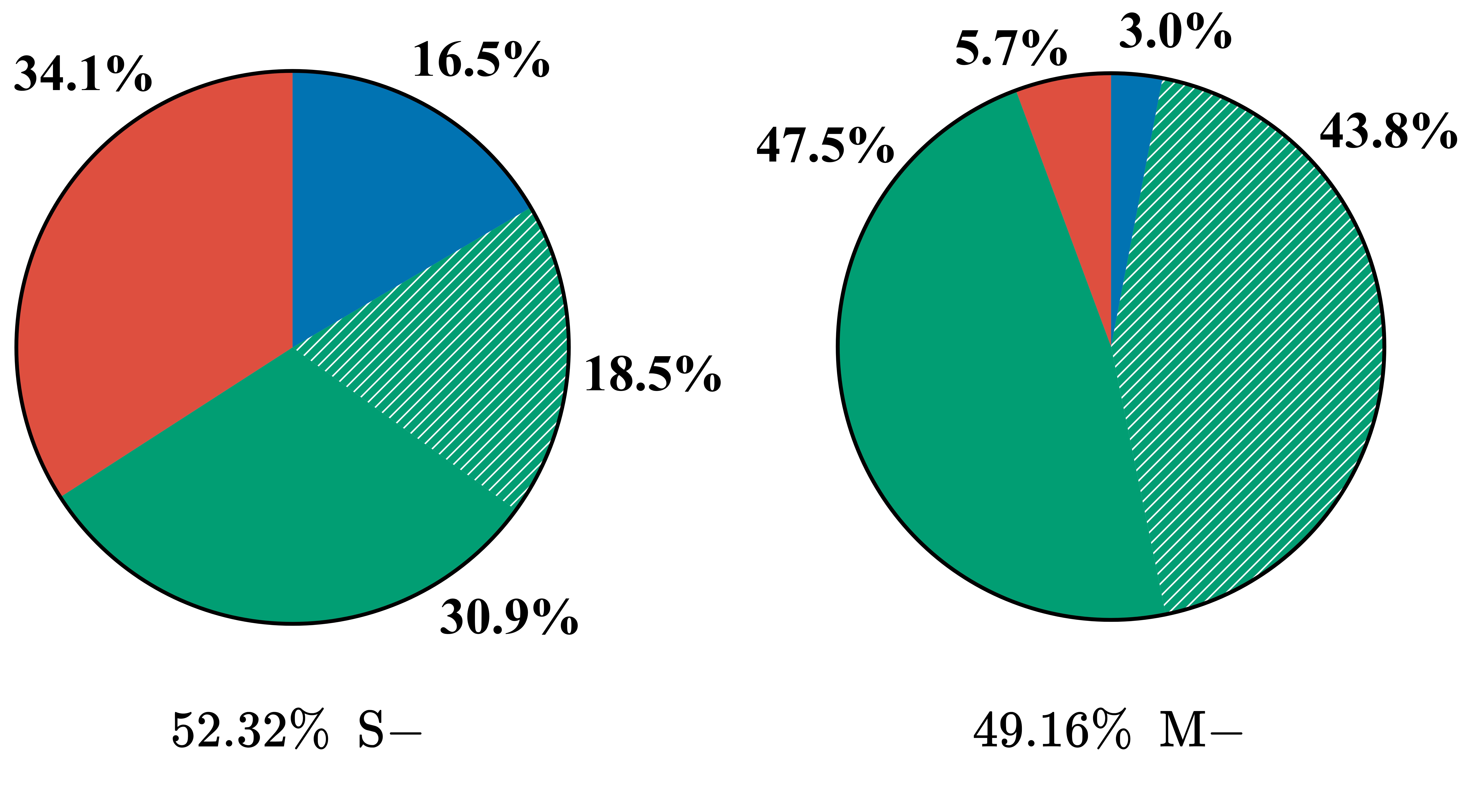}
    }
    \vspace{-2.5em}
    \caption{\hspace{-1ex} Second annotation pass results on images where \mbox{\cgpt{}} gave a valid prediction and disagreed with \regt{} for $\mathrm{S-}$ and $\mathrm{M-}$ label categories (52.32\% and 49.16\% of images in these categories, respectively; challenging since \gt{} $\notin$ \regt{}). Color coding: Annotators \textcolor[HTML]{DE4F3F}{\rule{1.2ex}{1.2ex}} preferred the \cgpt{} prediction (\regt{} was incorrect); \textcolor[HTML]{029E73}{\rule{1.2ex}{1.2ex}}\, preserved correct \regt{} without adding \cgpt{} prediction; {\rlap{\textcolor{white}{\rule{1.2ex}{1.2ex}}}\textcolor[HTML]{029E73}{\rule{0.25ex}{1.2ex}\kern0.25ex\rule{0.25ex}{1.2ex}\kern0.25ex\rule{0.25ex}{1.2ex}}}\, combined \cgpt{} prediction with \regt{} (\cgpt{} $\notin$ \regt{} before); \textcolor[HTML]{0173B2}{\rule{1.2ex}{1.2ex}}\, made other changes (\cgpt{} \& \regt{} $\notin$ ReReGT). %Percentages below pie charts show proportions of $\mathrm{S-}$ and $\mathrm{M-}$ images where valid \cgpt{} predictions differ from ReGT.
    }
    
    %\textcolor[HTML]{0173B2}{\rule{1.2ex}{1.2ex}}\, + \,\textcolor[HTML]{ECB01F}{\rule{1.2ex}{1.2ex}}\ ChatGPT was right and we had an error in reannottaions. \textcolor[HTML]{0173B2}{\rule{1.2ex}{1.2ex}}\, + \,\textcolor[HTML]{CC79A7}{\rule{1.2ex}{1.2ex}}\, + \,\textcolor[HTML]{DE4F3F}{\rule{1.2ex}{1.2ex}}\,  Reannotated labels were completely wrong (did not contain any correctly marked object)

\label{fig:rere_split}
\end{figure}

%For nearly half of the image, the annotaters have added the \cgpt{} prediction to the re-annotated labels (the human annotators have missed the \cgpt{} prediction). For the remaining portion, we confirm that the reannotations were at least partially accurate, whereas \cgpt{}’s predictions were inaccurate.
    % It can be seen that in the second pass, for nearly half of images, \cgpt{} provided the correct prediction while our reannotated labels were incorrect. For the remaining portion, we confirm that the reannotations were at least partially accurate, whereas \cgpt{}’s predictions were inaccurate.

\noindent\textbf{Out-Of-Prompt.}
\acp{mllm} still often predict texts outside of the provided prompt class names despite clear instructions not to, an effect known as hallucinations \cite{liu2024revisitingmllmsindepthanalysis, zhang2024visuallygroundedlanguagemodelsbad}, which we investigate by label category in
\cref{tab:ood}.

%The results reveal the hallucinations are far from random - we can observe that the model tends to hallucinate more on %1.  multilabel images than single label ones, 2. challenging images whose label was reannotated (S- and M-) 3. images which our annotators labelled as "no ImageNet-1k object", where the OOD rate more than doubles.

\noindent\textbf{NN in text embedding space mapping (CW+).} To address this, we encode model predictions into the text embedding space (similarly to the OW setup) and assign each to its nearest class name embedding (CW+, \cref{tab:ov_res}).
%To address the OOP issue, we encode model predictions into the text embedding space (similarly to the OW setup), so that the predicted outputs are semantically aligned with the provided class names. Each predicted text embedding is then assigned to its nearest class name embedding. The results, denoted as CW+, are shown in \cref{tab:ov_res}.
The performance of CW+ consistently outperforms pure CW across all subsets and models, with the largest gains on the more challenging S- and M- subsets. The accuracy improvement achieved by LLaVA-OV (the highest among all evaluated models) is $\approx$54$\times$ greater than that of InternVL3.5 (the model with the lowest improvement) when evaluated against \gt{}, and  $\approx$67$\times$ greater when evaluated using \regt{}. The difference is explained by LLaVA-OV’s consistently higher OOP rate (\cref{tab:ood}), which also corresponds to higher correct mapping rates\footnote{See \cref{tab:qwen_ood} in the Supplementary for a comprehensive breakdown.}. Examples of successful mappings are shown in \cref{fig:chatgpt_mapping}.

%\noindent\textbf{NN in text embedding space mapping (CW+).} To address this, we encode 
%model predictions into the text embedding space and assign each to its nearest class name 
%embedding (CW+, \cref{tab:ov_res}). CW+ consistently outperforms CW across all subsets, 
%with the largest gains on $\mathrm{S-}$ and $\mathrm{M-}$. The accuracy improvement of 
%LLaVA-OV is approximately 54$\times$ greater than that of InternVL3.5 against \gt{} 
%(67$\times$ with \regt{}), explained by LLaVA-OV's higher OOP rate (\cref{tab:ood}) and 
%correspondingly higher correct mapping rates\footnote{see \cref{tab:qwen_ood} in the 
%Supplementary for a comprehensive breakdown.}. Examples are shown in \cref{fig:chatgpt_mapping}.

\subsection{Multiple-Choice}
Results in \cref{tab:mc_res} reveal the commonly adopted setup with random distractors significantly inflates performance, while harder distractors reduce it by 10--15\,\%. When both \gt{} and \regt{} are provided, models select between them at similar rates on challenging single label images ($\mathrm{S-}$), where confidence intervals are also 
wider, as they are for $\mathrm{M-}$.

\subsection{Open-World}
% \acp{mllm}' performance in the free-form output setup is reported in 
\cref{tab:ov_res} indicates that overall performance is noticeably lower than in the CW setup for InternVL3.5 
% ($\approx 5\,\%$) 
and \cgpt{} 
% ($\approx 4\,\%$)
, whereas it is higher for LLaVA-OV 
% ($\approx 19\,\%$)
and \qwen{} 
% ($\approx 5\,\%$)
. The latter contrasts with previous studies \cite{liu2024revisitingmllmsindepthanalysis, zhang2024visuallygroundedlanguagemodelsbad}, which report that accuracy in the CW setup generally exceeds that of the OW for various \acp{mllm} - a trend we also observe for InternVL3.5 and \cgpt{}. 
Consistent with the CW results, all models show sizable improvement on \regt{} compared to \gt{}.

% Despite the general decline for \cgpt{}, OW achieves an improvement of over 1\,\% on the subset $\mathrm{S-}$, which is among the most challenging.

%The \acp{mllm} performance in the free-form output setup is reported in \cref{tab:ov_res}. The results reveal that the overall performance is significantly lower than in the CW setup (~4 \% lower on most subsets). Despite the general decline, OW achieves an accuracy improvement of over 1\,\% on a single subset $S-$, which is among the most challenging.

%Consistently with the CW results, the performance of both models significantly improves on the reannotated labels comapred to the original labels.

\subsection{Case study: ChatGPT vs. humans}
\label{subsec:case_study}

We conducted a second annotation pass on the challenging $\mathrm{S-}$ and $\mathrm{M-}$ label categories\footnote{See \cref{sec:case_study} in the Supplementary for full experimental setup.}. It revealed that \cgpt{} predictions were frequently adopted either fully or in combination with \regt{} (see \cref{fig:rere_split}). At the same time, a non-negligible fraction of cases highlighted inconsistencies in human corrections, suggesting that \cgpt{} can serve as a useful assistive tool for error detection and verification.

\noindent\textbf{Annotation challenges.}
\cref{fig:discussion} illustrates common failure cases in image classification and dataset annotation.
\section{Conclusions}
\label{sec:conclusion}
\label{sec:discussion}

We presented an evaluation of both closed- and open-source \acp{mllm} on ImageNet-1k, enabled by a new large-scale reannotation (\regt{}) that 
% substantially
reduces label noise and resolves long-standing ambiguities. The results showed \acp{mllm} perceived performance is the most affected by incorrect ground truth. GT corrections  narrowed the apparent performance gap w.r.t. supervised and self-supervised vision models, while also revealing strong \acp{mllm} sensitivity to task formulation and distractor selection.
The \regt{} labels further expose how model families differ in their robustness to mislabeled and multilabel images, and demonstrate that \acp{mllm} can meaningfully assist human annotators in a controlled curation pipeline.
Although closed-source systems retain an advantage under strict Closed-World prompting, this gap largely disappears in Open-World settings.

We showed both the promise and the pitfalls of current \acp{mllm} for visual recognition, underscoring the need for high-quality benchmarks and careful integration of model assistance in dataset construction. 

%often generate free-form class descriptions similar to how a human might describe an image. However, evaluating such outputs still requires relying on image labels in some form, and prior work shows that OW typically performs worse than CW for most models (with \qwen{} being a notable exception). We further examine this and other classification challenges in \cref{fig:discussion}.

\noindent\textbf{Acknowledgments.} Nikita Kisel and Illia Volkov were supported by EquiLibre Technologies, Klara Janouskova by the CTU student grant SGS. The work was also supported by the Technology Agency of the Czech Republic within the SIGMA Programme project No. TQ28000003 and by the
 EC Digital Europe Programme
project CEDMO 2.0 no. 101158609. The infrastructure of the OP VVV funded project CZ.02.1.01/0.0/0.0/16\_019/0000765 ``Research Center for Informatics''.

{
    \small
    \bibliographystyle{ieeenat_fullname}
    \bibliography{main}
}


\clearpage

\appendix
\setcounter{page}{1}

\maketitlesupplementary
\section{Related Work}
\label{ap:related}

\noindent\textbf{\ac{mllm} evaluation.}
\acp{mllm} benchmarks \cite{fu2025mmecomprehensiveevaluationbenchmark, liu2024mmbenchmultimodalmodelallaround, li2023seedbenchbenchmarkingmultimodalllms, liu2024mibenchevaluatingmultimodallarge} have largely converged on multiple-choice question answering as the dominant evaluation format, favored for its simplicity and unambiguous scoring. However, this format makes direct comparison with traditional vision models (\acp{vlm}, supervised classifiers) difficult, as it measures different capabilities. Some benchmarks \cite{li2023seedbenchbenchmarkingmultimodalllms, liu2024mibenchevaluatingmultimodallarge} design more challenging distractors, while others \cite{gaur2024detectdescribediscriminatemoving} argue that selecting from a list is fundamentally easier than generating an answer, promoting self-retrieval methods that test a model's ability to describe and distinguish images without provided options. Our work fills this gap by evaluating \acp{mllm} on standard image classification under conditions comparable to traditional vision models.

\noindent\textbf{\acp{mllm} in Image Classification.}
Zhang et al. \cite{zhang2024visuallygroundedlanguagemodelsbad} show that generative \acp{mllm} perform far below CLIP on ImageNet-1k, even in a Closed-World task, though limited to 100 classes due to token length constraints of older models. To address OOP predictions, they introduce Probabilistic Inference, which constrains generation to only tokens from the provided class list. While effective, this is computationally intensive: for 1,000 classes, inference is 1,000 times slower than direct generation. This contrasts with our CW+ approach, which resolves OOP via fast post-processing. The authors also evaluate an OW task, using string matching (\ie text inclusion) to map free-text predictions to class names. This mapping strategy is inefficient, which explains why they find OW to always perform worse than CW — a finding we do not replicate with our embedding-based mapping.

Liu et al. \cite{liu2024revisitingmllmsindepthanalysis} report that newer \acp{mllm} approach or surpass CLIP-like systems, though typically under MC conditions. They explore increasing the number of answer options up to 26 — the number of letters in the alphabet — and find that accuracy decreases as options grow. To increase difficulty, they construct harder distractors using the top-25 most semantically similar class names identified with BERT \cite{devlin-etal-2019-bert}, observing only a slight accuracy drop of 2.3 percentage points for Qwen2-VL on ImageNet-1k. Notably, their experiments are run once without confidence intervals. Our confidence-interval-backed experiments show a considerably larger accuracy drop under their sampling strategy, and a further increase with our harder distractor selection strategy. Similarly to \cite{zhang2024visuallygroundedlanguagemodelsbad}, the authors do not evaluate on the full class list due to OOP, a limitation we address with CW+.

Conti et al. \cite{conti2025largemultimodalmodelsopenworld} evaluate \acp{mllm} in an Open-World task across several metrics, including cosine similarity between model predictions and ground-truth labels in the Sentence-BERT \cite{reimers2019sentencebertsentenceembeddingsusing} embedding space, which is an inspiration for our embedding-space approach, but we utilize newer models. We additionally include experiments with their encoder in \cref{tab:lang_knn_extended}. However, they do not report results for ImageNet-1k and focus solely on \acp{mllm} and the OW task, whereas we also observe performance on other tasks, and compare it against \acp{vlm} and supervised models.

%Wu et al. \cite{wu2024gpt4visgpt4zeroshotvisual} evaluated GPT-4V on the full 1,000-class ImageNet-1k in a closed-world setting, prompting the model with all class names and asking for top-5 predictions. This is the only prior work to evaluate on the full class list, achieving performance comparable to CLIP ViT-B.

%\noindent\textbf{ImageNet-1k.}
%ImageNet has served as a foundational dataset for pretraining and evaluating model performance. Thanks to its widespread adoption, its limitations are well-documented \cite{kisel2025flaws, beyer2020imagenet, tsipras2020imagenetimageclassificationcontextualizing, pmlr-v119-shankar20c, vasudevan2022doesdoughbagelanalyzing, northcutt2021pervasivelabelerrorstest, yun2021relabelingimagenetsinglemultilabels} and go beyond simple annotation errors. Beyer et al. \cite{beyer2020imagenet} addressed the widespread issue of multi-label images — estimated at 15–21\% of the validation set — where a single ground-truth label is insufficient. Northcutt et al. \cite{northcutt2021pervasivelabelerrorstest} estimate a 20\% error rate in validation labels. Broader issues include overlapping class definitions \cite{vasudevan2022doesdoughbagelanalyzing}, distribution shifts between training and validation sets \cite{kisel2025flaws}, and duplicate images \cite{vasudevan2022doesdoughbagelanalyzing}.
\section{Case study: ChatGPT vs. humans}
\label{sec:case_study}
% Having identified the $\mathrm{M-}$ and $\mathrm{S-}$ label categories as the most challenging, 
We conducted a second annotation pass by randomly assigning all challenging $\mathrm{S-}$ (1894) and $\mathrm{M-}$ (1078) images to the annotators.  
For each image, they received the following in random, anonymized order:  
1. \cgpt{} prediction, 2. \regt{}, 3. \gt{}, and 4. SigLIP~2 prediction.  
\cgpt{} was prompted to return all ImageNet-1k labels present in an image as class IDs rather than full class names, reducing token usage and associated costs.  

% (e.g., for “170 – Irish Wolfhound,” 
% the model outputs 170) 
% to minimize output token usage, as the model was not limited to predicting a single object.
ChatGPT accuracy on $\mathrm{S-}$ and $\mathrm{M-}$ images was initially 34.37\,\% and 34.32\,\%, respectively, with respect to \regt{}. After the second pass, accuracy increased to 55.20\,\% and 55.19\,\%.  
An analysis of the results from this verification pass is presented in \cref{fig:rere_split}. For 52.6\,\% of $\mathrm{S-}$ and 49.5\,\% of $\mathrm{M-}$ images (\textcolor[HTML]{DE4F3F}{\rule{1.2ex}{1.2ex}}\, + {\rlap{\textcolor{white}{\rule{1.2ex}{1.2ex}}}\textcolor[HTML]{029E73}{\rule{0.25ex}{1.2ex}\kern0.25ex\rule{0.25ex}{1.2ex}\kern0.25ex\rule{0.25ex}{1.2ex}}}\ sections in \cref{fig:rere_split}), annotators either confirmed the ChatGPT prediction as the only correct label or combined it with \regt{} label.  

% This led to a substantial accuracy improvement described above. 
% However, 

There remains a set of 30.9\,\% $\mathrm{S-}$ and 47.5\,\% $\mathrm{M-}$ mispredicted images (\textcolor[HTML]{029E73}{\rule{1.2ex}{1.2ex}}\ sections in \cref{fig:rere_split}) where at least some of \regt{} labels were preserved and no ChatGPT predictions were added. Examples of these reannotations are presented in \cref{fig:case_study}.  
We conclude:  
1. Error corrections made by trained annotators are not entirely reliable, being completely incorrect (\textcolor[HTML]{DE4F3F}{\rule{1.2ex}{1.2ex}}\, + \textcolor[HTML]{0173B2}{\rule{1.2ex}{1.2ex}}\,) in 50.6\,\% of $\mathrm{S-}$ and 8.7\,\% of $\mathrm{M-}$ images; and  
2. \cgpt{} can serve as a valuable assistant for flagging such errors for additional verification.

\begin{table*}[t]
\centering
\small
\setlength{\tabcolsep}{3pt}
\begin{tabular}{llll}
\toprule
Model & Vision Backbone & Lang. Encoder & Training Strategy \\
\midrule
PaliGemma2-mix-28B/448 & SigLIP so400M & Gemma2 27B & Joint multimodal pretraining + finetuning \\

LLaVA-OneVision-72B-Chat & SigLIP so400M & Qwen2 72B & Projector-based alignment + supervised finetuning \\

InternVL3.5-38B & InternViT 6B & Qwen3 32B & Progressive multimodal scaling + instruction tuning \\

Qwen3-VL-235B-A22B-Inst & SigLIP~2 so400M & Qwen3 MoE & Large-scale multimodal pretraining + instruction tuning \\

GPT-4o-2024-08-06 & (undisclosed) & (undisclosed) & Proprietary multimodal pretraining + post-training alignment (RLHF) \\

\bottomrule
\end{tabular}
\caption{Vision backbones and language models for all five evaluated \acp{mllm}. GPT-4o-2024-08-06 is a closed-source model; architectural details are not publicly disclosed.}
\label{tab:model_architectures}
\end{table*}

% SigLIP-so400M denotes the shape-optimized Vision Transformer. InternViT-6B is a vanilla ViT with 6 billion parameters optimized for MLLM tasks.
% \input{tables/exps_suppl/task_decomp}

\section{Evalutation Setup - Details}
\label{ap:setup}
All \acp{mllm} evaluated in our experiments, along with their corresponding vision backbones and language encoders, are summarized in \cref{tab:model_architectures}.

\subsection{Prompt overview}
The exact prompts used for InternVL3.5, LLaVA-OV, \qwen{}, and \cgpt{} are provided in \cref{fig:cs_prompt} for the CW and CW+ tasks, and in \cref{fig:ov_prompt} for the OW task. In the MC setup, all models share the same prompt, as shown in \cref{fig:mc_prompt}, whereas the OW and CW prompts differ. A key distinction is that \cgpt{} prompts generally omit detailed instructions on output formatting, since the desired output structure is enforced directly via the API. For PaliGemma~2, the standard \texttt{describe en} prompt was employed in the OW setup. The CW task could not be performed due to the input prompt exceeding the token length limit.

% \subsection{Classification task decomposition}
% We provide a detailed breakdown of task components in \cref{tab:task_decomposition}, including the input prompt format and the output processing. A ``Notes'' column is also included to highlight the subtleties defined by each task.

\subsection{Equal classes}
We list all class pairs treated as equal for evaluation in the main paper, along with image examples and brief explanations, in \cref{fig:equal_classes_uniform_1} and \cref{fig:equal_classes_uniform_2}.

\begin{figure*}[ht]
    \centering
    
    \begin{subfigure}[b]{0.24\textwidth}
        \centering
        \includegraphics[height=0.12\textheight, keepaspectratio]{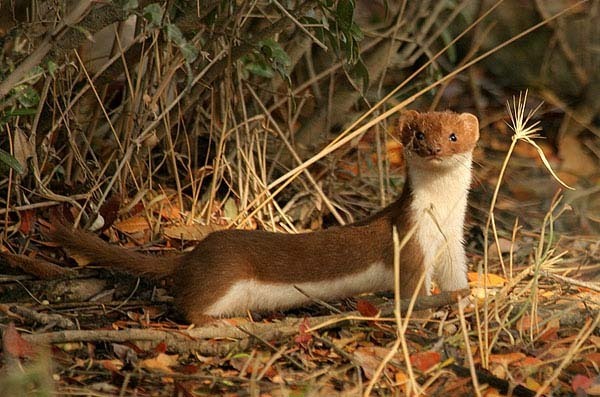}
        \caption{weasel}
    \end{subfigure}
    \hfill
    \begin{subfigure}[b]{0.24\textwidth}
        \centering
        \includegraphics[height=0.12\textheight, keepaspectratio]{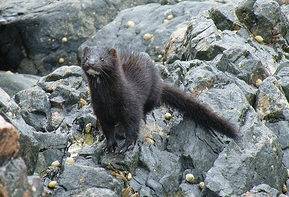}
        \caption{mink}
    \end{subfigure}
    \hfill
    \begin{subfigure}[b]{0.24\textwidth}
        \centering
        \includegraphics[height=0.12\textheight, keepaspectratio]{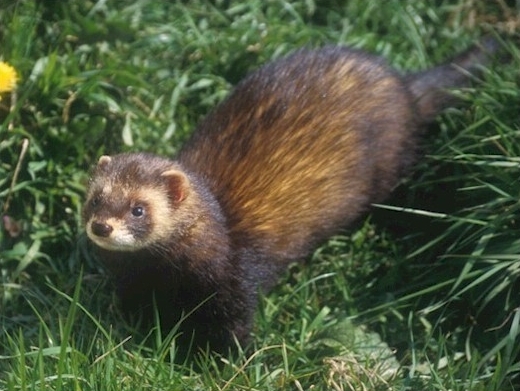}
        \caption{polecat}
    \end{subfigure}
    \hfill
    \begin{subfigure}[b]{0.24\textwidth}
        \centering
        \includegraphics[height=0.12\textheight, keepaspectratio]{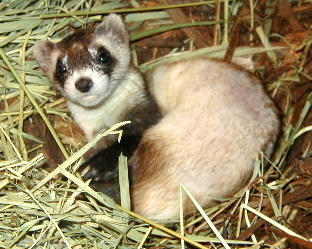}
        \caption{black-footed ferret}
    \end{subfigure}

    \caption{
    % Weasel family case study: Class overview. Correctly labeled ImageNet-1k images from the “weasel”, “mink”, “polecat”, and “black-footed ferret” classes are shown.
    % (a) The term \textbf{`weasel'} is often used to describe the whole \textit{mustelidae} family, which encompasses all the four species, but the images correspond to the least weasel species and similar. (b) \textbf{Minks} are easier to recognize, but still easily confused with the others for non-experts. (c) The European \textbf{polecat} is the ancestor of domestic ferret and they are extremely hard to recognize and they can also interbreed. Often, they are recognized by the environment (domestic ferrets are typically found in a man-made environment, polecats in the wild), as opposed by the animals' features. For this reason, the images are a mix. (d) The \textbf{black-footed ferret} is a rare ferret species and most images are actually the domesticated ferret, easily confused with polecats, as explained previously. The class got renamed to `domestic ferret'.
    % For more details on the reannotation and the issues, please refer to Kisel et. al. \cite{kisel2025flaws}.
    Weasel family case study: Correctly labeled ImageNet-1k images for “weasel”, “mink”, “polecat”, and “black-footed ferret”. (a) \textbf{Weasel} often refers broadly to \textit{mustelidae} family, but images mainly depict least weasels. (b) \textbf{Minks} are easier to recognize but still confused with similar species. (c) European \textbf{polecat} is the ancestor of domestic ferret and they are extremely hard to classify. Often, they are recognized by the environment, as opposed by the animals' features. (d) \textbf{Black-footed ferret} images are frequently domestic ferrets, leading to renaming. See \cite{kisel2025flaws} for details.
    }
    \label{fig:case_study_examples}
\end{figure*}

\begin{table*}[t!]
\centering
\setlength{\tabcolsep}{5pt}
\small

\begin{tabular}{llllllllll}
\toprule
& & \multicolumn{2}{c}{\gt{}} & & \multicolumn{1}{c}{\regt{}} & & \multicolumn{3}{c}{WeaselGT} \\ 
\cmidrule{3-4} \cmidrule{6-6} \cmidrule{8-10} 

Model & Task & 
\multicolumn{1}{c}{$\mathrm{A_{31250}}$} & $\mathrm{A_{159}}$\twemoji{otter}\twemoji{badger} & & 
% $\mathrm{A_{31250}^{625}}$ & $\mathrm{A_{159}^4}$ & & 
\multicolumn{1}{c}{$\mathrm{A_{31250}}$} & & $\mathrm{A_{159}}$\twemoji{otter}\twemoji{badger} & $\mathrm{{S+_{112}}}$ & $\mathrm{{S-_{47}}}$ \\
 \midrule
 % \multirow{2}{*}{(A)}
\gt{}
& CW & 100.00 & 100.00 & & 91.20 {\scriptsize\textcolor{red}{-8.8}} & & 70.44 {\scriptsize\textcolor{red}{-29.6}} & 100.00 & 00.00 \\
\midrule

\multirow{2}{*}{LLaVA-OV}
& CW & 42.56 & 25.79 & & 52.00 {\scriptsize\textcolor{ForestGreen}{+9.4}} & & 28.93 {\scriptsize\textcolor{ForestGreen}{+3.1}} & 35.71 & 12.77 \\ \vspace{0.4em}
& MC & 82.17 & 45.91 & & 90.01 {\scriptsize\textcolor{ForestGreen}{+7.8}} & & 61.01 {\scriptsize\textcolor{ForestGreen}{+15.1}} & 59.82 & 63.83 \\ 
\multirow{2}{*}{Intern-VL3.5}
& CW & 63.68 & 44.65 & & 72.96 {\scriptsize\textcolor{ForestGreen}{+9.3}} & & 61.01 {\scriptsize\textcolor{ForestGreen}{+16.4}} & 60.71 & 61.70 \\ \vspace{0.4em}
& MC & 85.89 & 55.97 & & 90.53 {\scriptsize\textcolor{ForestGreen}{+4.6}} & & 75.47 {\scriptsize\textcolor{ForestGreen}{+19.5}} & 75.89 & 74.47 \\

\multirow{2}{*}{\qwen{}}
& CW & 64.16 & 40.25 & & 72.16 {\scriptsize\textcolor{ForestGreen}{+8.0}} & & 50.94 {\scriptsize\textcolor{ForestGreen}{+10.7}} & 54.46 & 42.55 \\ \vspace{0.4em}
& MC & 86.32 & 64.15 & & 90.77 {\scriptsize\textcolor{ForestGreen}{+4.5}} & & 83.65 {\scriptsize\textcolor{ForestGreen}{+19.5}} & 86.61 & 76.60 \\

\multirow{2}{*}{\cgpt{}}
& CW & 74.69 & 64.15 & & 81.32 {\scriptsize\textcolor{ForestGreen}{+6.6}} & & 84.28 {\scriptsize\textcolor{ForestGreen}{+20.1}} & 88.39 & 74.47 \\
& MC & 88.31 & 67.92 & & 92.52 {\scriptsize\textcolor{ForestGreen}{+4.2}} & & 91.19 {\scriptsize\textcolor{ForestGreen}{+23.3}} & 92.86 & 87.23 \\
\midrule

DINOv3 (dino.txt)
& CW & 79.36 & 69.18 & & 85.12 {\scriptsize\textcolor{ForestGreen}{+5.8}} & & 86.79 {\scriptsize\textcolor{ForestGreen}{+17.6}} & 91.96 & 74.47 \\

SigLIP so400M
& CW & 82.88 & 66.04 & & 87.20 {\scriptsize\textcolor{ForestGreen}{+4.3}} & & 88.05 {\scriptsize\textcolor{ForestGreen}{+22.0}} & 91.07 & 80.85 \\

SigLIP~2 so400M
& CW & 83.36 & 66.67 & & 87.68 {\scriptsize\textcolor{ForestGreen}{+4.3}} & & 88.68 {\scriptsize\textcolor{ForestGreen}{+22.0}} & 90.18 & 85.11 \\

SigLIP~2 giant 
& CW & 83.84 & 67.92 & & 86.24 {\scriptsize\textcolor{ForestGreen}{+2.4}} & & 91.82 {\scriptsize\textcolor{ForestGreen}{+23.9}} & 92.86 & 89.36 \\
% & CW & 84.14 & 85.83 & & 87.12 {\scriptsize\textcolor{ForestGreen}{+3.0}} & & 84.91 {\scriptsize\textcolor{red}{-0.9}} & 90.18 & 72.34 \\

\midrule

DINOv3 ($k$-NN)
& CW & 83.36 & 71.07 & & 87.20 {\scriptsize\textcolor{ForestGreen}{+3.8}} & & 79.87 {\scriptsize\textcolor{ForestGreen}{+8.8}} & 87.50 & 61.70 \\

EfficientNetV2-XL
& CW & 85.12 & 71.07 & & 86.88 {\scriptsize\textcolor{ForestGreen}{+1.8}} & & 78.62 {\scriptsize\textcolor{ForestGreen}{+7.6}} & 87.50 & 57.45 \\

EfficientNet-L2
& CW & 87.52 & 75.47 & & 89.12 {\scriptsize\textcolor{ForestGreen}{+1.6}} & & 80.50 {\scriptsize\textcolor{ForestGreen}{+5.0}} & 91.07 & 55.32 \\

EVA-02
& CW & 90.24 & 73.58 & & 91.68 {\scriptsize\textcolor{ForestGreen}{+1.4}} & & 80.50 {\scriptsize\textcolor{ForestGreen}{+6.9}} & 90.18 & 57.45 \\

\bottomrule
\end{tabular}
\caption{Accuracy on the full 31250-image \regt{} set and on the WeaselGT 159-image subset \twemoji{otter}\twemoji{badger}. \textcolor{ForestGreen}{Increase} or \textcolor{red}{decrease} indicates the accuracy change under \regt{} vs. \gt{}. For WeaselGT, deltas denote the accuracy difference between WeaselGT vs. \gt{} on the same 159-image subset \twemoji{otter}\twemoji{badger}. 
% Overall, the consistently positive gains on WeaselGT largest for \acp{vlm} and \acp{mllm} and comparatively modest for supervised models suggest that multimodal and vision-language models are more sensitive to finegrained semantic corrections introduced by the reannotations, whereas supervised models remain more anchored to the original label space.}
Gains are consistently higher on WeaselGT, especially for \acp{vlm} and \acp{mllm}—indicating greater sensitivity to fine-grained label corrections than supervised models.}
\label{tab:mink}
\end{table*}

% Positive deltas reflect improved alignment with the reannotations, while negative deltas indicate reduced agreement. 

\begin{table*}[t!]
\centering
\setlength{\tabcolsep}{2.5pt}
\small

\begin{tabular}{l *{4}{lll}}
\toprule

& 
\multicolumn{3}{c}{weasel} &
\multicolumn{3}{c}{mink} &
\multicolumn{3}{c}{polecat} &
\multicolumn{3}{c}{domestic ferret} \\

\cmidrule(lr){2-4}
\cmidrule(lr){5-7}
\cmidrule(lr){8-10}
\cmidrule(lr){11-13}

& \gt{} & & WeaselGT
& \gt{} & & WeaselGT
& \gt{} & & WeaselGT
& \gt{} & & WeaselGT \\

\cmidrule(lr){2-2}
\cmidrule(lr){4-4}
\cmidrule(lr){5-5}
\cmidrule(lr){7-7}
\cmidrule(lr){8-8}
\cmidrule(lr){10-10}
\cmidrule(lr){11-11}
\cmidrule(lr){13-13}

\multirow{1}{*}{\gt{}} 
& 100.00 & & 84.62 {\scriptsize\textcolor{red}{-15.4}}
& 100.00 & & 85.00 {\scriptsize\textcolor{red}{-15.0}}
& 100.00 & & 57.14 {\scriptsize\textcolor{red}{-42.9}}
& 100.00 & & 61.54 {\scriptsize\textcolor{red}{-38.5}} \\

\midrule
\multirow{1}{*}{LLaVA-OV}
& 40.00 & & 76.92 {\scriptsize\textcolor{ForestGreen}{+36.9}}
& 60.00 & & 65.00 {\scriptsize\textcolor{ForestGreen}{+5.0}}
& 0.00 & & 0.00 {\scriptsize\textcolor{ForestGreen}{+0.0}}
& 0.00 & & 0.00 {\scriptsize\textcolor{ForestGreen}{+0.0}} \\

\multirow{1}{*}{Intern-VL3.5}
& 30.00 & & 65.38 {\scriptsize\textcolor{ForestGreen}{+35.4}}
& 32.00 & & 37.50 {\scriptsize\textcolor{ForestGreen}{+5.5}}
& 0.00 & & 0.00 {\scriptsize\textcolor{ForestGreen}{+0.0}}
& 92.00 & & 100.00 {\scriptsize\textcolor{ForestGreen}{+8.0}} \\

\multirow{1}{*}{\qwen{}} 
& 36.00 & & 76.92 {\scriptsize\textcolor{ForestGreen}{+40.9}}
& 72.00 & & 80.00 {\scriptsize\textcolor{ForestGreen}{+8.0}}
& 0.00 & & 3.57 {\scriptsize\textcolor{ForestGreen}{+3.6}}
& 42.00 & & 43.08 {\scriptsize\textcolor{ForestGreen}{+1.1}} \\

\multirow{1}{*}{\cgpt{}} 
& 46.00 & & 100.00 {\scriptsize\textcolor{ForestGreen}{+54.0}}
& 94.00 & & 95.00 {\scriptsize\textcolor{ForestGreen}{+1.0}}
& 10.00 & & 17.86 {\scriptsize\textcolor{ForestGreen}{+7.9}}
& 92.00 & & 100.00 {\scriptsize\textcolor{ForestGreen}{+8.0}} \\

\midrule

DINOv3 (dino.txt) 
& 44.00 & & 92.31 {\scriptsize\textcolor{ForestGreen}{+48.3}}
& 92.00 & & 100.00 {\scriptsize\textcolor{ForestGreen}{+8.0}}
& 36.00 & & 64.29 {\scriptsize\textcolor{ForestGreen}{+28.3}}
& 84.00 & & 86.15 {\scriptsize\textcolor{ForestGreen}{+2.2}} \\

SigLIP so400M 
& 42.00 & & 88.46 {\scriptsize\textcolor{ForestGreen}{+46.5}}
& 94.00 & & 97.50 {\scriptsize\textcolor{ForestGreen}{+3.5}}
& 26.00 & & 57.14 {\scriptsize\textcolor{ForestGreen}{+31.1}}
& 86.00 & & 95.38 {\scriptsize\textcolor{ForestGreen}{+9.4}} \\

SigLIP~2 so400M 
& 44.00 & & 92.31 {\scriptsize\textcolor{ForestGreen}{+48.3}}
& 92.00 & & 95.00 {\scriptsize\textcolor{ForestGreen}{+3.0}}
& 30.00 & & 67.86 {\scriptsize\textcolor{ForestGreen}{+37.9}}
& 86.00 & & 92.31 {\scriptsize\textcolor{ForestGreen}{+6.3}} \\

SigLIP~2 giant 
& 44.00 & & 100.00 {\scriptsize\textcolor{ForestGreen}{+56.0}}
& 94.00 & & 97.50 {\scriptsize\textcolor{ForestGreen}{+3.5}}
& 34.00 & & 78.57 {\scriptsize\textcolor{ForestGreen}{+44.6}}
& 78.00 & & 90.77 {\scriptsize\textcolor{ForestGreen}{+12.8}} \\

\midrule
DINOv3 ($k$-NN)
& 48.00 & & 100.00 {\scriptsize\textcolor{ForestGreen}{+52.0}}
& 94.00 & & 100.00 {\scriptsize\textcolor{ForestGreen}{+6.0}}
& 48.00 & & 71.43 {\scriptsize\textcolor{ForestGreen}{+23.4}}
& 74.00 & & 63.08 {\scriptsize\textcolor{red}{-10.9}} \\

EfficienNetV2-XL
& 48.00 & & 96.15 {\scriptsize\textcolor{ForestGreen}{+48.2}}
& 92.00 & & 100.00 {\scriptsize\textcolor{ForestGreen}{+8.0}}
& 60.00 & & 85.71 {\scriptsize\textcolor{ForestGreen}{+25.7}}
& 62.00 & & 55.38 {\scriptsize\textcolor{red}{-6.6}} \\

EfficienNet-L2
& 52.00 & & 100.00 {\scriptsize\textcolor{ForestGreen}{+48.0}}
& 96.00 & & 100.00 {\scriptsize\textcolor{ForestGreen}{+4.0}}
& 66.00 & & 82.14 {\scriptsize\textcolor{ForestGreen}{+16.1}}
& 74.00 & & 60.00 {\scriptsize\textcolor{red}{-14.0}} \\

EVA-02
& 52.00 & & 100.00 {\scriptsize\textcolor{ForestGreen}{+48.0}}
& 96.00 & & 100.00 {\scriptsize\textcolor{ForestGreen}{+4.0}}
& 60.00 & & 82.14 {\scriptsize\textcolor{ForestGreen}{+22.1}}
& 74.00 & & 60.00 {\scriptsize\textcolor{red}{-14.0}} \\

\bottomrule
\end{tabular}
\caption{Recall comparison between \gt{} and WeaselGT. 
\acp{mllm} (top) show large gains, often reaching perfect recall, indicating sensitivity to noisy or ambiguous \gt{} labels.
\acp{vlm} (center) also improve, reflecting reduced fine-grained sensitivity. 
Supervised models (bottom) sometimes drop, suggesting overfitting to \gt{}. 
% Overall, \acp{mllm} and \acp{vlm} appear more robust to annotation noise. 
\textit{Note: \acp{mllm} recall is computed only for the CW setup.}}

% \caption{Recall on \gt{} compared to the WeaselGT.  
% For \gt{}, we evaluate on 50 images per class based on the original annotations; for WeaselGT, we use the class assignments provided in \cite{kisel2025flaws}.  
% \acp{mllm} (top) show substantial recall increases under WeaselGT, often reaching perfect recall, indicating that many of their apparent ``errors'' under \gt{} stem from mislabeled or ambiguous images.  
% \textit{Note: for \acp{mllm}, the recall is computed only for the CW setup.}  
% \acp{vlm} (center) also improve, suggesting lingering sensitivity to finegrained distinctions.  
% Supervised models (bottom) sometimes exhibit substantial drops in recall on WeaselGT, suggesting that they overfit to \gt{}.  
% Overall, the shift from \gt{} to WeaselGT indicates that \acp{mllm} and \acp{vlm} are more robust to annotation noise.}

\label{tab:mink_recall}
\end{table*}

\subsection{Weasel family case study}

The reanotated dataset, as introduced in \cref{subsec:dataset}, only contains 625 out of the original 1000 classes, where the majority of wildlife, with the prominent exception of about 120 dog breeds, is excluded.

This is because wildlife is notoriously hard to annotate for non-experts, even if the annotators are trained.
To account for this limitation, we also introduce a case study on four classes from the weasel family: weasel, mink, polecat and black-footed ferret, whose reannotation conducted by an expert was introduced in \cite{kisel2025flaws}.
For a brief overview of the classes and image examples, see \cref{fig:case_study_examples}.

\paragraph{The weasel problem.}
Here we provide a brief overview of the classes and the main issues.
For more details, we refer the reader to the original publication.

Kisel et al.~\cite{kisel2025flaws} revisited four closely related mustelid classes in ImageNet and found severe problems arising from ambiguous synsets, inconsistent taxonomy, and extensive mislabeling. Their expert reannotation shows that these fine-grained wildlife categories cannot reliably serve as ground truth. The main issues are:

\begin{itemize}
    \item \textbf{weasel}: Synset corresponds to a broad colloquial category rather than a specific species; many images depict other mustelids (e.g., mink, ferrets), leading to more than half of the images being incorrect. However, the dominant species is the least weasel and highly similar species.
    In American English, the term weasel often refers to the weasel family as a whole, encompassing all four classes. 
    % The class is renamed to `least weasel'.
    \item \textbf{mink}: Although somewhat cleaner, many images still mix American and European mink and include other small mustelids due to visual similarity and poor source metadata.
    \item \textbf{polecat (Mustela putorius)}: The term ``polecat'' is ambiguous across English varieties and is also used for skunks; images contain a heterogeneous mix of species, and only about one-third match the intended European polecat. The European polecat is the ancestor of the domestic ferret, which makes them hard to distinguish, and they can also interbreed.
    \item \textbf{black-footed ferret (Mustela nigripes)}: Synset conflates the endangered wild species with domestic ferrets; almost all images show domestic animals rather than M.~nigripes, leaving the class effectively without valid examples. The class is renamed to `domestic ferret'.
\end{itemize}

These issues stem from WordNet synset definitions that do not align with real-world photographic data, making the original ImageNet labels highly unreliable for fine-grained species evaluation.

\paragraph{Performance on reannotated data.}
Across model families, accuracy generally increases when evaluated on the reannotated labels, with substantially larger gains on the 159-image WeaselGT subset, see \cref{tab:mink}. The improvements are most pronounced for \acp{mllm} and \acp{vlm}, indicating that many of their apparent errors under \gt{} labels stem from annotation noise. In contrast, supervised models show smaller gains, suggesting stronger anchoring to the original label space.
\cref{tab:mink_recall} complements this by reporting class-level recall under \gt{} labels versus WeaselGT labels.
Taken together, the tables show that careful reannotation is essential for reliably evaluating model performance on closely related wildlife categories.

\begin{table*}[ht]
\small
\centering
\begin{tabular}{lllcllllllll}
\toprule
& & \multicolumn{1}{c}{ImGT} & & \multicolumn{7}{c}{ReGT} \\ 
\cmidrule{3-3} \cmidrule{5-11} 
& Batch & $\mathrm{A_{625}}$ & & $\mathrm{A_{625}}$ & $\mathrm{S_{352}}$ & $\mathrm{{S+}_{316}}$ & $\mathrm{{S-}_{36}}$ & $\mathrm{M_{240}}$ & $\mathrm{{M+}_{221}}$ & $\mathrm{{M-}_{19}}$ & $\mathrm{OOP}$ \\
\cmidrule{1-12}

\multirow{3}{*}{LLaVA-OV} & 1 & 42.56 & & 52.00 & 46.31 & 48.10 & 30.56 & 53.75 & 55.20 & 36.84 & 168 \\
& 5 & 34.88 & & 42.56 & 37.78 & 40.19 & 16.67 & 41.67 & 43.44 & 21.05 & 233 \\
& 10 & 27.20 & & 35.36 & 30.97 & 32.28 & 19.44 & 32.92 & 34.39 & 15.79 & 258 \\ \midrule

\multirow{3}{*}{\qwen{}} & 1 & 64.96 & & 73.12 & 69.32 & 73.10 & 36.11 & 75.00 & 77.83 & 42.11 & 28 \\
& 5 & 63.52 & & 72.00 & 70.45 & 74.37 & 36.11 & 70.42 & 72.85 & 42.11 & 44 \\
& 10 & 64.16 & & 72.16 & 71.88 & 75.95 & 36.11 & 68.75 & 72.40 & 26.32 & 61 \\ \midrule

\multirow{3}{*}{\cgpt{}} & 1 & 73.92 & & 80.80 & 79.55 & 84.49 & 36.11 & 80.00 & 83.71 & 36.84 & 23 \\
& 5 & 74.40 & & 81.76 & 80.40 & 84.81 & 41.67 & 81.25 & 85.07 & 36.84 & 27 \\
& 10 & 74.24 & & 81.92 & 80.68 & 85.44 & 38.89 & 81.25 & 85.07 & 36.84 & 23  \\
\bottomrule
\end{tabular}
\caption{Image batch size comparison for \acp{mllm} on 625 sampled images (one per class) using batches of 1, 5, and 10. \qwen{} and \cgpt{} show consistent performance across batch sizes, while LLaVA-OV accuracy drops with larger batches. Thus, batch size 10 is used for \qwen{} and \cgpt{}, and 1 for LLaVA-OV.}

% \caption{Batch size comparison is conducted for \acp{mllm} across a randomly sampled subset of 625 images (one image per class) by sending batches of 1, 5, and 10 images in a single request. For \qwen{} and \cgpt{}, the results remain consistent across all batch sizes, with only negligible variation. In contrast, LLaVA-OV shows a significant decrease in accuracy as the batch size increases. Since the differences are minor for \qwen{} and \cgpt{} and larger batches are more resource-efficient, a batch size of 10 is selected for the experiments. In contrast, LLaVA-OV is evaluated using a batch size of 1 in all experiments.
% }
\label{tab:batch_comparison}
\end{table*}

\begin{table}[ht]
\small
\setlength{\tabcolsep}{4pt}
\centering
\begin{tabular}{llcccc}
\toprule
& & \multicolumn{1}{c}{\gt{}} & & \multicolumn{1}{c}{\regt{}} \\ 
\cmidrule{3-3} \cmidrule{5-5} 
Model & In-Batch Ordering & $\mathrm{A_{6250}}$ &  & $\mathrm{A_{6250}}$ & $\mathrm{OOP}$ \\
\cmidrule{1-6}

\multirow{2}{*}{\qwen{}}
& Random  & 63.47 & & 70.83 & 734 \\
& Same-Class & 76.96 & & 78.19 & 470 \\
\midrule
\multirow{2}{*}{\cgpt{}}
& Random  & 75.78 & & 80.82 & 324 \\
& Same-Class & 86.03 & & 84.61 & 155 \\

\bottomrule
\end{tabular}
\caption{
We compare random class mixtures to same-class batches on 6250 images (10 per class). Grouping by class causes \acp{mllm} to assign identical labels within a batch, inflating \gt{} accuracy and agreement with \regt{}. Random ordering reduces this bias and is used in all experiments with image-batched inputs.}
% \caption{
% We evaluated the effect of in-batch ordering by comparing random class mixtures with batches containing images from the same \gt{} class across a randomly sampled subset of 6250 images (10 images per class). Grouping images by class leads \acp{mllm} to frequently assign the same label to all images within a batch, thereby increasing \gt{} accuracy and alignment with the reannotated labels \regt{}. Random ordering mitigates this batch-class bias. Therefore, random in-batch ordering is used in all experiments for models that process batches of images per request.
% }
\label{tab:shuffle_table}
\end{table}

\begin{table}[ht]
\small
\centering
\setlength{\tabcolsep}{4pt}
\begin{tabular}{lllllll}
\toprule
& & & \multicolumn{1}{c}{ImGT} & & \multicolumn{1}{c}{ReGT} \\ 
\cmidrule{4-4} \cmidrule{6-6} 
& Img Pos. & Batch & \multicolumn{1}{c}{$\mathrm{A_{63(62)}}$} & & \multicolumn{1}{c}{$\mathrm{A_{63(62)}}$} & \multicolumn{1}{c}{$\mathrm{OOP}$} \\

\midrule

\multirow{9}{*}{\rotatebox[origin=c]{90}{LLaVA-OV}} & 1st & 1 & 44.44 &  & 60.32 & 18 \\
& 1st & 5 & 42.86 \textcolor{red}{-1.6} &  & 58.73 \textcolor{red}{-1.6} & 20 \textcolor{red}{+2} \\
& 1st & 10 & 33.33 \textcolor{red}{-11.1} &  & 52.38 \textcolor{red}{-7.9} & 23 \textcolor{red}{+5} \\
\cmidrule{2-7}
& 5th & 1 & 55.56 & & 69.84 & 11 \\
& 5th & 5 & 26.98 \textcolor{red}{-28.6} & & 34.92 \textcolor{red}{-34.9} & 28 \textcolor{red}{+17} \\
& 5th & 10 & 31.75 \textcolor{red}{-23.8} & & 39.68 \textcolor{red}{-30.2} & 25 \textcolor{red}{+14} \\
\cmidrule{2-7}
& 10th & 1 & 37.10 & & 50.00 & 22 \\
& 10th & 5 & 16.13 \textcolor{red}{-21.0} & & 29.03 \textcolor{red}{-21.0} & 38 \textcolor{red}{+16} \\
& 10th & 10 & 17.74 \textcolor{red}{-19.4} & & 30.65 \textcolor{red}{-19.4} & 33 \textcolor{red}{+11} \\
\midrule

\multirow{9}{*}{\rotatebox[origin=c]{90}{\qwen{}}} & 1st & 1 & 63.49 & & 73.02 & 1 \\
& 1st & 5 & 66.67 \textcolor{ForestGreen}{+3.2} & & 76.19 \textcolor{ForestGreen}{+3.2} & 2 \textcolor{red}{+1} \\
& 1st & 10 & 66.67 \textcolor{ForestGreen}{+3.2} & & 76.19 \textcolor{ForestGreen}{+3.2} & 2 \textcolor{red}{+1} \\
\cmidrule{2-7}
& 5th & 1 & 71.43 & & 85.71 & 0 \\
& 5th & 5 & 68.25 \textcolor{red}{-3.2} & & 80.95 \textcolor{red}{-4.8} & 5 \textcolor{red}{+5} \\
& 5th & 10 & 66.67 \textcolor{red}{-4.8} & & 77.78 \textcolor{red}{-7.9} & 5 \textcolor{red}{+5} \\
\cmidrule{2-7}
& 10th & 1 & 61.29 & & 72.58 & 7 \\
& 10th & 5 & 53.23 \textcolor{red}{-8.1} & & 69.35 \textcolor{red}{-3.2} & 8 \textcolor{red}{+1} \\
& 10th & 10 & 56.45 \textcolor{red}{-4.8} & & 70.97 \textcolor{red}{-1.6} & 9 \textcolor{red}{+2} \\
\midrule

\multirow{9}{*}{\rotatebox[origin=c]{90}{\cgpt{}}} & 1st & 1 & 69.84 &  & 79.37 & 2 \\
& 1st & 5 & 73.02 \textcolor{ForestGreen}{+3.2} &  & 82.54 \textcolor{ForestGreen}{+3.2} & 2 \\
& 1st & 10 & 69.84 &  & 79.37 & 1 \textcolor{ForestGreen}{-1} \\
\cmidrule{2-7}
& 5th & 1 & 76.19 &  & 88.89 & 1 \\
& 5th & 5 & 74.60 \textcolor{red}{-1.6} & & 88.89 & 3 \textcolor{red}{+2} \\
& 5th & 10 & 74.60 \textcolor{red}{-1.6} & & 85.71 \textcolor{red}{-3.2} & 2 \textcolor{red}{+1} \\
\cmidrule{2-7}
& 10th & 1 & 70.97 & & 83.87 & 4 \\
& 10th & 5 & 70.97 & & 85.48 \textcolor{ForestGreen}{+1.6} & 3 \textcolor{ForestGreen}{-1} \\
& 10th & 10 & 80.65 \textcolor{ForestGreen}{+9.7} & & 90.32 \textcolor{ForestGreen}{+6.5} & 3 \textcolor{ForestGreen}{-1} \\

\bottomrule
\end{tabular}
\vspace{1em}
\caption{Accuracy of the 1st, 5th, and 10th images from request batches of size 1, 5, and 10, evaluated on 625 sampled images (one per class). Image positions vary by batch size. \qwen{} and \cgpt{} show stable accuracy across positions, while LLaVA-OV’s accuracy declines for later images, consistent with its sensitivity to larger batches (see \cref{tab:batch_comparison}).}
% \caption{The 1st, 5th, and 10th images from each set of requests with batch sizes 1, 5, and 10 were selected, and their accuracies were computed. Original requests were performed across a randomly sampled subset of 625 images (one image per class). The image positions vary depending on the batch size: in batch size 1, the image is always first; in batch size 5, the 1st image is first while the 5th and 10th are last; and in batch size 10, the 1st image is first, the 5th is in the middle, and the 10th is last. The results show that \qwen{} and \cgpt{} maintain nearly stable accuracy regardless of image position, with only small deviations. In contrast, LLaVA-OV exhibits a noticeable decrease in accuracy for images appearing later in the batch, consistent with its sensitivity to larger batch sizes (see \cref{tab:batch_comparison}).}
\label{tab:context_batch}
\end{table}

\begin{table*}[ht]
\small
\centering
\begin{tabular}{lllcllllllll}
\toprule
& & \multicolumn{1}{c}{\gt{}} & & \multicolumn{7}{c}{\regt{}} \\ 
\cmidrule{3-3} \cmidrule{5-11} 
Model & Response &  $\mathrm{A_{625}}$ & & $\mathrm{A_{625}}$ & $\mathrm{S_{352}}$ & $\mathrm{{S+}_{316}}$ & $\mathrm{{S-}_{36}}$ & $\mathrm{M_{240}}$ & $\mathrm{{M+}_{221}}$ & $\mathrm{{M-}_{19}}$ & $\mathrm{OOP}$\\
\midrule
\multirow{2}{*}{LLaVA-OV} & ID & 7.68 & & 14.24 & 9.38 & 9.81 & 5.56 & 9.58 & 9.95 & 5.26 & 0 \\
& Class name & 42.56 {\scriptsize\textcolor{ForestGreen}{+34.9}} & & 52.00 {\scriptsize\textcolor{ForestGreen}{+37.8}} & 46.31 & 48.10 & 30.56 & 53.75 & 55.20 & 36.84 & 168 \\
\midrule
\multirow{2}{*}{InternVL3.5} & ID & 57.60 & & 67.20 & 62.22 & 64.24 & 44.44 & 70.00 & 72.85 & 36.84 & 6 \\
& Class name & 63.68 {\scriptsize\textcolor{ForestGreen}{+6.1}} & & 72.96 {\scriptsize\textcolor{ForestGreen}{+5.8}} & 67.05 & 69.30 & 47.22 & 77.92 & 80.54 & 47.37 & 6 \\
\midrule
\multirow{2}{*}{\qwen{}} & ID & 48.00 & & 57.28 & 54.55 & 56.96 & 33.33 & 55.42 & 58.37 & 21.05 & 0 \\
& Class name & 64.16 {\scriptsize\textcolor{ForestGreen}{+16.2}} & & 72.16 {\scriptsize\textcolor{ForestGreen}{+14.9}} & 71.88 & 75.95 & 36.11 & 68.75 & 72.40 & 26.32 & 61 \\
\midrule
\multirow{2}{*}{\cgpt{}} & ID & 70.24 & & 78.08 & 75.00 & 79.43 & 36.11 & 79.58 & 83.26 & 36.84 & 0  \\
& Class name & 74.24 {\scriptsize\textcolor{ForestGreen}{+4.0}} & & 81.92 {\scriptsize\textcolor{ForestGreen}{+3.8}} & 80.68 & 85.44 & 38.89 & 81.25 & 85.07 & 36.84 & 23 \\
\bottomrule
\end{tabular}
\vspace{1em}
\caption{Comparison of two response formats for \acp{mllm} on 625 sampled images (one per class): \textit{ID}: predicting a class ID via a mapping (\eg{} 0 -- ``tench'', \ldots) and \textit{Class Name}: selecting directly from class names. Accuracy \textcolor{ForestGreen}{deltas} relative to the ID format. While accuracy gains over the ID format vary by model, Class Name consistently performs better and can produce OOP outputs useful for CW+. We therefore use Class Name in all experiments.}

% \caption{We compare two response formats for \acp{mllm} across a randomly sampled subset of 625 images (one image per class): \textit{ID}, where the model outputs a class ID using a provided ID--class name mapping (\eg{} 0 -- ``tench'', \ldots), and \textit{Class Name}, where the model directly selects a class name from the supplied list. Although the \textcolor{ForestGreen}{accuracy deltas relative to the ID format} vary substantially across models, the Class Name setup yields higher accuracy for all models. Moreover, this setting occasionally produces out-of-prompt (OOP) outputs, which can later be leveraged in the CW+ setup. Therefore, the Class Name setup is used in all experiments.
% }
\label{tab:id_vs_name}
\end{table*}

\subsection{Preliminary experiments}
Research on image classification with \acp{mllm} remains limited, 
with only three methodologies proposed in the literature, corresponding to OW, MC, and CW tasks. 
There is little guidance on how to properly configure \acp{mllm} for such evaluations, 
and no prior studies have systematically examined the impact of different parameter settings. 
Therefore, we present an ablation study on the parameters influencing image classification using \acp{mllm}.

\paragraph{Image batch size.}
We evaluate different batch sizes by sending batches of 1, 5, and 10 images per request to the LLaVA-OV, \qwen{}, and \cgpt{} models. We do not perform this evaluation for InternVL3.5, as it only accepts a single image per request. The results in \cref{tab:batch_comparison} show that \qwen{} and \cgpt{} maintain nearly identical accuracy across all batch sizes, whereas LLaVA-OV experiences a dramatic drop in accuracy with larger batch sizes. Since batch size 10 provides stable results for both \cgpt{} and \qwen{} and is more resource-efficient, we select it for all remaining experiments. In contrast, LLaVA-OV is evaluated using batch size 1, as it would be unfair to assess it under conditions known to reduce its performance.

\paragraph{Effect of image position within a batch on accuracy.} 
The impact of image position within a batch is presented in \cref{tab:context_batch}. \qwen{} and \cgpt{} consistently maintain high accuracy regardless of image order, while LLaVA-OV experiences a significant drop in performance for images appearing later in the batch, consistent with its sensitivity to larger batch sizes.
%We additionally examine whether an image’s position within the batch affects accuracy (\cref{tab:context_batch}); \cgpt{} again remains stable across positions, whereas \qwen{} shows modest accuracy drops for later-positioned images, consistent with its sensitivity to larger batch sizes.

\paragraph{Composition bias.}
Using this fixed image batch size for \qwen{} and \cgpt{}, we evaluate the effect of in-batch image ordering on model behavior. We compare randomly mixed batches with batches containing only images from the same \gt{} class. As shown in \cref{tab:shuffle_table}, class-grouped batches frequently cause \acp{mllm} to assign the same label to every image, inflating \gt{} and \regt{} accuracy. Random ordering mitigates this batch-class bias; therefore, we adopt random ordering for all experiments with these models.

\paragraph{Class names or class IDs?}
With image batch size and ordering fixed, we compare two response formats: requiring the models to output a class ID from a provided mapping or directly output a Class Name from the supplied list. As shown in \cref{tab:id_vs_name}, \cgpt{} performs similarly in both settings, although the Class Name format occasionally produces out-of-prompt (OOP) outputs that can later be leveraged in the CW+ experiment. For LLaVA-OV, InternVL3.5, and \qwen{}, the Class Name format yields noticeably higher accuracy compared to the ID format, while still providing OOP predictions. Therefore, we use the Class Name format in all experiments.

%\paragraph{Stability under repeated evaluation.}
%To assess the stability of the \acp{mllm} outputs, we repeatedly evaluate \cgpt{} on the same 625-image subset, running the full inference procedure 31 times under identical settings. These repeated evaluations reveal small but consistent deviations in accuracy despite using temperature~0, as shown in \cref{tab:deviation}. This indicates that \cgpt{} is not perfectly deterministic in practice. The observed variability is non-zero but sufficiently small that it does not meaningfully affect the overall conclusions of our experiments. In contrast, \qwen{} produces identical predictions across repetitions, confirming fully deterministic behaviour under the same conditions.

\paragraph{Language-models (LMs) comparison for mapping.}
We evaluate different LMs for our OW setup: (i) Sentence-BERT \cite{reimers2019sentencebertsentenceembeddingsusing}, which is optimized for general-purpose sentence-level semantic similarity and provides embeddings aligned for textual comparison (following a similar approach to our OW semantic-similarity baseline in Conti et al.~\cite{conti2025largemultimodalmodelsopenworld}); (ii) SigLIP~2 \cite{tschannen2025siglip2multilingualvisionlanguage}, a state-of-the-art language model designed to produce high-quality, image-grounded text embeddings; and (iii) the newly released Qwen3 text encoder \cite{yang2025qwen3technicalreport}, tailored for multimodal models and naturally integrated with the Qwen3-VL architecture. The results are presented in \cref{tab:lang_knn_extended}. Templating improves performance for all encoders.

PaliGemma~2 and \cgpt{} perform best when paired with the SigLIP~2, while LLaVA-OV, InternVL3.5 and \qwen{} achieve the highest performance with Qwen3-Embedding-8B. The model-specific LM selected based on OW accuracy is used for evaluation in both the OW and CW+ setups in the main paper.

\paragraph{Distractor choice in the MC setup}
We explore several methods for the informed selection of challenging distractors for a class \texttt{c} in the 4-choice MC setup: (i) using the confusion matrix of the EVA-02 model, and (ii) selecting classes closest to \texttt{c} in the BERT embedding space, similar to Liu et al.~\cite{liu2024revisitingmllmsindepthanalysis}. The results are shown in \cref{tab:mc_ext_res}. 

The BERT-based method can yield highly challenging distractors (\eg{} distractors for ``computer keyboard'' include ``typewriter keyboard'', ``keyboard space bar'', and ``laptop computer''), but it may also produce distractors that are no more difficult than random choices (\eg{} the nearest classes to ``lens cap'' are ``swimming cap'', ``bottle cap'', and ``shower cap''). Overall, BERT-based distractors are more challenging than completely random selection, yet remain less difficult than those derived from the EVA-02. 

For example, in the pure ImGT with distractors setup (a common baseline for \acp{mllm} image recognition), \mbox{\cgpt{}'s} performance on ImGT drops nearly twice as much with EVA-02 distractors vs. BERT ones (90.66\% vs. 95.86\%), compared to random distractors (99.62\%).

\section{Additional Results}
\label{ap:exps}

\begin{table}[t!]
\centering
\small
\begin{tabular}{lccccc}
\toprule
&  
& \multicolumn{2}{c}{OOP} 
& \multicolumn{2}{c}{Correctly Mapped} \\
\cmidrule(lr){3-4} \cmidrule(lr){5-6}
& & \# & \% & \# & \% \\
\midrule

\multirow{8}{*}{\rotatebox[origin=c]{90}{InternVL3.5}}
& $\mathrm{A_{31250}}$ & 271 & 0.87 & 83 & 30.63 \\
& $\mathrm{S_{18071}}$ & 145 & 0.80 & 34 & 23.45 \\
& $\mathrm{{S+}_{16177}}$ & 130 & 0.80 & 34 & 26.15 \\
& $\mathrm{{S-}_{1894}}$ & 15 & 0.79 & 0 & 0.00 \\
& $\mathrm{M_{11834}}$ & 92 & 0.78 & 15 & 16.30 \\
& $\mathrm{{M+}_{10756}}$ & 79 & 0.73 & 12 & 15.19 \\
& $\mathrm{{M-}_{1078}}$ & 13 & 1.21 & 3 & 23.08 \\
& $\mathrm{{N}_{1345}}$ & 34 & 2.53 & -- & -- \\
\midrule

\multirow{8}{*}{\rotatebox[origin=c]{90}{\cgpt{}}}
& $\mathrm{A_{31250}}$ & 1647 & 5.27 & 554 & 33.64 \\
& $\mathrm{S_{18071}}$ & 717 & 3.97 & 142 & 19.80 \\
& $\mathrm{{S+}_{16177}}$ & 553 & 3.42 & 119 & 21.52 \\
& $\mathrm{{S-}_{1894}}$ & 164 & 8.66 & 23 & 14.02 \\
& $\mathrm{M_{11834}}$ & 710 & 6.00 & 192 & 27.04 \\
& $\mathrm{{M+}_{10756}}$ & 603 & 5.61 & 177 & 29.35 \\
& $\mathrm{{M-}_{1078}}$ & 107 & 9.93 & 15 & 14.02 \\
& $\mathrm{{N}_{1345}}$ & 220 & 16.36 & -- & -- \\
\midrule

\multirow{8}{*}{\rotatebox[origin=c]{90}{\qwen{}}}
& $\mathrm{A_{31250}}$ & 3290 & 10.53 & 1275 & 38.75 \\
& $\mathrm{S_{18071}}$ & 1671 & 9.25 & 497 & 29.74 \\
& $\mathrm{{S+}_{16177}}$ & 1410 & 8.72 & 456 & 32.34 \\
& $\mathrm{{S-}_{1894}}$ & 261 & 13.78 & 41 & 15.71 \\
& $\mathrm{M_{11834}}$ & 1294 & 10.93 & 453 & 35.01 \\
& $\mathrm{{M+}_{10756}}$ & 1135 & 10.55 & 421 & 37.09 \\
& $\mathrm{{M-}_{1078}}$ & 159 & 14.75 & 32 & 20.13 \\
& $\mathrm{{N}_{1345}}$ & 325 & 24.16 & -- & -- \\
\midrule

\multirow{8}{*}{\rotatebox[origin=c]{90}{LLaVA-OV}}
& $\mathrm{A_{31250}}$ & 8364 & 26.76 & 3720 & 44.48 \\
& $\mathrm{S_{18071}}$ & 4355 & 24.10 & 1797 & 41.26 \\
& $\mathrm{{S+}_{16177}}$ & 3823 & 23.63 & 1679 & 43.92 \\
& $\mathrm{{S-}_{1894}}$ & 532 & 28.09 & 118 & 22.18 \\
& $\mathrm{M_{11834}}$ & 3437 & 29.04 & 1351 & 39.31 \\
& $\mathrm{{M+}_{10756}}$ & 3101 & 28.83 & 1282 & 41.34 \\
& $\mathrm{{M-}_{1078}}$ & 336 & 31.17 & 69 & 20.54 \\
& $\mathrm{{N}_{1345}}$ & 572 & 42.53 & -- & -- \\

\bottomrule
\end{tabular}
\caption{
Overview of \acp{mllm} out-of-prompt across label categories. Counts (\#) indicate predictions falling outside the provided class names in the CW setup and the number correctly mapped (\#) in CW+. Harder label categories: M-- and S--, which do not contain the ImGT label, and N, which contains no label at all -- exhibit a higher ratio of OOP predictions.
}
\label{tab:qwen_ood}
\end{table}

% \multirow{7}{*}{\rotatebox[origin=c]{90}{GPT-4o}}
% & $\mathrm{A_{31,250}}$ & 81.29 & 25404 & 1647 & 5.27 & 307 & 18.64 \\
% & $\mathrm{S_{17,532}}$ & 81.41 & 14273 & 675 & 3.85 & 117 & 17.33 \\
% & $\mathrm{{S+}_{15,862}}$ & 85.76 & 13604 & 534 & 3.37 & 97 & 18.16 \\
% & $\mathrm{{S-}_{1,670}}$ & 40.06 & 669 & 141 & 8.44 & 20 & 14.18 \\
% & $\mathrm{M_{12,373}}$ & 79.09 & 9786 & 752 & 6.08 & 190 & 25.27 \\
% & $\mathrm{{M+}_{11,071}}$ & 84.56 & 9362 & 622 & 5.62 & 175 & 28.14 \\
% & $\mathrm{{M-}_{1,302}}$ & 32.57 & 424 & 130 & 9.98 & 15 & 11.54 \\
% & $\mathrm{{N}_{1,345}}$ & 100.00 & 1345 & 220 & 16.36 & -- & 0.00 \\
% \midrule

% \multirow{7}{*}{\rotatebox[origin=c]{90}{Qwen3-VL}}
% & $\mathrm{A_{31,250}}$ & 71.39 & 22309 & 3285 & 10.51 & 799 & 24.32 \\
% & $\mathrm{S_{17,532}}$ & 71.17 & 12478 & 1592 & 9.08 & 429 & 26.95 \\
% & $\mathrm{{S+}_{15,862}}$ & 74.57 & 11828 & 1367 & 8.62 & 386 & 28.24 \\
% & $\mathrm{{S-}_{1,670}}$ & 38.92 & 650 & 225 & 13.47 & 43 & 19.11 \\
% & $\mathrm{M_{12,373}}$ & 68.58 & 8486 & 1368 & 11.06 & 370 & 27.05 \\
% & $\mathrm{{M+}_{11,071}}$ & 73.01 & 8083 & 1175 & 10.61 & 347 & 29.53 \\
% & $\mathrm{{M-}_{1,302}}$ & 30.95 & 403 & 193 & 14.82 & 23 & 11.92 \\
% & $\mathrm{{N}_{1,345}}$ & 100.00 & 1345 & 325 & 24.16 & -- & 0.00 \\

\begin{table}[t!]
\centering
\setlength{\tabcolsep}{2.5pt}
\small
\begin{tabular}{lllll}
\toprule
Model & Partial & ImageNet & Abstain & Wrong \\
\midrule

InternVL3.5 & 21.77 & 0.0 & 1.11 & 77.12 \\

\cgpt{} & 35.34 & 3.22 & 6.38 & 55.07 \\

\qwen{} & 26.81 & 1.67 & 0.0 & 71.52 \\

LLaVA-OV & 36.29 & 2.59 & 0.04 & 61.08 \\

\bottomrule
\end{tabular}
\caption{
Distribution of out-of-prompt \acp{mllm} predictions. ``Partial'' denotes cases where the predicted string is a subset of any of the provided class names, regardless of word order. ``ImageNet'' corresponds to predictions that exactly match the commonly used OpenAI ImageNet-1k class names. ``Abstain'' refers to predictions in the form of ``I don't know'' or similar expressions (including LLM-generated variations). ``Wrong'' includes all remaining predictions.
}
\label{tab:oop_distr}
\end{table}
\begin{figure*}[t!]
    \centering
    \includegraphics[width=0.48\textwidth]{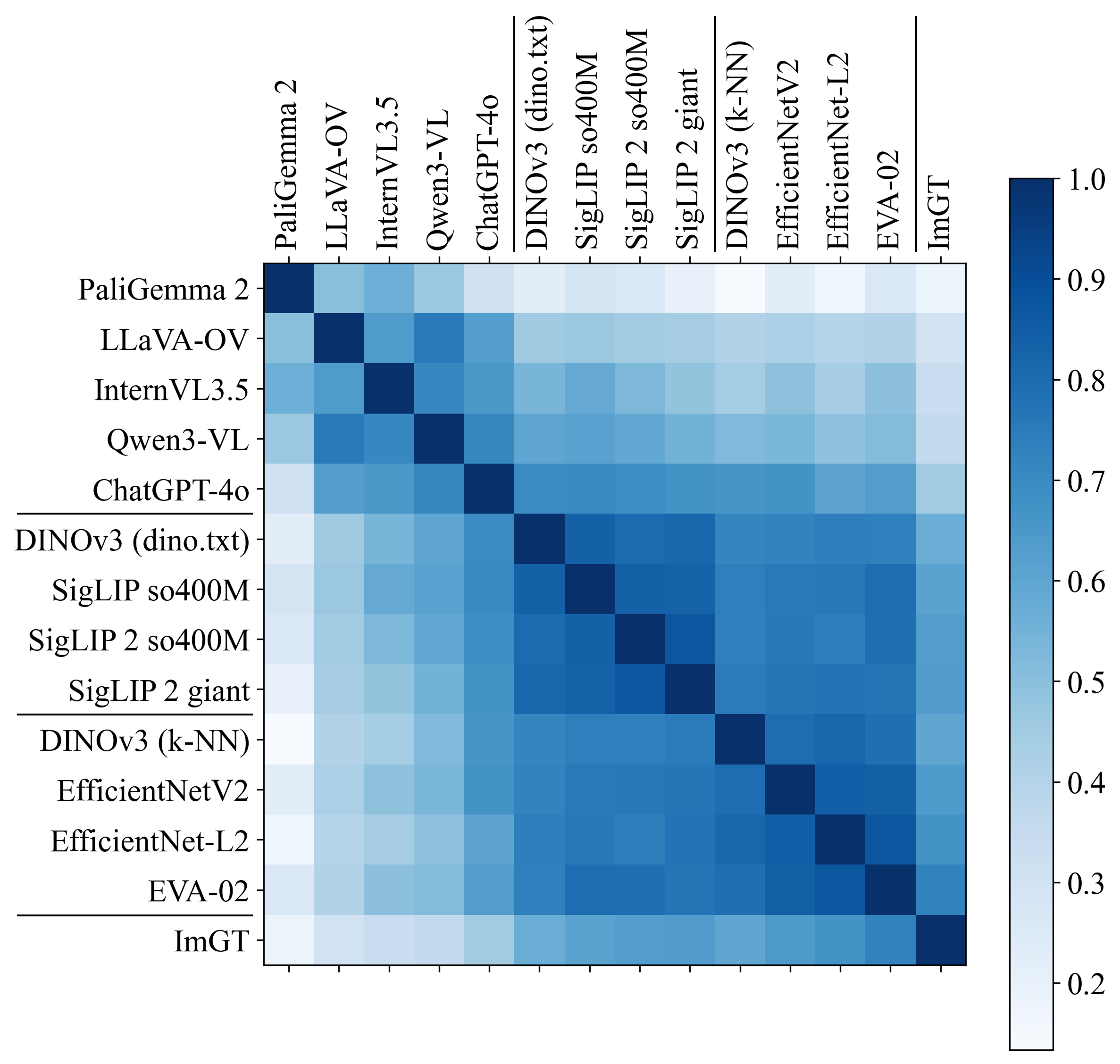}
    \hfill
    \includegraphics[width=0.48\textwidth]{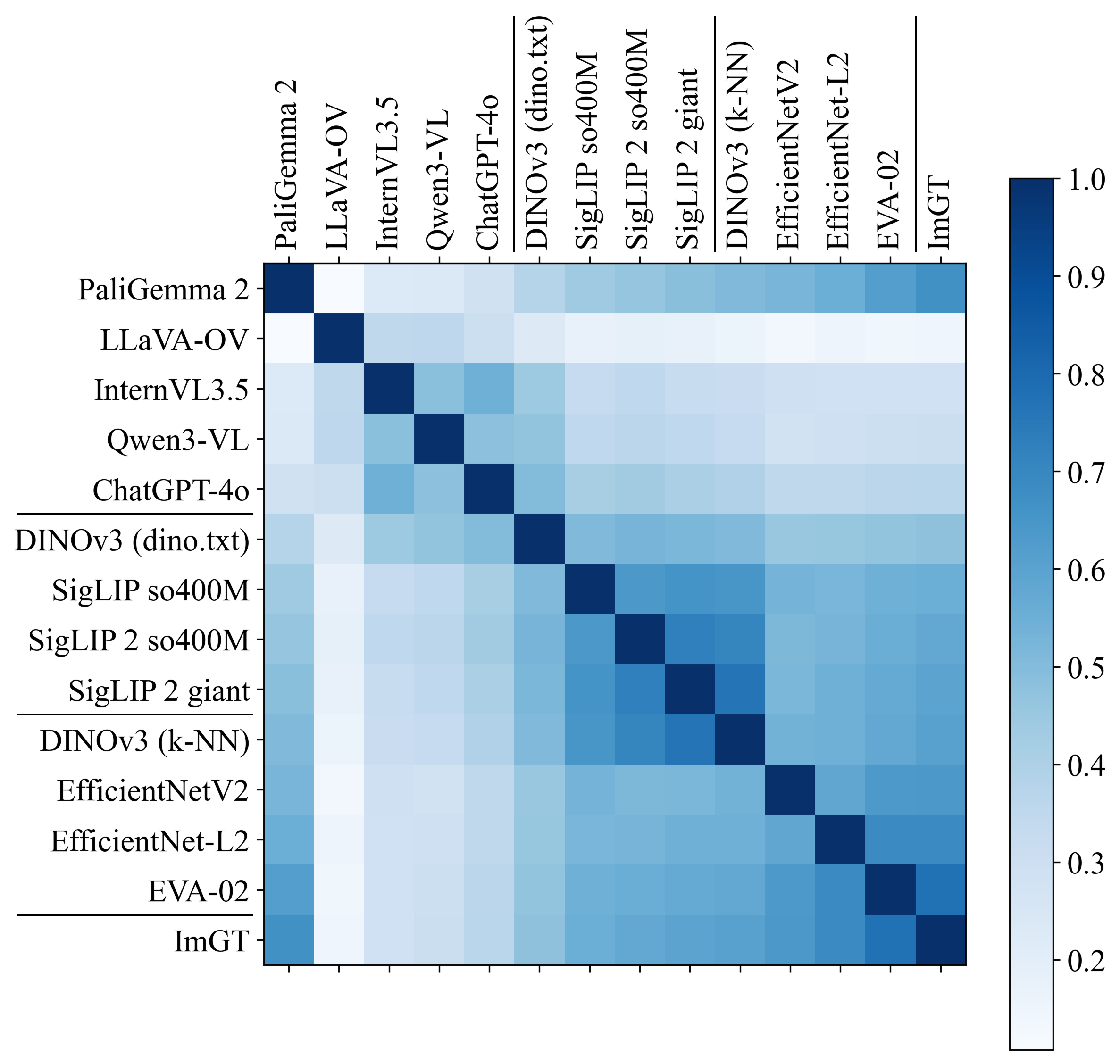}
    \caption{Correlation matrices across models evaluated on reannotated labels (\regt{}). 
    \textbf{Left:} Computes per-class recall for single-label images only and measures cross-model similarity using Spearman correlation. 
    \textbf{Right:} Builds an image correctness vector including all images with valid label(s) and computes cross-model similarity using the Phi coefficient. 
    The blocks indicate \acp{mllm}, \acp{vlm}, supervised models and original ground-truth (\gt{}).}
    \label{fig:recall_corr}
\end{figure*}

\begin{table*}[ht]
\scriptsize
\centering
\setlength{\tabcolsep}{3pt}
\begin{tabular}{llllllllllll}
\toprule

& & \multicolumn{1}{c}{\gt{}} & & \multicolumn{7}{c}{\regt{}} & \\
\cmidrule{3-3} \cmidrule{5-11}

Model & Batch & $\mathrm{A_{625}}$ & & $\mathrm{A_{625}}$ & $\mathrm{S_{352}}$ & $\mathrm{{S+}_{316}}$ & $\mathrm{{S-}_{36}}$ & $\mathrm{M_{240}}$ & $\mathrm{{M+}_{221}}$ & $\mathrm{{M-}_{19}}$ & $\mathrm{OOP}$ \\
\midrule{}
\cgpt{} 
& 10 & 74.69 \textsuperscript{\scriptsize \textcolor{gray}{±0.19}} & & 81.32 \textsuperscript{\scriptsize \textcolor{gray}{±0.18}} & 79.87 \textsuperscript{\scriptsize \textcolor{gray}{±0.19}} & 84.49 \textsuperscript{\scriptsize \textcolor{gray}{±0.21}} & 39.25 \textsuperscript{\scriptsize \textcolor{gray}{±0.82}} & 80.89 \textsuperscript{\scriptsize \textcolor{gray}{±0.35}} & 84.72 \textsuperscript{\scriptsize \textcolor{gray}{±0.36}} & 36.33 \textsuperscript{\scriptsize \textcolor{gray}{±0.58}} & 30.13 \textsuperscript{\scriptsize \textcolor{gray}{±1.11}} \\

\bottomrule
\end{tabular}
\vspace{1em}
\caption{Accuracy deviation over 31 repeated evaluations of a 625 sampled images (one per class) using \cgpt{} at temperature 0. Despite deterministic settings, small non-zero variability appears, shown by 95\,\% confidence intervals. LLaVA-OV, InternVL3.5, and \qwen{} are fully deterministic and omitted.}

% \caption{Accuracy deviation observed over 31 repeated evaluations of the same 625-image subset using the \cgpt{} model with temperature set to 0. Despite these deterministic settings, the model exhibits small but non-zero variability, reflected in the 95\,\% confidence intervals, which may cause slight differences when replicating the experiments. LLaVA-OV, InternVL3.5, and \qwen{} are fully deterministic and are therefore not included in this table.
% }
\label{tab:deviation}
\end{table*}

\subsection{Detailed overview of the OOP predictions}
We present the distribution of out-of-prompt \acp{mllm} predictions under the CW setup across four types in \cref{tab:oop_distr}. 
We also report the exact number of OOP predictions for each label category, along with the number correctly mapped (using the best model-specific LM) in the CW+ setup, in \cref{tab:qwen_ood}. Predictions in category $N$ do not require mapping, as any prediction for images in this category is considered correct in our setup.

%\subsection{Comparison of different setups}
%\cref{fig:compare_tasks} is presented to visualize the results reported in \cref{tab:ov_res} of the main paper.

\subsection{Correlation of recall and correctness patterns}  

To analyze similarity in model behavior, we compute two complementary correlation measures, corresponding to the two panels in \cref{fig:recall_corr}.

\paragraph{Per-class recall correlation (left).}
For each model, we compute per-class recall for every class in the reannotated image subset, restricting the analysis to single-label images only. Although the subset was reannotated for 625 classes, the images span 665 ImageNet-1k classes. We calculate the Spearman correlation between these per-class recall vectors across models. This correlation matrix captures similarity in class-wise error patterns rather than raw accuracy.

A clear structure emerges according to the training paradigm. Supervised models (EfficientNetV2, EfficientNet-L2, EVA-02) and the $k$-NN variant of DINOv3 form a tight, highly correlated cluster. \acp{vlm}: DINOv3 (dino.txt), SigLIP, and SigLIP~2 are strongly aligned with each other and moderately correlated with supervised models. In contrast, \acp{mllm} exhibit lower similarity to traditional vision models, indicating that their failure modes differ substantially from supervised and self-supervised approaches. We also compute per-class recall by treating \gt{} as predictions. This vector shows strong alignment with high-capacity supervised and self-supervised models.

\paragraph{Image-level correctness correlation (right).}
We also construct, for each model, a binary image-level correctness vector over all reannotated images with valid label(s), where each entry indicates whether the model prediction is correct. We compute pairwise similarity between these binary vectors using the Phi coefficient. This analysis operates at the image level and includes multi-label cases. The resulting matrix reflects agreement on which specific images are classified correctly or incorrectly. While the results differ slightly from the left panel, they still highlight the \acp{mllm} cluster, which is weakly correlated with other models.

\subsection{Stability under repeated evaluation}
The results of accuracy deviation are presented in \cref{tab:deviation}. With the temperature set to 0, \cgpt{} exhibits non-zero variability, whereas LLaVA-OV, InternVL3.5, and \qwen{} behave deterministically.

\begin{table*}[ht]
\small
\centering
\begin{tabular}{lllllllllll}
\toprule
& & \multicolumn{1}{c}{ImGT} & \multicolumn{7}{c}{ReGT} \\ 
\cmidrule{3-3} \cmidrule{5-11} 
& Emb. Space & $\mathrm{A_{31250}}$ & & $\mathrm{A_{31250}}$ & $\mathrm{S_{18071}}$ & $\mathrm{{S+}_{16177}}$ & $\mathrm{{S-}_{1894}}$ & $\mathrm{M_{11834}}$ & $\mathrm{{M+}_{10756}}$ & $\mathrm{{M-}_{1078}}$ \\

\midrule

\multirow{6}{*}{\rotatebox[origin=c]{90}{PaliGemma~2}} 
& Sentnce-BERT & 30.60 & & 41.41 & 32.58 & 34.01 & 20.38 & 48.24 & 49.58 & 34.88 \\
& Sentence-BERT$^\dagger$     & 31.89 & & 42.81 & 33.86 & 35.32 & 21.44 & 49.98 & 51.33 & 36.55   \\
& Qwen3-Embedding-8B & 33.24 & & 43.53 & 35.49 & 37.05 & 22.23 & 49.37 & 50.99 & 33.21 \\
& Qwen3-Embedding-8B$^\dagger$         & 36.52 & & 47.51 & 39.05 & 40.71 & \textbf{24.92} & 54.45 & 56.16 & 37.38  \\
& SigLIP~2 & 34.56 & & 45.04 & 36.45 & 38.23 & 21.28 & 51.90 & 53.51 & 35.81 \\
& SigLIP~2$^\dagger$      & \textbf{37.11} & & \textbf{47.94} & \textbf{39.74} & \textbf{41.49} & 24.76 & \textbf{54.55} & \textbf{56.24} & \textbf{37.66}  \\
\midrule

\multirow{6}{*}{\rotatebox[origin=c]{90}{LLaVA-OV}} 
& Sentence-Bert & 55.72 & & 64.56 & 61.13 & 64.42 & 33.05 & 65.76 & 68.60 & 37.38 \\
& Sentence-BERT$^\dagger$     & 56.57 & & 65.60 & 61.76 & 65.03 & 33.84 & 67.54 & 70.43 & 38.78   \\
& Qwen3-Embedding-8B & 57.56 & & 65.71 & 62.22 & 65.42 & 34.90 & 67.14 & 70.22 & 36.36 \\
& Qwen3-Embedding-8B$^\dagger$        & \textbf{62.00} & & \textbf{70.58} & 67.11 & 70.67 & 36.69 & \textbf{72.52} & \textbf{75.87} & 39.05   \\
& SigLIP~2 & 59.40 & & 67.24 & 64.58 & 68.12 & 34.42 & 67.57 & 70.59 & 37.38 \\
& SigLIP~2$^\dagger$       & 61.98 & & 70.35 & \textbf{67.38} & \textbf{70.84} & \textbf{37.91} & 71.50 & 74.67 & \textbf{39.89}   \\
\midrule

\multirow{6}{*}{\rotatebox[origin=c]{90}{InternVL3.5}} 
& Sentnce-BERT & 51.94 & & 61.15 & 55.51 & 58.48 & 30.15 & 65.33 & 68.34 & 35.25 \\
& Sentence-SBERT$^\dagger$     & 53.11 & & 62.63 & 56.75 & 59.71 & 31.47 & 67.36 & 70.30 & 38.03   \\
& Qwen3-Embedding-8B & 53.72 & & 62.07 & 56.54 & 59.51 & 31.20 & 66.21 & 69.20 & 36.36 \\
& Qwen3-Embedding-8B$^\dagger$         & \textbf{59.23} & & \textbf{68.18} & 62.44 & \textbf{65.78} & 33.90 & \textbf{73.33} & \textbf{76.66} & \textbf{40.07}   \\
& SigLIP~2 & 55.25 & & 63.09 & 58.67 & 61.85 & 31.47 & 65.65 & 68.70 & 35.25 \\
& SigLIP~2$^\dagger$      & 58.91 & & 67.16 & \textbf{62.53} & 65.76 & \textbf{34.90} & 70.51 & 73.72 & 38.50   \\
\midrule

\multirow{6}{*}{\rotatebox[origin=c]{90}{\qwen{}}}   
& Sentence-BERT & 57.65 & & 66.51 & 62.72 & 66.22 & 32.84 & 68.49 & 71.46 & 38.87 \\
& Sentence-BERT$^\dagger$ & 60.35 & & 69.21 & 65.41 & 69.05 & 34.32 & 71.51 & 74.65 & 40.26 \\
& Qwen3-Embedding-8B & 59.28 & & 66.89 & 62.45 & 65.96 & 32.42 & 69.90 & 73.21 & 36.83 \\
& Qwen3-Embedding-8B$^\dagger$ & \textbf{68.74} & & \textbf{76.68} & \textbf{73.61} & \textbf{77.79} & \textbf{37.91} & \textbf{78.71} & \textbf{82.46} & \textbf{41.28} \\
& SigLIP~2 & 59.27 & & 65.66 & 63.86 & 67.96 & 28.83 & 64.50 & 67.83 & 31.26  \\
& SigLIP~2$^\dagger$ & 65.82 & & 73.07 & 71.23 & 75.40 & 35.59 & 72.82 & 76.40 & 37.01 \\
\midrule

\multirow{6}{*}{\rotatebox[origin=c]{90}{\cgpt{}}}  
& Sentence-BERT & 64.14 & & 71.49 & 70.22 & 74.57 & 33.05 & 70.20 & 73.77 & 34.51  \\
& Sentence-BERT$^\dagger$ & 64.84 & & 72.58 & 71.03 & 75.28 & 34.69 & 71.83 & 75.51 & 35.06 \\
& Qwen3-Embedding-8B & 63.91 & & 70.59 & 68.62 & 72.72 & 33.63 & 70.25 & 74.00 & 32.84 \\
& Qwen3-Embedding-8B$^\dagger$ & 70.32 & & 77.29 & 75.61 & 80.23 & 36.22 & \textbf{77.28} & \textbf{81.41} & \textbf{36.09} \\
& SigLIP~2 ViT-gopt-384& 67.42 & & 73.94 & 73.11 & 77.41 & 36.38 & 72.26 & 76.20 & 32.93 \\
& SigLIP~2$^\dagger$ ViT-gopt-384& \textbf{70.92} & & \textbf{77.58} & \textbf{76.59} & \textbf{81.08} & \textbf{38.23} & 76.55 & 80.75 & 34.69 \\

\bottomrule
\end{tabular}
\vspace{1em}
\caption{Results for the OW setup with different language models (LMs) used for prediction mapping are reported. $\dagger{}$ denotes the use of templating following Radford et al.~\cite{radford2021learningtransferablevisualmodels}. Sentence-BERT is included, as it is used for semantic similarity in Conti et al.~\cite{conti2025largemultimodalmodelsopenworld}. Templating improves performance for all LMs. SigLIP~2 encoding performs best for PaliGemma~2 and \cgpt{}, while Qwen3-Embedding-8B encoding achieves the highest performance for LLaVA-OV, InternVL3.5, and \qwen{}. The model-specific LM selected based on OW accuracy is used for evaluation in both the OW and CW+ setups in the main paper.}
\label{tab:lang_knn_extended}
\end{table*}

\begin{table*}[ht]
\footnotesize
\setlength{\tabcolsep}{2pt}
\centering
\begin{tabular}{lllllllllll}
\toprule
% Model & Distractors & $\mathrm{Acc}_{gt}^{625}$ & $\mathrm{Acc}_A^{625}$ & $\mathrm{Acc}_S^{340}$ & $\mathrm{Acc}_{S+}^{309}$ & $\mathrm{Acc}_{S-}^{31}$ & $\mathrm{Acc}_M^{252}$ & $\mathrm{Acc}_{M+}^{228}$ & $\mathrm{Acc}_{M-}^{24}$ \\
& & \multicolumn{1}{c}{ImGT} & & \multicolumn{7}{c}{ReGT} \\ \cmidrule{3-3} \cmidrule{5-11} 
 & Distractors & $\mathrm{A_{625}}$ & & $\mathrm{A_{625}}$ & $\mathrm{S_{352}}$ & $\mathrm{{S+}_{316}}$ & $\mathrm{{S-}_{36}}$ & $\mathrm{M_{240}}$ & $\mathrm{{M+}_{221}}$ & $\mathrm{{M-}_{19}}$ \\
\midrule

% --- GPT-4o results first (multirow) ---

\multirow{8}{*}{\rotatebox[origin=c]{90}{LLaVA-OV}} 
& \gt{} + random & 99.28 \textsuperscript{\scriptsize \textcolor{gray}{±0.15}} & & 90.75 \textsuperscript{\scriptsize \textcolor{gray}{±0.11}} & 89.36 \textsuperscript{\scriptsize \textcolor{gray}{±0.13}} & 99.52 \textsuperscript{\scriptsize \textcolor{gray}{±0.14}} & 0.18 \textsuperscript{\scriptsize \textcolor{gray}{±0.25}} & 91.52 \textsuperscript{\scriptsize \textcolor{gray}{±0.18}} & 99.39 \textsuperscript{\scriptsize \textcolor{gray}{±0.20}} & 0.00 \textsuperscript{\scriptsize \textcolor{gray}{±0.00}} \\ 
& \gt{} + \texttt{confEVA(\gt{})} & 79.93 \textsuperscript{\scriptsize \textcolor{gray}{±0.32}} & & 76.02 \textsuperscript{\scriptsize \textcolor{gray}{±0.30}} & 73.13 \textsuperscript{\scriptsize \textcolor{gray}{±0.45}} & 81.46 \textsuperscript{\scriptsize \textcolor{gray}{±0.50}} & 0.00 \textsuperscript{\scriptsize \textcolor{gray}{±0.00}} & 76.96 \textsuperscript{\scriptsize \textcolor{gray}{±0.47}} & 83.58 \textsuperscript{\scriptsize \textcolor{gray}{±0.51}} & 0.00 \textsuperscript{\scriptsize \textcolor{gray}{±0.00}}  \\

& \gt{} + \texttt{confBERT(\gt{})} & 88.59 \textsuperscript{\scriptsize \textcolor{gray}{±0.29}} & & 82.65 \textsuperscript{\scriptsize \textcolor{gray}{±0.26}} & 79.66 \textsuperscript{\scriptsize \textcolor{gray}{±0.42}} & 88.73 \textsuperscript{\scriptsize \textcolor{gray}{±0.46}} & 0.00 \textsuperscript{\scriptsize \textcolor{gray}{±0.00}} & 84.65 \textsuperscript{\scriptsize \textcolor{gray}{±0.32}} & 91.93 \textsuperscript{\scriptsize \textcolor{gray}{±0.35}} & 0.00 \textsuperscript{\scriptsize \textcolor{gray}{±0.00}} \\

& ReGT + \texttt{confEVA(ReGT)} & 44.28 \textsuperscript{\scriptsize \textcolor{gray}{±0.26}} & & 78.36 \textsuperscript{\scriptsize \textcolor{gray}{±0.39}} & 81.42 \textsuperscript{\scriptsize \textcolor{gray}{±0.43}} & 81.49 \textsuperscript{\scriptsize \textcolor{gray}{±0.48}} & 80.83 \textsuperscript{\scriptsize \textcolor{gray}{±1.40}} & 70.90 \textsuperscript{\scriptsize \textcolor{gray}{±0.69}} & 71.77 \textsuperscript{\scriptsize \textcolor{gray}{±0.64}} & 60.78 \textsuperscript{\scriptsize \textcolor{gray}{±3.86}} \\

& ReGT + \texttt{confBERT(ReGT)} & 49.19 \textsuperscript{\scriptsize \textcolor{gray}{±0.25}} & & 84.92 \textsuperscript{\scriptsize \textcolor{gray}{±0.42}} & 89.08 \textsuperscript{\scriptsize \textcolor{gray}{±0.42}} & 89.21 \textsuperscript{\scriptsize \textcolor{gray}{±0.47}} & 87.99 \textsuperscript{\scriptsize \textcolor{gray}{±1.24}} & 76.75 \textsuperscript{\scriptsize \textcolor{gray}{±0.82}} & 76.57 \textsuperscript{\scriptsize \textcolor{gray}{±0.91}} & 78.78 \textsuperscript{\scriptsize \textcolor{gray}{±2.66}} \\

& \gt{} + ReGT + random & 89.99 \textsuperscript{\scriptsize \textcolor{gray}{±0.28}} & & 95.21 \textsuperscript{\scriptsize \textcolor{gray}{±0.17}} & 95.44 \textsuperscript{\scriptsize \textcolor{gray}{±0.21}} & 99.55 \textsuperscript{\scriptsize \textcolor{gray}{±0.14}} & 59.32 \textsuperscript{\scriptsize \textcolor{gray}{±1.38}} & 94.22 \textsuperscript{\scriptsize \textcolor{gray}{±0.27}} & 99.71 \textsuperscript{\scriptsize \textcolor{gray}{±0.16}} & 30.39 \textsuperscript{\scriptsize \textcolor{gray}{±2.89}} \\
& \gt{} + ReGT + \texttt{confEVA(\gt{},ReGT)} & 82.17 \textsuperscript{\scriptsize \textcolor{gray}{±0.31}} & & 90.01 \textsuperscript{\scriptsize \textcolor{gray}{±0.20}} & 87.78 \textsuperscript{\scriptsize \textcolor{gray}{±0.23}} & 90.74 \textsuperscript{\scriptsize \textcolor{gray}{±0.25}} & 61.74 \textsuperscript{\scriptsize \textcolor{gray}{±1.08}} & 91.92 \textsuperscript{\scriptsize \textcolor{gray}{±0.31}} & 97.23 \textsuperscript{\scriptsize \textcolor{gray}{±0.25}} & 30.22 \textsuperscript{\scriptsize \textcolor{gray}{±2.95}}  \\

& \gt{} + 999 (Closed World (CW)) & 42.56 \textsuperscript{\scriptsize \textcolor{gray}{±0.00}} & & 52.00 \textsuperscript{\scriptsize \textcolor{gray}{±0.00}} & 46.31 \textsuperscript{\scriptsize \textcolor{gray}{±0.00}} & 48.10 \textsuperscript{\scriptsize \textcolor{gray}{±0.00}} & 30.56 \textsuperscript{\scriptsize \textcolor{gray}{±0.00}} & 53.75 \textsuperscript{\scriptsize \textcolor{gray}{±0.00}} & 55.20 \textsuperscript{\scriptsize \textcolor{gray}{±0.00}} & 36.84 \textsuperscript{\scriptsize \textcolor{gray}{±0.00}} \\ \midrule

\multirow{8}{*}{\rotatebox[origin=c]{90}{InternVL3.5}} 
& \gt{} + random & 99.64 \textsuperscript{\scriptsize \textcolor{gray}{±0.11}} & & 90.97 \textsuperscript{\scriptsize \textcolor{gray}{±0.08}} & 89.54 \textsuperscript{\scriptsize \textcolor{gray}{±0.10}} & 99.72 \textsuperscript{\scriptsize \textcolor{gray}{±0.11}} & 0.18 \textsuperscript{\scriptsize \textcolor{gray}{±0.25}} & 91.83 \textsuperscript{\scriptsize \textcolor{gray}{±0.13}} & 99.72 \textsuperscript{\scriptsize \textcolor{gray}{±0.14}} & 0.00 \textsuperscript{\scriptsize \textcolor{gray}{±0.00}} \\

& \gt{} + \texttt{confEVA(\gt{})} & 83.72 \textsuperscript{\scriptsize \textcolor{gray}{±0.30}} & & 78.95 \textsuperscript{\scriptsize \textcolor{gray}{±0.24}} & 76.04 \textsuperscript{\scriptsize \textcolor{gray}{±0.34}} & 84.71 \textsuperscript{\scriptsize \textcolor{gray}{±0.38}} & 0.00 \textsuperscript{\scriptsize \textcolor{gray}{±0.00}} & 80.31 \textsuperscript{\scriptsize \textcolor{gray}{±0.40}} & 87.21 \textsuperscript{\scriptsize \textcolor{gray}{±0.44}} & 0.00 \textsuperscript{\scriptsize \textcolor{gray}{±0.00}} \\

& \gt{} + \texttt{confBERT(\gt{})} & 92.68 \textsuperscript{\scriptsize \textcolor{gray}{±0.22}} & & 85.42 \textsuperscript{\scriptsize \textcolor{gray}{±0.18}} & 82.64 \textsuperscript{\scriptsize \textcolor{gray}{±0.26}} & 92.06 \textsuperscript{\scriptsize \textcolor{gray}{±0.29}} & 0.00 \textsuperscript{\scriptsize \textcolor{gray}{±0.00}} & 87.50 \textsuperscript{\scriptsize \textcolor{gray}{±0.28}} & 95.02 \textsuperscript{\scriptsize \textcolor{gray}{±0.30}} & 0.00 \textsuperscript{\scriptsize \textcolor{gray}{±0.00}} \\

& ReGT + \texttt{confEVA(ReGT)} & 46.49 \textsuperscript{\scriptsize \textcolor{gray}{±0.19}} & & 82.83 \textsuperscript{\scriptsize \textcolor{gray}{±0.37}} & 84.41 \textsuperscript{\scriptsize \textcolor{gray}{±0.33}} & 84.71 \textsuperscript{\scriptsize \textcolor{gray}{±0.38}} & 81.81 \textsuperscript{\scriptsize \textcolor{gray}{±0.82}} & 78.14 \textsuperscript{\scriptsize \textcolor{gray}{±0.75}} & 79.10 \textsuperscript{\scriptsize \textcolor{gray}{±0.65}} & 67.06 \textsuperscript{\scriptsize \textcolor{gray}{±3.56}} \\

& ReGT + \texttt{confBERT(ReGT)} & 51.32 \textsuperscript{\scriptsize \textcolor{gray}{±0.19}} & & 88.97 \textsuperscript{\scriptsize \textcolor{gray}{±0.24}} & 91.84 \textsuperscript{\scriptsize \textcolor{gray}{±0.26}} & 92.06 \textsuperscript{\scriptsize \textcolor{gray}{±0.29}} & 89.88 \textsuperscript{\scriptsize \textcolor{gray}{±0.62}} & 83.24 \textsuperscript{\scriptsize \textcolor{gray}{±0.54}} & 83.35 \textsuperscript{\scriptsize \textcolor{gray}{±0.53}} & 82.00 \textsuperscript{\scriptsize \textcolor{gray}{±2.58}} \\

& \gt{} + ReGT + random & 91.98 \textsuperscript{\scriptsize \textcolor{gray}{±0.22}} & & 94.90 \textsuperscript{\scriptsize \textcolor{gray}{±0.15}} & 94.99 \textsuperscript{\scriptsize \textcolor{gray}{±0.16}} & 99.72 \textsuperscript{\scriptsize \textcolor{gray}{±0.11}} & 53.41 \textsuperscript{\scriptsize \textcolor{gray}{±1.20}} & 94.07 \textsuperscript{\scriptsize \textcolor{gray}{±0.30}} & 99.83 \textsuperscript{\scriptsize \textcolor{gray}{±0.10}} & 27.17 \textsuperscript{\scriptsize \textcolor{gray}{±3.60}} \\

& \gt{} + ReGT + \texttt{confEVA(\gt{},ReGT)} & 85.89 \textsuperscript{\scriptsize \textcolor{gray}{±0.29}} & & 90.53 \textsuperscript{\scriptsize \textcolor{gray}{±0.24}} & 88.78 \textsuperscript{\scriptsize \textcolor{gray}{±0.38}} & 92.86 \textsuperscript{\scriptsize \textcolor{gray}{±0.36}} & 52.96 \textsuperscript{\scriptsize \textcolor{gray}{±1.76}} & 91.79 \textsuperscript{\scriptsize \textcolor{gray}{±0.36}} & 97.35 \textsuperscript{\scriptsize \textcolor{gray}{±0.23}} & 27.17 \textsuperscript{\scriptsize \textcolor{gray}{±2.91}} \\

& \gt{} + 999 (Closed World (CW)) & 63.68 \textsuperscript{\scriptsize \textcolor{gray}{±0.00}} & & 72.96 \textsuperscript{\scriptsize \textcolor{gray}{±0.00}} & 67.05 \textsuperscript{\scriptsize \textcolor{gray}{±0.00}} & 69.30 \textsuperscript{\scriptsize \textcolor{gray}{±0.00}} & 47.22 \textsuperscript{\scriptsize \textcolor{gray}{±0.00}} & 77.92 \textsuperscript{\scriptsize \textcolor{gray}{±0.00}} & 80.54 \textsuperscript{\scriptsize \textcolor{gray}{±0.00}} & 47.37 \textsuperscript{\scriptsize \textcolor{gray}{±0.00}} \\

\midrule

% --- Qwen3-VL results after (multirow) ---
\multirow{8}{*}{\rotatebox[origin=c]{90}{\qwen{}}} 
& \gt{} + random & 99.34 \textsuperscript{\scriptsize \textcolor{gray}{±0.09}} & & 90.67 \textsuperscript{\scriptsize \textcolor{gray}{±0.08}} & 89.41 \textsuperscript{\scriptsize \textcolor{gray}{±0.08}} & 99.59 \textsuperscript{\scriptsize \textcolor{gray}{±0.09}} & 0.09 \textsuperscript{\scriptsize \textcolor{gray}{±0.18}} & 91.22 \textsuperscript{\scriptsize \textcolor{gray}{±0.17}} & 99.07 \textsuperscript{\scriptsize \textcolor{gray}{±0.18}} & 0.00 \textsuperscript{\scriptsize \textcolor{gray}{±0.00}} \\
 
& \gt{} + \texttt{confEVA(\gt{})} & 85.80 \textsuperscript{\scriptsize \textcolor{gray}{±0.29}} & & 80.74 \textsuperscript{\scriptsize \textcolor{gray}{±0.26}} & 77.67 \textsuperscript{\scriptsize \textcolor{gray}{±0.27}} & 86.52 \textsuperscript{\scriptsize \textcolor{gray}{±0.31}} & 0.00 \textsuperscript{\scriptsize \textcolor{gray}{±0.00}} & 82.59 \textsuperscript{\scriptsize \textcolor{gray}{±0.42}} & 89.70 \textsuperscript{\scriptsize \textcolor{gray}{±0.46}} & 0.00 \textsuperscript{\scriptsize \textcolor{gray}{±0.00}} \\

& \gt{} + \texttt{confBERT(\gt{})} & 93.13 \textsuperscript{\scriptsize \textcolor{gray}{±0.22}} & & 86.54 \textsuperscript{\scriptsize \textcolor{gray}{±0.17}} & 84.41 \textsuperscript{\scriptsize \textcolor{gray}{±0.27}} & 94.03 \textsuperscript{\scriptsize \textcolor{gray}{±0.30}} & 0.00 \textsuperscript{\scriptsize \textcolor{gray}{±0.00}} & 87.81 \textsuperscript{\scriptsize \textcolor{gray}{±0.23}} & 95.36 \textsuperscript{\scriptsize \textcolor{gray}{±0.25}} & 0.00 \textsuperscript{\scriptsize \textcolor{gray}{±0.00}}  \\

& ReGT + \texttt{confEVA(ReGT)} & 47.14 \textsuperscript{\scriptsize \textcolor{gray}{±0.17}} & & 82.01 \textsuperscript{\scriptsize \textcolor{gray}{±0.36}} & 86.40 \textsuperscript{\scriptsize \textcolor{gray}{±0.30}} & 86.52 \textsuperscript{\scriptsize \textcolor{gray}{±0.31}} & 85.39 \textsuperscript{\scriptsize \textcolor{gray}{±1.26}} & 73.09 \textsuperscript{\scriptsize \textcolor{gray}{±0.77}} & 73.92 \textsuperscript{\scriptsize \textcolor{gray}{±0.67}} & 63.50 \textsuperscript{\scriptsize \textcolor{gray}{±3.69}} \\

& ReGT + \texttt{confBERT(ReGT)} & 51.88 \textsuperscript{\scriptsize \textcolor{gray}{±0.18}} & & 88.57 \textsuperscript{\scriptsize \textcolor{gray}{±0.34}} & 93.70 \textsuperscript{\scriptsize \textcolor{gray}{±0.32}} & 94.03 \textsuperscript{\scriptsize \textcolor{gray}{±0.30}} & 90.77 \textsuperscript{\scriptsize \textcolor{gray}{±1.13}} & 79.48 \textsuperscript{\scriptsize \textcolor{gray}{±0.66}} & 79.39 \textsuperscript{\scriptsize \textcolor{gray}{±0.64}} & 80.47 \textsuperscript{\scriptsize \textcolor{gray}{±2.78}} \\

& \gt{} + ReGT + random & 91.95 \textsuperscript{\scriptsize \textcolor{gray}{±0.25}} & & 94.23 \textsuperscript{\scriptsize \textcolor{gray}{±0.11}} & 94.12 \textsuperscript{\scriptsize \textcolor{gray}{±0.14}} & 99.59 \textsuperscript{\scriptsize \textcolor{gray}{±0.09}} & 46.06 \textsuperscript{\scriptsize \textcolor{gray}{±1.46}} & 93.60 \textsuperscript{\scriptsize \textcolor{gray}{±0.21}} & 99.64 \textsuperscript{\scriptsize \textcolor{gray}{±0.11}} & 23.43 \textsuperscript{\scriptsize \textcolor{gray}{±2.48}} \\
& \gt{} + ReGT + \texttt{confEVA(\gt{},ReGT)} & 86.32 \textsuperscript{\scriptsize \textcolor{gray}{±0.26}} & & 90.77 \textsuperscript{\scriptsize \textcolor{gray}{±0.21}} & 89.14 \textsuperscript{\scriptsize \textcolor{gray}{±0.27}} & 93.86 \textsuperscript{\scriptsize \textcolor{gray}{±0.28}} & 47.67 \textsuperscript{\scriptsize \textcolor{gray}{±1.18}} & 91.88 \textsuperscript{\scriptsize \textcolor{gray}{±0.28}} & 97.65 \textsuperscript{\scriptsize \textcolor{gray}{±0.28}} & 24.79 \textsuperscript{\scriptsize \textcolor{gray}{±2.55}} \\

& \gt{} + 999 (Closed World (CW)) & 64.16 \textsuperscript{\scriptsize \textcolor{gray}{±0.00}} & & 72.16 \textsuperscript{\scriptsize \textcolor{gray}{±0.00}} & 71.88 \textsuperscript{\scriptsize \textcolor{gray}{±0.00}} & 75.95 \textsuperscript{\scriptsize \textcolor{gray}{±0.00}} & 36.11 \textsuperscript{\scriptsize \textcolor{gray}{±0.00}} & 68.75 \textsuperscript{\scriptsize \textcolor{gray}{±0.00}} & 72.40 \textsuperscript{\scriptsize \textcolor{gray}{±0.00}} & 26.32 \textsuperscript{\scriptsize \textcolor{gray}{±0.00}} \\ \midrule

\multirow{8}{*}{\rotatebox[origin=c]{90}{\cgpt{}}} 
& \gt{} + random & 99.62 \textsuperscript{\scriptsize \textcolor{gray}{±0.07}} & & 90.95 \textsuperscript{\scriptsize \textcolor{gray}{±0.05}} & 89.49 \textsuperscript{\scriptsize \textcolor{gray}{±0.07}} & 99.67 \textsuperscript{\scriptsize \textcolor{gray}{±0.07}} & 0.09 \textsuperscript{\scriptsize \textcolor{gray}{±0.18}} & 91.85 \textsuperscript{\scriptsize \textcolor{gray}{±0.10}} & 99.75 \textsuperscript{\scriptsize \textcolor{gray}{±0.11}} & 0.00 \textsuperscript{\scriptsize \textcolor{gray}{±0.00}} \\
& \gt{} + \texttt{confEVA(\gt{})} & 90.66 \textsuperscript{\scriptsize \textcolor{gray}{±0.22}} & & 85.12 \textsuperscript{\scriptsize \textcolor{gray}{±0.21}} & 84.27 \textsuperscript{\scriptsize \textcolor{gray}{±0.26}} & 93.87 \textsuperscript{\scriptsize \textcolor{gray}{±0.29}} & 0.00 \textsuperscript{\scriptsize \textcolor{gray}{±0.00}} & 84.31 \textsuperscript{\scriptsize \textcolor{gray}{±0.39}} & 91.56 \textsuperscript{\scriptsize \textcolor{gray}{±0.42}} & 0.00 \textsuperscript{\scriptsize \textcolor{gray}{±0.00}} \\

& \gt{} + \texttt{confBERT(\gt{})} & 95.86 \textsuperscript{\scriptsize \textcolor{gray}{±0.17}} & & 88.76 \textsuperscript{\scriptsize \textcolor{gray}{±0.12}} & 87.71 \textsuperscript{\scriptsize \textcolor{gray}{±0.19}} & 97.70 \textsuperscript{\scriptsize \textcolor{gray}{±0.21}} & 0.00 \textsuperscript{\scriptsize \textcolor{gray}{±0.00}} & 88.75 \textsuperscript{\scriptsize \textcolor{gray}{±0.18}} & 96.38 \textsuperscript{\scriptsize \textcolor{gray}{±0.20}} & 0.00 \textsuperscript{\scriptsize \textcolor{gray}{±0.00}}  \\

& ReGT + \texttt{confEVA(ReGT)} & 51.35 \textsuperscript{\scriptsize \textcolor{gray}{±0.21}} & & 84.26 \textsuperscript{\scriptsize \textcolor{gray}{±0.37}} & 92.45 \textsuperscript{\scriptsize \textcolor{gray}{±0.36}} & 93.90 \textsuperscript{\scriptsize \textcolor{gray}{±0.31}} & 79.66 \textsuperscript{\scriptsize \textcolor{gray}{±1.50}} & 70.09 \textsuperscript{\scriptsize \textcolor{gray}{±0.89}} & 70.78 \textsuperscript{\scriptsize \textcolor{gray}{±0.83}} & 62.14 \textsuperscript{\scriptsize \textcolor{gray}{±3.36}} \\

& ReGT + \texttt{confBERT(ReGT)} & 54.10 \textsuperscript{\scriptsize \textcolor{gray}{±0.12}} & & 89.37 \textsuperscript{\scriptsize \textcolor{gray}{±0.35}} & 96.72 \textsuperscript{\scriptsize \textcolor{gray}{±0.23}} & 97.75 \textsuperscript{\scriptsize \textcolor{gray}{±0.20}} & 87.63 \textsuperscript{\scriptsize \textcolor{gray}{±1.11}} & 77.14 \textsuperscript{\scriptsize \textcolor{gray}{±0.76}} & 76.91 \textsuperscript{\scriptsize \textcolor{gray}{±0.77}} & 79.80 \textsuperscript{\scriptsize \textcolor{gray}{±2.69}}  \\

& \gt{} + ReGT + random & 91.77 \textsuperscript{\scriptsize \textcolor{gray}{±0.26}} & & 94.89 \textsuperscript{\scriptsize \textcolor{gray}{±0.13}} & 94.79 \textsuperscript{\scriptsize \textcolor{gray}{±0.16}} & 99.67 \textsuperscript{\scriptsize \textcolor{gray}{±0.07}} & 51.97 \textsuperscript{\scriptsize \textcolor{gray}{±1.35}} & 94.31 \textsuperscript{\scriptsize \textcolor{gray}{±0.27}} & 99.80 \textsuperscript{\scriptsize \textcolor{gray}{±0.09}} & 30.56 \textsuperscript{\scriptsize \textcolor{gray}{±2.97}} \\

& \gt{} + ReGT + \texttt{confEVA(\gt{},ReGT)} & 88.31 \textsuperscript{\scriptsize \textcolor{gray}{±0.29}} & & 92.52 \textsuperscript{\scriptsize \textcolor{gray}{±0.15}} & 91.59 \textsuperscript{\scriptsize \textcolor{gray}{±0.20}} & 96.36 \textsuperscript{\scriptsize \textcolor{gray}{±0.15}} & 49.73 \textsuperscript{\scriptsize \textcolor{gray}{±1.27}} & 92.86 \textsuperscript{\scriptsize \textcolor{gray}{±0.31}} & 98.16 \textsuperscript{\scriptsize \textcolor{gray}{±0.20}} & 31.24 \textsuperscript{\scriptsize \textcolor{gray}{±3.30}}  \\

& \gt{} + 999 (Closed World (CW)) & 74.69 \textsuperscript{\scriptsize \textcolor{gray}{±0.19}} & & 81.32 \textsuperscript{\scriptsize \textcolor{gray}{±0.18}} & 79.87 \textsuperscript{\scriptsize \textcolor{gray}{±0.19}} & 84.49 \textsuperscript{\scriptsize \textcolor{gray}{±0.21}} & 39.25 \textsuperscript{\scriptsize \textcolor{gray}{±0.82}} & 80.89 \textsuperscript{\scriptsize \textcolor{gray}{±0.35}} & 84.72 \textsuperscript{\scriptsize \textcolor{gray}{±0.36}} & 36.33 \textsuperscript{\scriptsize \textcolor{gray}{±0.58}} \\

\bottomrule
\end{tabular}
\caption{4-way multiple-choice (MC) results across 625 sampled images (one per class), reported with 95\,\% confidence intervals. \texttt{confEVA()} and \texttt{confBERT()} denote functions that return challenging distractors based on the confusion matrix of EVA-02 and the distance between class names in the BERT embedding space, respectively. The Closed World result corresponds to using 999 distractors.}
\label{tab:mc_ext_res}
\end{table*}

\begin{table*}[ht]
\centering
\footnotesize 
\begin{tabular}{p{0.95\linewidth}}
\hline
\textbf{\cgpt{} Prompt} \\ \hline
{\ttfamily
You are an image classifier. \par
You will receive up to 50 images in order (image "1" = first, "2" = second, etc.). \par
You are also provided with a list of class names: \{class\_list\}. \vspace{0.8em}

Classification Rules: \par
- For each image, return the single class name that best represents the main subject of the image. \par
- Choose only one class per image - the most relevant or dominant one. \par
- Only return classes from the provided list. \vspace{0.8em}

Output Rules: \par
- Return exactly one output per image. \par
- Each output must be only a single class name (no separators or lists). \par
- Do not include explanations, confidence scores, or extra text. \par

} \\ \hline

\\ \rule{0pt}{1.2em} \\
\hline
\textbf{LLaVA-OV \& InternVL3.5 \& \qwen{} Prompt} \\ \hline
{\ttfamily
You are an image classifier. \par
You will receive up to 50 images in order (image "1" = first, "2" = second, etc.). \par
You are also provided with a list of class names: \{class\_list\}. \vspace{0.8em}

Your output will be automatically structured as JSON with keys "1", "2", "3", etc. corresponding to the order of images in the request. \par
Each value should be the predicted class name for that image. \vspace{0.8em}

Classification Rules: \par
- For each image, return the class name only from the provided list. \par
- Only return classes from the provided list. \vspace{0.8em}

Output Rules: \par
- Return exactly one JSON key per image ("1", "2", "3", etc.). \par
- Each value must be only class names. \par
- Do not include explanations, confidence scores, or extra text. \par

} \\ \hline
\end{tabular}
\caption{Prompts for the Closed World setup.}

\label{fig:cs_prompt}
\end{table*}

\begin{table*}[ht]
\centering
\footnotesize 
\begin{tabular}{p{0.95\linewidth}}
\hline
\textbf{\cgpt{} Prompt} \\ \hline
{\ttfamily
You are an open-set fine-grained image classifier. \par
You will receive up to 50 images in order (image "1" = first, "2" = second, etc.). \vspace{0.8em}

Classification Rules: \par
- For each image, identify the dominant object. \par
- Return the most fine-grained, specific label that accurately describes that object (e.g., "golden retriever puppy", "1950s red convertible", "blue morpho butterfly", "ceramic coffee mug with floral pattern"). \par
- Use natural-language labels that reflect detailed visual distinctions such as species, make/model, style, color, or material. \par
- Avoid generic terms like "dog", "car", or "bird" when a more specific subtype or description is visually inferable. \par
- If the dominant object cannot be clearly identified, return a concise descriptive label of its appearance (e.g., "abstract metal sculpture", "blurry human silhouette"). \par
- Focus only on the dominant object, even if multiple are present. \vspace{0.8em}

Output Rules: \par
- Return exactly one output per image. \par
- The output must contain only the final label (no punctuation beyond normal text, no explanations, confidence scores, or extra text). \par

} \\ \hline

\\ \rule{0pt}{1.2em} \\
\hline
\textbf{LLaVA-OV \& InternVL3.5 \& \qwen{} Prompt} \\ \hline
{\ttfamily
You are an open-set fine-grained image classifier. \par
You will receive up to 50 images in order (image "1" = first, "2" = second, etc.). \vspace{0.8em}

Your output will be automatically structured as JSON with keys "1", "2", "3", etc. corresponding to the order of images in the request. \par
Each value should be the predicted label for that image. \vspace{0.8em}

Classification Rules: \par
- For each image, identify the dominant object. \par
- Return the most fine-grained, specific label that accurately describes that object (e.g., "golden retriever puppy", "1950s red convertible", "blue morpho butterfly", "ceramic coffee mug with floral pattern"). \par
- Use natural-language labels that reflect detailed visual distinctions such as species, make/model, style, color, or material. \par
- Avoid generic terms like "dog", "car", or "bird" when a more specific subtype or description is visually inferable. \par
- If the dominant object cannot be clearly identified, return a concise descriptive label of its appearance (e.g., "abstract metal sculpture", "blurry human silhouette"). \par
- Focus only on the dominant object, even if multiple are present. \vspace{0.8em}

Output Rules: \par
- Return exactly one output per image. \par
- The output must contain only the final label (no punctuation beyond normal text, no explanations, confidence scores, or extra text). \par

} \\ \hline

\end{tabular}
\caption{Prompts for the Open World setup.}

\label{fig:ov_prompt}
\end{table*}

\begin{table*}[ht]
\centering
\footnotesize 
\begin{tabular}{p{0.95\linewidth}}
\hline
\textbf{LLaVA-OV \& InternVL3.5 \& \qwen{} \& \cgpt{} Prompt} \\ \hline
{\ttfamily
You are an image classifier. You will receive one image. You are also provided with four multiple-choice options (A, B, C, D). \vspace{0.8em}

What is the main object in this image? \{dynamic\_choices\} \vspace{0.8em}

Classification Rules:\par
- For the image, return the letter (A, B, C, or D) that corresponds to the correct option. \par
- Only return one letter. \vspace{0.8em}

Output Rules:\par
- Return exactly one letter (A, B, C, or D).\par
- Do not include explanations, the class name, or extra text.\par
- Your answer must be only the letter.\par

} \\ \hline
\end{tabular}
\caption{Prompt for the Multiple-Choice setup. The same prompt is used for all \acp{mllm}.
}
\label{fig:mc_prompt}
\end{table*}

% 1. Define Image Dimensions
\newcommand{\IMGW}{2.8cm}
\newcommand{\IMGH}{2.8cm}
% New: Fixed height for the label area (enough for 2 lines of text)
\newcommand{\LABELH}{1.1cm} 

% 3. The Magic Macro
\newcommand{\mergedpair}[5]{%
    \begin{minipage}[t]{0.48\textwidth}
        \centering
        % --- BLOCK 1: IMAGES (Fixed Height) ---
        % We lock this area to exactly \IMGH
        \begin{minipage}[t][\IMGH][t]{\linewidth}
            \centering
            \begin{minipage}[b]{0.48\linewidth}\centering
                \includegraphics[width=\linewidth,height=\IMGH,keepaspectratio]{#1}
            \end{minipage}%
            \hfill
            \begin{minipage}[b]{0.48\linewidth}\centering
                \includegraphics[width=\linewidth,height=\IMGH,keepaspectratio]{#3}
            \end{minipage}
        \end{minipage}
        
        % Space between image and label
        \vspace{0.1em}
        
        % --- BLOCK 2: LABELS (Fixed Height) ---
        % We lock this area to exactly \LABELH. 
        % This ensures the description below always starts at the same level.
        \begin{minipage}[t][\LABELH][t]{\linewidth}
            \centering
            \begin{minipage}[t]{0.48\linewidth}\centering
                \ttfamily \small #2
            \end{minipage}%
            \hfill
            \begin{minipage}[t]{0.48\linewidth}\centering
                \ttfamily \small #4
            \end{minipage}
        \end{minipage}
        
        % --- BLOCK 3: DESCRIPTION ---
        % No italics, exactly as you requested
        \vspace{-0.5em} % Tweak this if you want the text closer/further from labels
        \justifying
        \noindent \small #5
    \end{minipage}
}

\begin{figure*}[ht]
\centering

\begin{tabular}{cc}

% --- ROW 1 ---
% Note: I removed \vspace{3em} from here because it breaks column alignment. 
\mergedpair{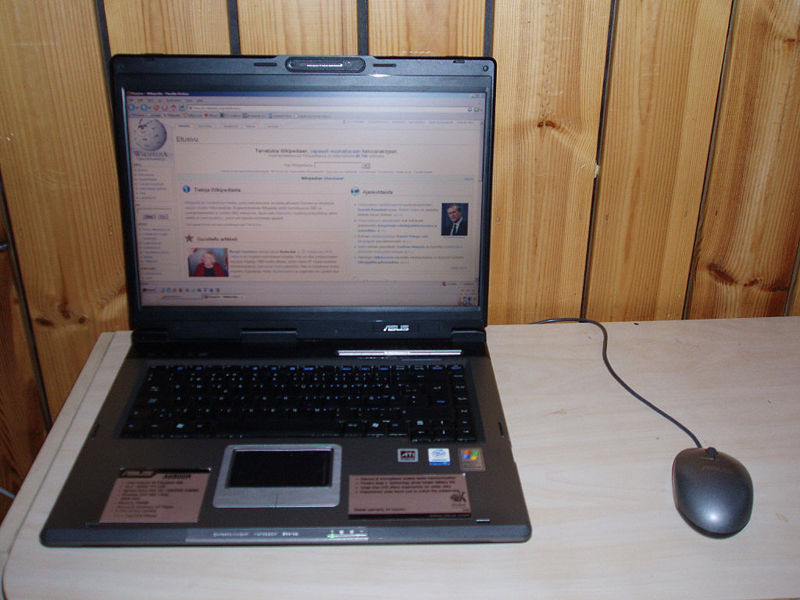}{620 - laptop computer}
           {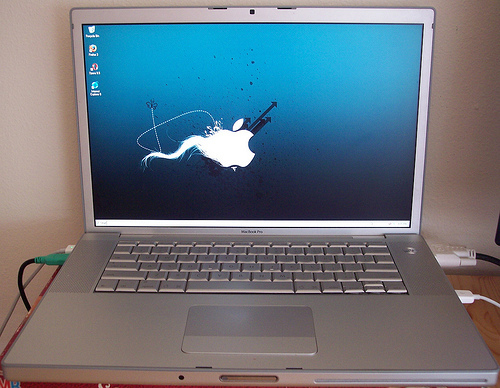}{681 - notebook computer}
           {In the modern context, these terms effectively mean the same thing. Distinguishing between them solely from visual features is close to impossible. The image content of the classes is the same.}
&
\mergedpair{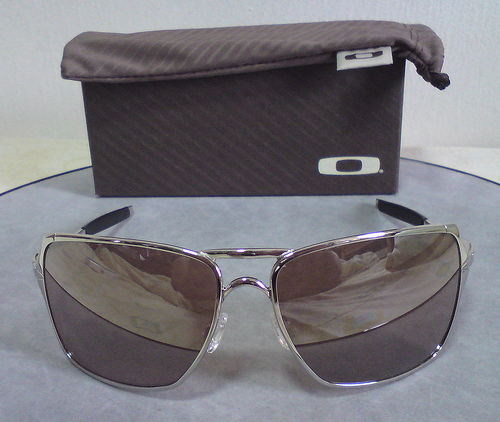}{836 - sunglass}
           {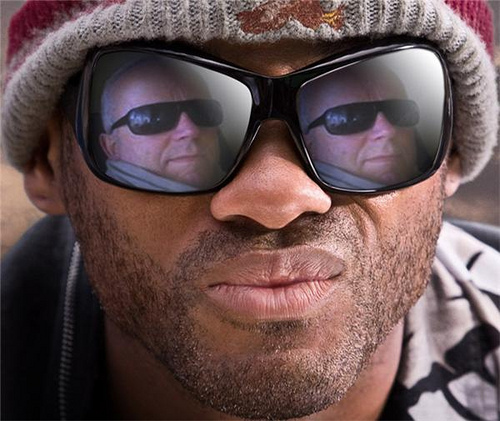}{837 - sunglasses}
           {Sunglass is defined as ``a convex lens that focuses the rays of the sun; used to start a fire'' in WordNet, the difference was lost in the original annotation process. The image content of the classes is the same.} 
\\\noalign{\vspace{2em}} 

% --- ROW 2 ---
\mergedpair{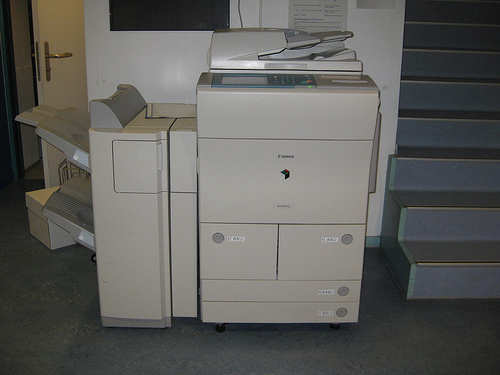}{742 - printer}
           {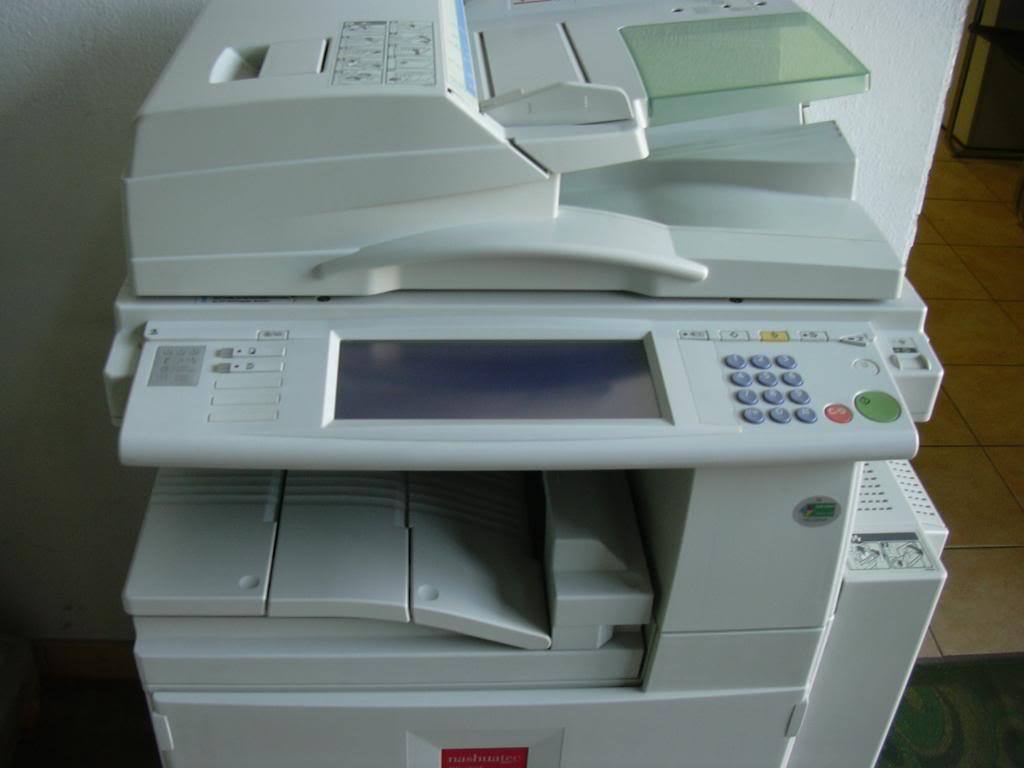}{713 - photocopier}
           {Printers and photocopiers used to be separate devices, but there are also multi-functional ones. Distinguishing them visually is difficult, and in modern context, stand-alone photocopiers are no longer commonly used.} % and dedicated standalone photocopiers have largely fallen out of regular use.
&
\mergedpair{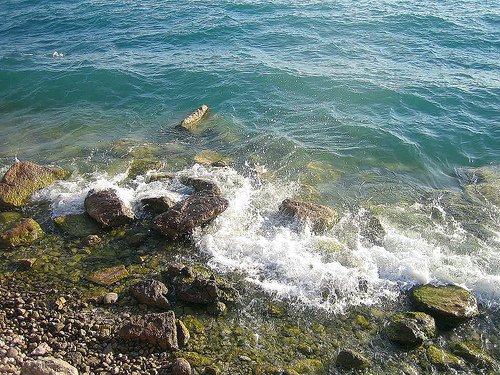}{975 - lakeshore}
           {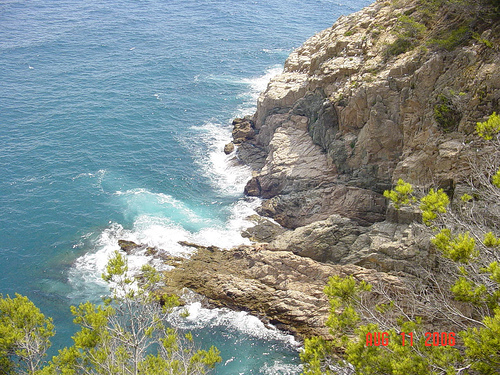}{978 - seashore}
           {It’s difficult to identify whether the water comes from ocean, sea, lake or river, unless it is a well-known geographical location or a characteristic wildlife can be observed.} 
\\ \\\noalign{\vspace{2em}} 

% --- ROW 3 ---
\mergedpair{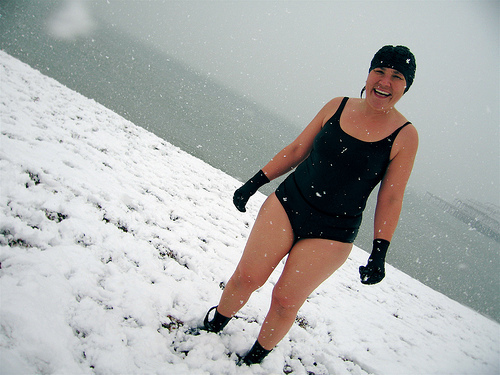}{638 - tights/leotard}
           {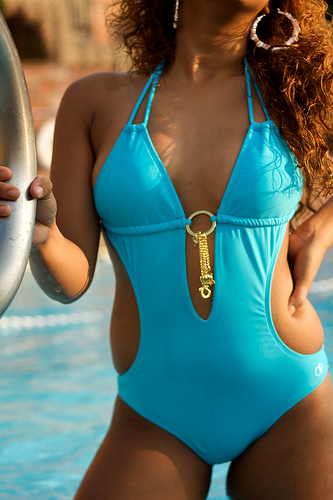}{639 - bathing suit}
           {The original meaning of the first term was lost during annotation, similar to what happened with ``sunglass''. As a result, the image content of the classes is now indistinguishable based on visual features.} 
&
\mergedpair{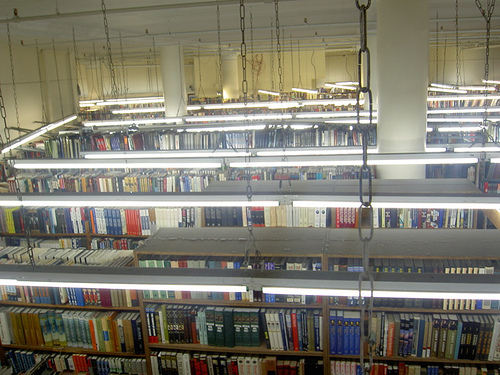}{454 - bookstore}
           {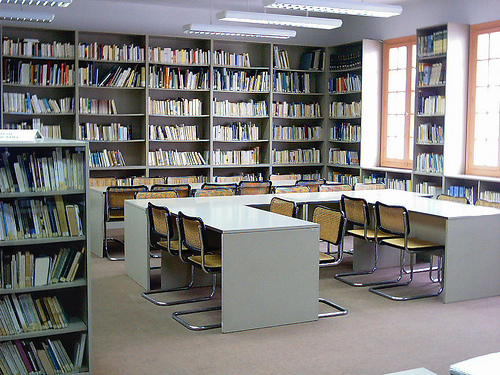}{624 - library}
           {It is challenging to distinguish a bookstore from a library solely from an interior view, unless a distinguishing object like a cash register is visible. This leaves room for ambiguity and speculation.} 
\\ 
\end{tabular}

\caption{Class pairs (part 1/2) considered equivalent during evaluation (\ie for a pair of classes $\{c_1, c_2\}$ and an image labeled $c_1$, a predicted label $c_2$ is also considered correct). Each pair is accompanied by a brief note explaining the decision.}
\label{fig:equal_classes_uniform_1}
\end{figure*}

\begin{figure*}[ht]
\centering

\begin{tabular}{cc}

% --- ROW 1: Missile & Breastplate ---
\mergedpair{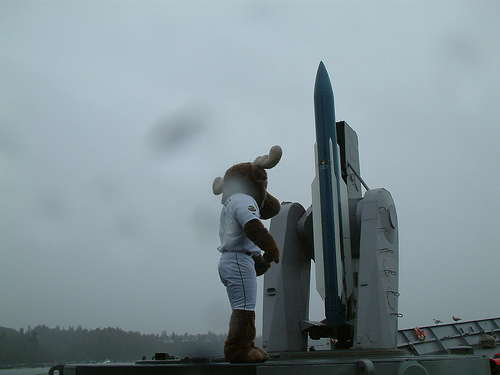}{657 - missile}
           {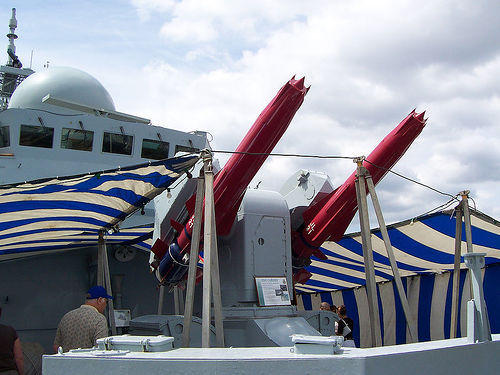}{744 - projectile, missile}
           {The image content of the classes is mostly the same: the majority of the images are missiles.} 
&
\mergedpair{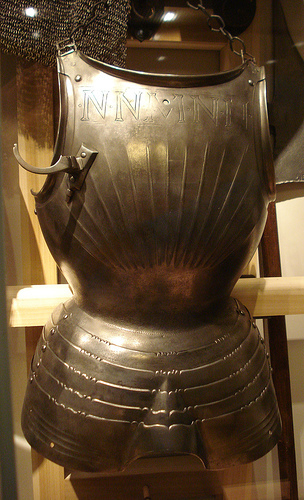}{461 - breastplate}
           {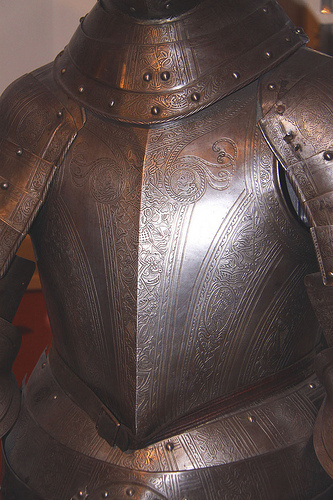}{524 - cuirass}
           {Breastplate only provides front coverage, while a cuirass covers the whole body. They share the same visual features from the front view, which is the predominant one. } 
\\ \noalign{\vspace{2em}} 

% --- ROW 2: Bathtub & Husky ---
\mergedpair{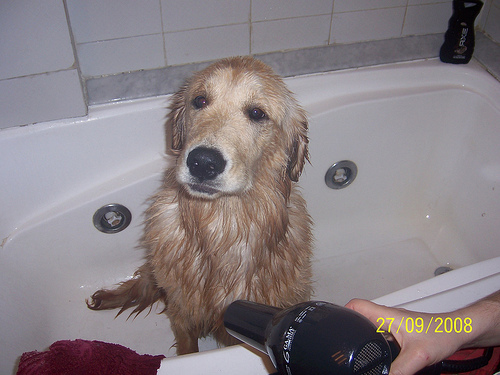}{435 - bathtub}
           {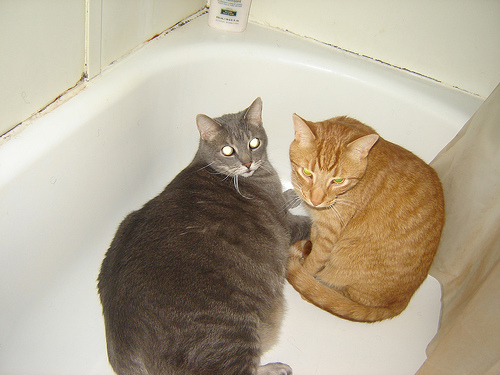}{876 - tub, vat}
           {The image content of the classes is mostly the same.} 
&
\mergedpair{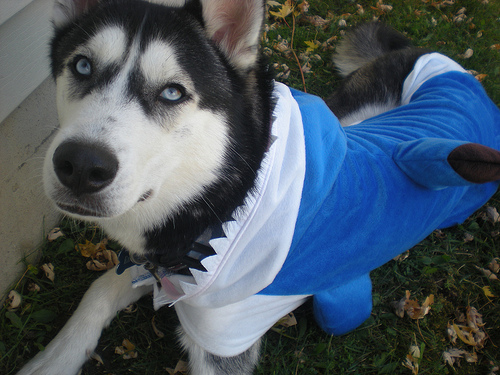}{248 - Eskimo dog}
           {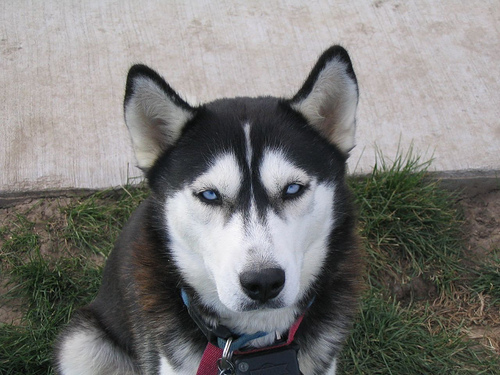}{250 - Siberian Husky}
           {The majority of the ``Eskimo dog'' class are huskies.} 
\\ \noalign{\vspace{2em}} 

% --- ROW 3: Cassette Player & Pitcher ---
\mergedpair{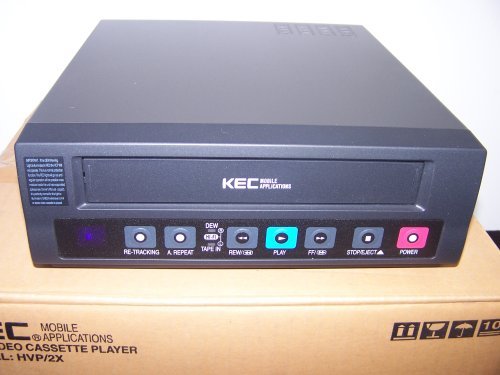}{482 - cassette player}
           {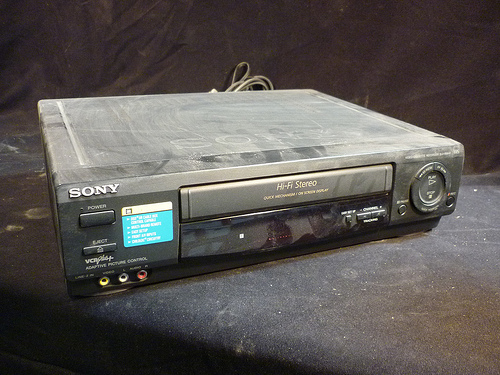}{848 - tape player}
           {Distinguishing between a cassette player and tape player from images is often unreliable. The terms are commonly used interchangeably, as the visual differences are minimal.} 
&
\mergedpair{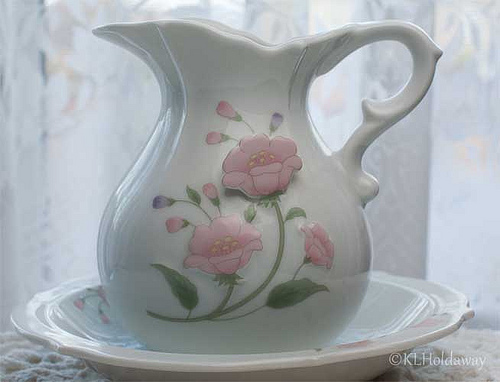}{899 - water jug}
           {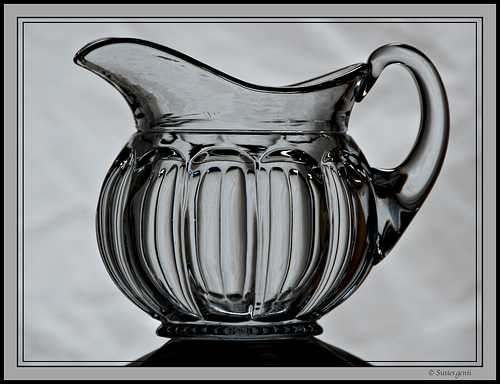}{725 - drink pitcher}
           {Both classes represent containers for pouring liquids with identical structures; it is hard to tell when a water jug becomes a pitcher and vice versa.} 
\\ 

\end{tabular}

\caption{Class pairs (part 2/2) considered equivalent during evaluation (\ie for a pair of classes $\{c_1, c_2\}$ and an image labeled $c_1$, a predicted label $c_2$ is also considered correct). Each pair is accompanied by a brief note explaining the decision.}
\label{fig:equal_classes_uniform_2}
\end{figure*}

\end{document}